\pdfoutput=1

\documentclass[11pt]{article}

\usepackage[final]{acl}

\usepackage{times}
\usepackage{latexsym}
\usepackage{pdfpages}

\usepackage[T1]{fontenc}

\usepackage[utf8]{inputenc}

\usepackage{microtype}
\usepackage{comment}
\usepackage{pdfpages}
\usepackage{inconsolata}
\usepackage{subcaption}
\usepackage{graphicx}

\title{Let’s Play Across Cultures: A Large Multilingual, Multicultural Benchmark for Assessing Language Models' Understanding of Sports}

\author{
  \textbf{Punit Kumar Singh}$^{1}$, \quad
  \textbf{Nishant Kumar}$^{1}$,\quad
  \textbf{Akash Ghosh}$^{1}$,\quad
  \textbf{Kunal Pasad}$^{2}$,\\
  \textbf{Khushi Soni}$^{2}$,\quad
  \textbf{Manisha Jaishwal}$^{1}$,\quad
  \textbf{Sriparna Saha}$^{1}$,\quad
  \textbf{Syukron Abu Ishaq Alfarozi}$^{3}$,\\
  \textbf{Asres Temam Abagissa}$^{1}$,\quad
  \textbf{Kitsuchart Pasupa}$^{4}$,\quad
  \textbf{Haiqin Yang}$^{5}$\footnotemark[1],\quad
  \textbf{Jose G Moreno}$^{6}$\footnotemark[1]
  \\
  {\small
  $^{1}$Indian Institute of Technology Patna, India \quad
  $^{2}$Sardar Patel Institute of Technology, Mumbai
  }
  \\
  {\small
  $^{3}$Universitas Gadjah Mada, Indonesia \quad
  $^{4}$King Mongkut's Institute of Technology Ladkrabang, Thailand 
  }
  \\
  {\small
  $^{5}$Shenzhen Technology University, China
  \quad
  $^{6}$Université de Toulouse, France
  }
}

\begin{document}
\maketitle
\renewcommand{\thefootnote}{\fnsymbol{footnote}}
\footnotetext[1]{Corresponding authors: \texttt{yanghaiqin@sztu.edu.cn} and \texttt{jose.moreno@irit.fr}.}
\begin{abstract}
Language Models (LMs) are primarily evaluated on globally popular sports, often overlooking regional and indigenous sporting traditions. To address this gap, we introduce \textbf{\textit{CultSportQA}}, a benchmark designed to assess LMs' understanding of traditional sports across 60 countries and 6 continents, encompassing four distinct cultural categories. The dataset features 33,000 multiple-choice questions (MCQs) across text and image modalities, each of which is categorized into three key types: history-based, rule-based, and scenario-based. To evaluate model performance, we employ zero-shot, few-shot, and chain-of-thought (CoT) prompting across a diverse set of Large Language Models (LLMs), Small Language Models (SLMs), and Multimodal Large Language Models (MLMs). By providing a comprehensive multilingual and multicultural sports benchmark, \textbf{\textit{CultSportQA}} establishes a new standard for assessing AI’s ability to understand and reason about traditional sports.

\end{abstract}

\section{Introduction}

\begin{figure*}
\includegraphics[height = 6 cm,width=1\textwidth]{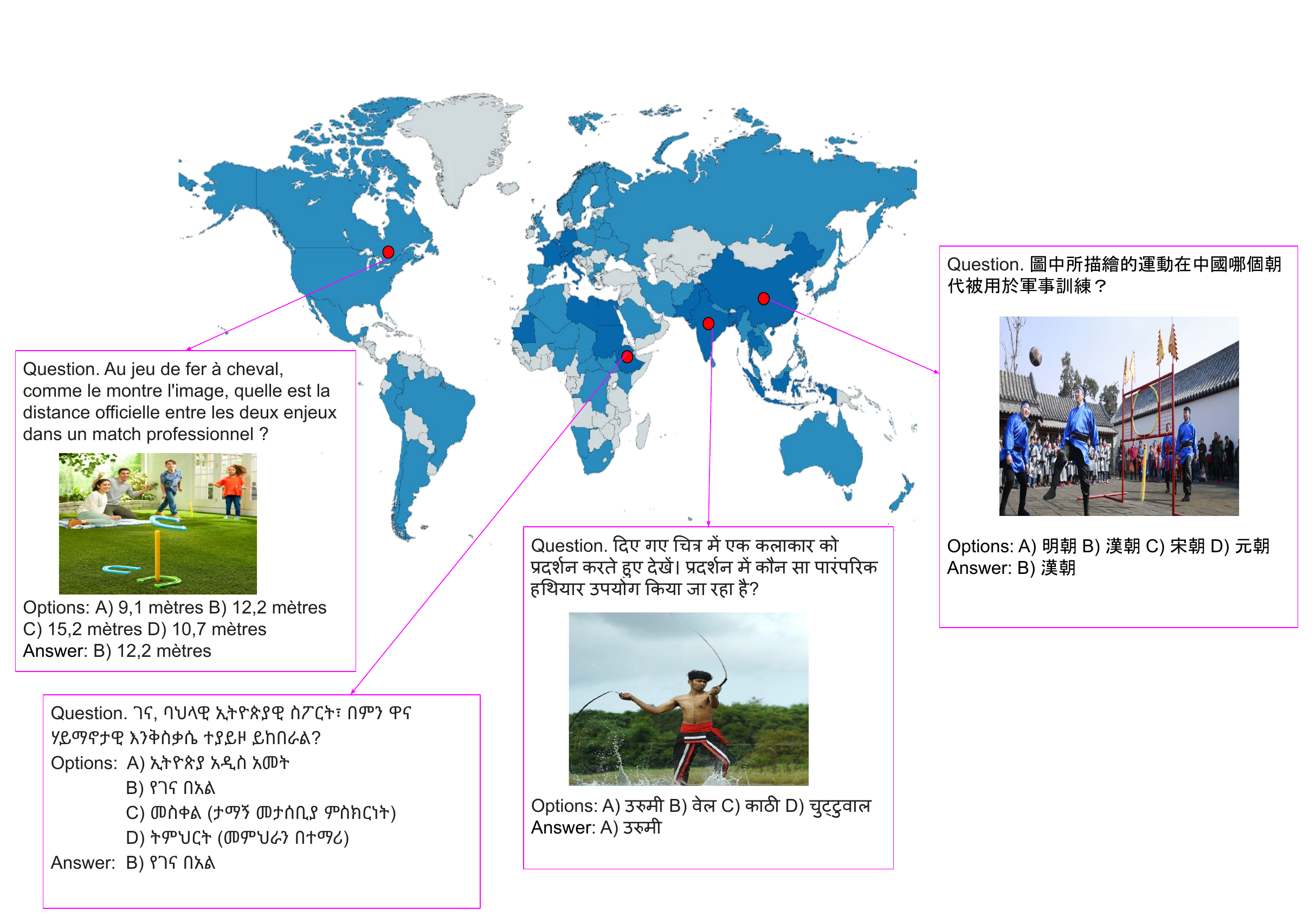}
\caption{\textbf{CultSportQA} is a diverse benchmark featuring 11 languages, with questions manually created and verified by native language experts. It covers three key aspects of traditional sports across two modalities, text and image, emphasizing mid to low-resource languages and sports originating from 11 countries across 3 continents. These sports, now played in 60 countries across 6 continents, are depicted with dark blue for their origins and light blue for their current reach. \textbf{CultSportQA} offers a wide range of question formats, including multiple-choice questions (MCQs) and both short and long visual question-answering (VQA) tasks.
} 
\label{CultSportQA_wm}
\end{figure*}
Sports serve as a powerful medium for cultural exchange, uniting people across diverse backgrounds and traditions \cite{Coakley2021}. The study of various sports and athletic practices provides valuable insights into societal values, historical narratives, and social structures of the communities that develop and embrace them \cite{Guttmann2004}. Furthermore, sports play a crucial role in shaping language, acting as a conduit for cultural knowledge and identity formation \cite{Maguire2011}. Sports' terminology, rituals, and adaptations showcase community history, societal change, and cultural identity \cite{Dyck2012}.\par
Researchers have long utilized sports as a lens to analyze cultural dynamics, providing a framework for quantifying differences in athletic traditions across regions \cite{Bairner2015}. Many sports share fundamental principles but have evolved uniquely in different societies, leading to variations in rules, playing styles, and even terminologies \cite{Eichberg2010}. Different nations have adapted bat-and-ball games uniquely, such as baseball, cricket, and pesäpallo, with distinct rules. Similarly, "football" varies in meaning across regions, referring to American football, soccer, or Australian rules football \cite{Mangan1996}.

Language Models (LMs) have revolutionized natural language understanding, content generation, and decision-making, becoming indispensable across industries such as education, governance, and entertainment, healthcare \cite{jain2022survey,brown2020language, kenton2019bert,ghosh2024clipsyntel,ghosh2024healthalignsumm,ghosh2025infogen,ghosal2025relic}. From Large Language Models (LLMs) to Multimodal Language Models (MLMs) and Small Language Models (SLMs) \footnote{Any model with less than or equal to 7B parameters are considered as Small Language Models(SLMs)}, these advancements have enabled seamless communication and efficient problem-solving \cite{ouyang2022training,ghosh2024exploring}. However, a persistent challenge remains: ensuring that these models effectively recognize and reason about diverse linguistic and cultural contexts, particularly in underrepresented domains such as traditional sports \cite{bender2021dangers}.\par

Traditional and indigenous sports are deeply intertwined with local histories, societal values, and cultural identities \cite{blodgett2020language}. Despite their significance, LMs are predominantly trained and evaluated on globally popular sports, often overlooking regional variations and culturally unique athletic traditions. This oversight risks reinforcing biases, inaccuracies, and stereotypes, further marginalizing underrepresented communities. Conversely, models capable of understanding cultural contexts not only enhance performance but also promote inclusivity and equity in AI applications.\par

\par

\textbf{Motivation for \textbf{\textit{CultSportQA}} Dataset:} Existing benchmarks in sports understanding and reasoning primarily focus on globally well-known sports and are often limited in scope. For instance, \textit{SportQA} \cite{xia2024sportqa} is an unimodal dataset that only supports English and covers widely recognized sports. Similarly, \textit{SportU} \cite{xia2024sportu} is the first dataset benchmarking multimodal large language models (MLLMs), but it includes only seven globally popular sports, all in English. However, no existing benchmark comprehensively captures the cultural nuances of sports reasoning across multiple languages, diverse cultural contexts, and visual question answering (VQA). To bridge this gap, we introduce the \textbf{largest multicultural and multilingual sports benchmark} to date, namely, \textbf{\textit{CultSportQA}}, featuring approximately 33,000 sports-related questions spanning countries originating from 3 continents and 11 countries and now being expanded in 6 continents and across 60 countries\footnote{The complete list is present in the Appendix}. Our benchmark evaluates Large Language Models (LLMs), Small Language Models (SLMs)\footnote{Any model below 7B parameters is considered a Small Language Model in this work.}, and Multimodal Language Models (MLLMs) across eleven languages. The questions are systematically categorized across two modalities and each modality is subdivided into three key types of questions: history-based, rule-based, and scenario-based\footnote{The MLLMs are used only for benchmarking image-based questions.}. By offering a comprehensive, culturally diverse, and multilingual benchmark, \textbf{\textit{CultSportQA}} establishes a new standard for assessing the ability of language models to understand and reason about sports in a more inclusive, global, and culturally aware manner.
\textbf{Research Questions:} This research aims to address the following key questions:

\begin{itemize}
    \item \textit{How do different categories of models—Large Language Models (LLMs), Small Language Models (SLMs), and Multimodal Large Language Models (MLMs)—perform on the \textbf{\textit{CultSportQA}} dataset?}
    \item \textit{What trends and patterns emerge in model performance across the various question types in the \textbf{\textit{CultSportQA}} dataset, including two modalities, text and image, and each modality covering three key types of questions: history-based, rule-based, and scenario-based}
    \item \textit{What are the performance trends of language models across different countries and languages in Asia, Africa, and Europe?}
\end{itemize}

Our \textbf{key contributions} in this research are summarized as follows:

\textbf{1. \textbf{{\em CultSportQA}} Dataset:} We introduce the first and most comprehensive QA dataset focusing on traditional sports, covering games played across 60 countries, 6 continents, and 11 languages.

\textbf{2. Diverse Question Types:} The dataset includes 33,000 questions spanning two modalities of data and each modality covers three categories, challenging AI models to reason through textual and visual input while incorporating multilingual and cultural contexts.

\textbf{3. Comprehensive Benchmarking:} We evaluate 8 state-of-the-art LLMs and five SLMs alongside four MLLMs, identifying critical gaps in their ability to reason about traditional and culturally sports-nuanced queries.

\textbf{4. Insights on AI Performance:} Using zero-shot, few-shot learning, and chain-of-thought (CoT) prompting, we analyze model strengths and limitations, advancing the understanding of AI performance in culturally rich domains.

\textbf{5. Expanding NLP in Sports:} Our work explores new applications of NLP in preserving cultural heritage, enriching sports journalism, and enhancing communication between athletes and coaches, particularly in regional and traditional sports contexts.

\textbf {6. Public Availability:} The CultSportQA dataset
 is available at: \url{https://github.com/M-Groot7/CultSportQA}.

By addressing the challenges of cultural underrepresentation in AI, {\em CultSportQA } establishes itself as a robust benchmark for evaluating and improving AI systems. This research contributes to fostering inclusivity and equity in AI applications while advancing the intersection of NLP and culturally rich domains like traditional sports.

\section{Related Work}
\subsection{Sports Datasets and Benchmarks}
Sports datasets are rapidly expanding, enabling diverse applications, such as sentiment analysis \cite{baca2023deep, ljajic2015sentiment}, game prediction, and video enhancement using computer vision \cite{beal2021combining, oved2020predicting}. While datasets like \textbf{SportQA} \cite{xia2024sportqa} and \textbf{BoolQ} \cite{clark2019boolq} have significantly advanced sports-related question answering (QA), many existing datasets primarily focus on historical events and overlook critical aspects such as \textbf{rules, strategies, and complex situational analysis} \cite{oved2020predicting, huang2020generating}. For instance, the \textbf{Sports Understanding} subtask in \textbf{BIG-bench (2023)} assesses \textbf{athlete recognition and action identification} but lacks depth in \textbf{situational comprehension}. Among existing benchmarks, \citet{xia2024sportqa} introduced one of the largest \textbf{unimodal text-based} datasets, covering approximately \textbf{70,000 questions in English}. Meanwhile, \citet{xia2024sportu} developed a \textbf{multimodal sports dataset} with \textbf{12,048 questions}, benchmarked on leading \textbf{Multimodal Large Language Models (MLLMs)}. Additionally, \citet{yang2024sports} explored \textbf{multimodal sports understanding} by benchmarking various \textbf{video-language models} for \textbf{sports-related tasks}.  

\subsection{Cultural Benchmaks for MLLMs and LLMs} Several previous studies have focused on developing culturally relevant VQA benchmarks, including FM-IQA \cite{gao2015you}, MCVQA \cite{gupta2020unified}, xGQA \cite{pfeiffer2021xgqa}, MaXM \cite{changpinyo2022maxm}, MTVQA \cite{tang2024mtvqa}, MABL \cite{kabra2023multi}, MAPS \cite{liu2024multilingual}, and MaRVL \cite{liu2021visually}. Additionally, datasets such as CVQA \cite{romero2024cvqa}, CulturalVQA \cite{nayak2024benchmarking} and ALM-bench \cite{vayani2024all} provide VQA resources that encompass a wide range of regions and cultural themes, including food, with CVQA offering multilingual questions alongside English translations. SEA-VQA \cite{urailertprasert2024sea} specifically benchmarks the Southeast Asian region, while FoodieQA \cite{li2024foodieqa}, World Wide Dishes \cite{magomere2024you} and WORLDCUISINES \cite{winata2025worldcuisines} focus exclusively on food-related benchmarks. Our work is driven by a similar objective, using traditional sports as a cultural lens; however, it distinguishes itself with a significantly larger dataset and broader coverage of languages. Recent research has assessed LLMs' sociocultural reasoning using frameworks like the World Values Survey and Hofstede’s dimensions, highlighting gaps in adapting to user-specific and non-Western cultural contexts \cite{johnson2022ghost,atari2023morality,masoud2023cultural,seth-etal-2024-dosa,li2024culture,alkhamissi2024investigating,durmus2023towards}. While synthetic personas and fine-tuning have improved cultural adaptability and performance in tasks like hate speech detection, regional language evaluations still lag behind English benchmarks \cite{dwivedi2024exploring,shen2017style}. These findings emphasize the need for robust multilingual strategies to enhance LLMs' cultural competence.

\section{Construction of {\em CultSportQA}}
This section outlines the various stages of the creation of the benchmark \textbf{\textit{CultSportQA}}.
\begin{figure*}
\includegraphics[height = 6 cm,width=1\textwidth]{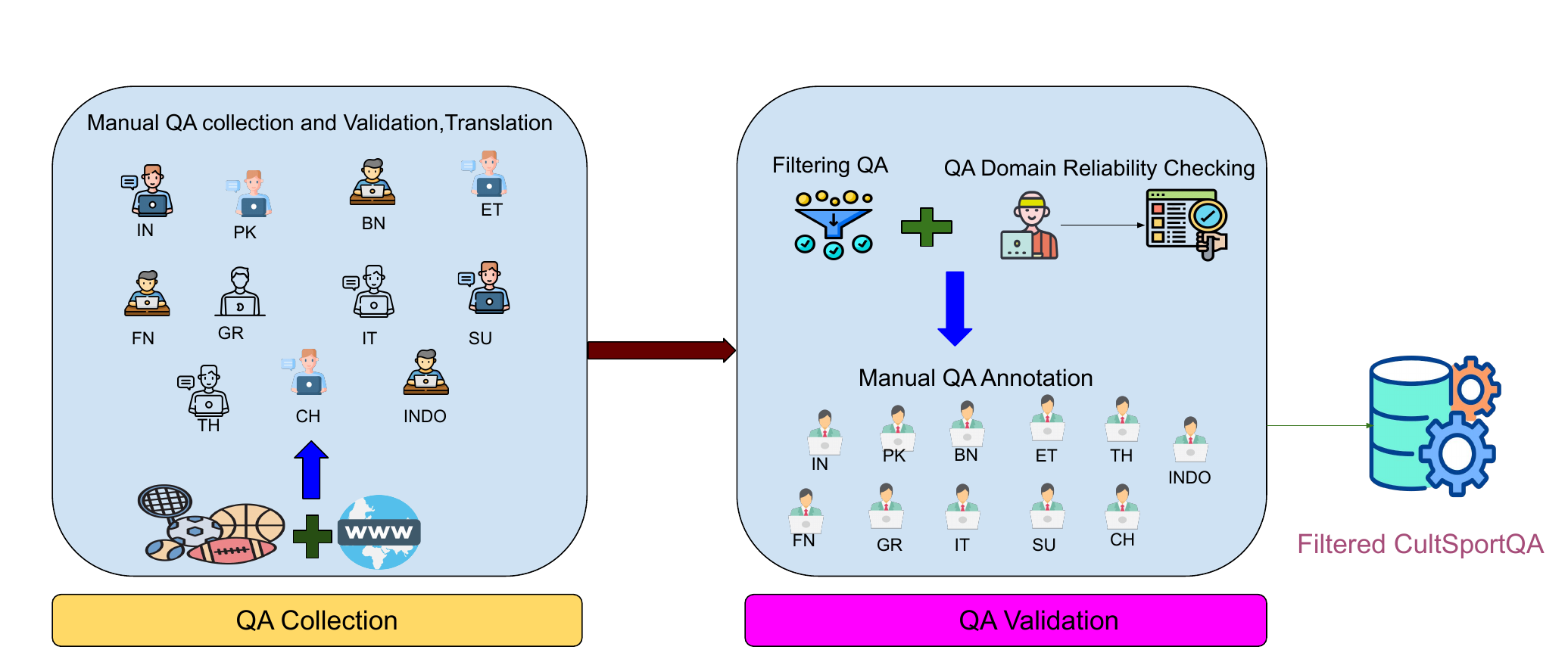}
\caption{\textbf{\textit{CultSportQA}} Manual Data Collection Pipeline: The data collection process involved two key stages. In the first stage, annotators gathered data sources and generated questions, drawing from their respective cultural backgrounds and languages. In the second stage, annotators reviewed and verified the questions to ensure cultural authenticity and maintain high translation quality.} 
\label{CultSportQA_sport}
\end{figure*}
\subsection{Manual Data Collection}
\label{subsec:preprocessing}
The creation of the \textbf{\textit{CultSportQA}} traditional sports dataset followed a carefully structured, multi-phase process to ensure comprehensive coverage and high-quality standards. Domain experts and country-specific annotators contributed at every stage, from data collection to question formulation and manual translation across multiple languages, incorporating their cultural knowledge and expertise.\par

\textbf{Data Sources:}~The dataset was curated using information from six carefully selected and credible sources to ensure comprehensive coverage of traditional sports across India, Pakistan, Bangladesh, Italy, France, China, Thailand, Indonesia, Sudan, Ethiopia, and Germany. These sources include Wikipedia, National Heritage and Sports Boards, Local Sports Blogs, Cultural Journals, News Outlets, and Academic Publications. Wikipedia served as a foundational resource for historical and rule-based information, while academic publications added scholarly depth with technical analyses. National heritage and sports boards contributed authentic cultural context and historical relevance. Cultural journals offered insights into the societal impact and evolution of these sports. Local blogs provided region-specific practices and community perspectives, and news outlets highlighted current events and preservation efforts. The resulting questions span historical facts, gameplay rules, scenario-based reasoning, and image-based understanding, capturing both the depth and diversity of traditional sports.

\textbf{Annotators Background:}~The \textbf{\textit{CultSportQA}} dataset was created with contributions from native speakers and cultural experts from 11 countries covering 11 languages across three continents. The annotators were selected such that they are fluent in the language of their respective country with most having over 10 years of residency in their respective regions. They were required to be fluent in their local languages and be aware of their cultural nuances. Contributors who provided significant input, such as validated question-answer pairs, were credited as co-authors. The team followed detailed guidelines and underwent training to ensure the questions reflected cultural relevance and diversity. Additionally, a peer-validation process ensured the accuracy and consistency of the annotations, resulting in a culturally rich and multilingual VQA benchmark. \textit{The detailed guidelines are discussed by showing an example in the Appendix.}

\textbf{Dataset Organization:}~The \textbf{\textit{CultSportQA}} dataset is organized into four question types: history-based, rule-based, scenario-based, and image-based. The dataset follows a multiple-choice question (MCQ) format with four options (A, B, C, D), where one is correct. Each question-answer (QA) pair includes metadata such as continent, country, sport name, and question type. The questions are divided into text-based, evaluated using Large Language Models (LLMs) and Small Language Models (SLMs), and image-based, assessed by Multimodal Large Language Models (MLLMs). \textbf{History-based} questions test the model’s knowledge of a sport’s origins and cultural significance. \textbf{Scenario-based} questions assess the model’s ability to determine the best move in a game situation to score maximum points. \textbf{Rule-based} questions evaluate the model’s understanding of the fundamental rules of the sport depicted in the text or image.

\begin{figure*}[htbp]
    \centering
    \begin{subfigure}{0.23\textwidth}
        \centering
        \includegraphics[width=\linewidth]{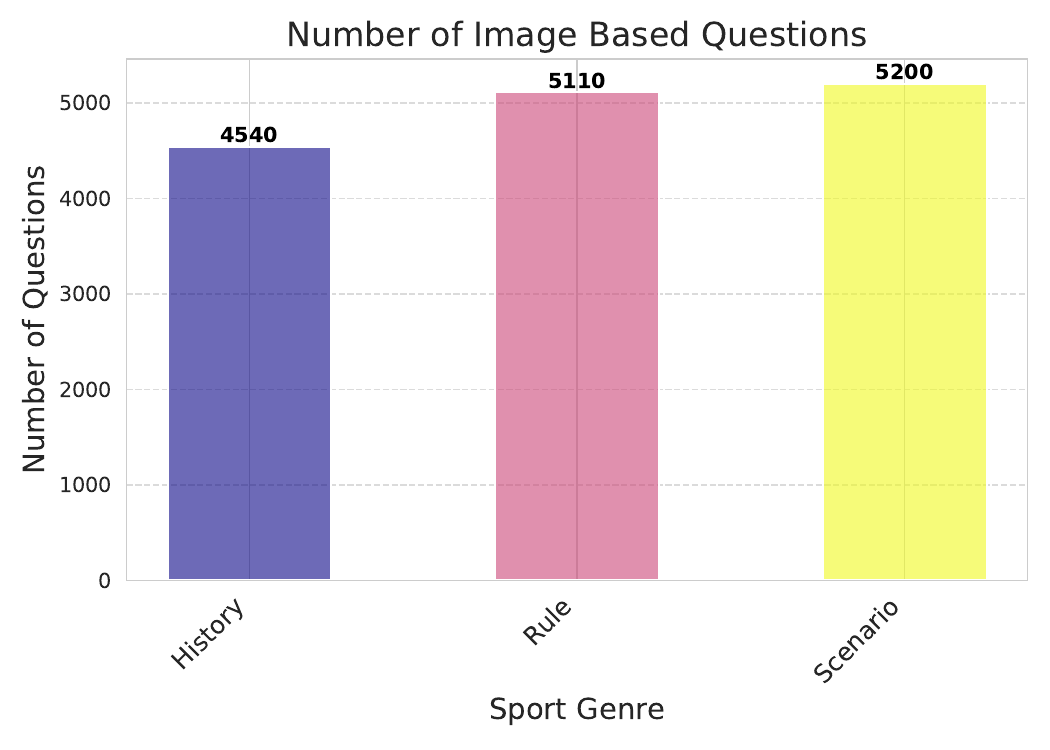}
        \caption{Image-based questions}
        \label{fig:total_image}
    \end{subfigure}
    \hfill
    \begin{subfigure}{0.23\textwidth}
        \centering
        \includegraphics[width=\linewidth]{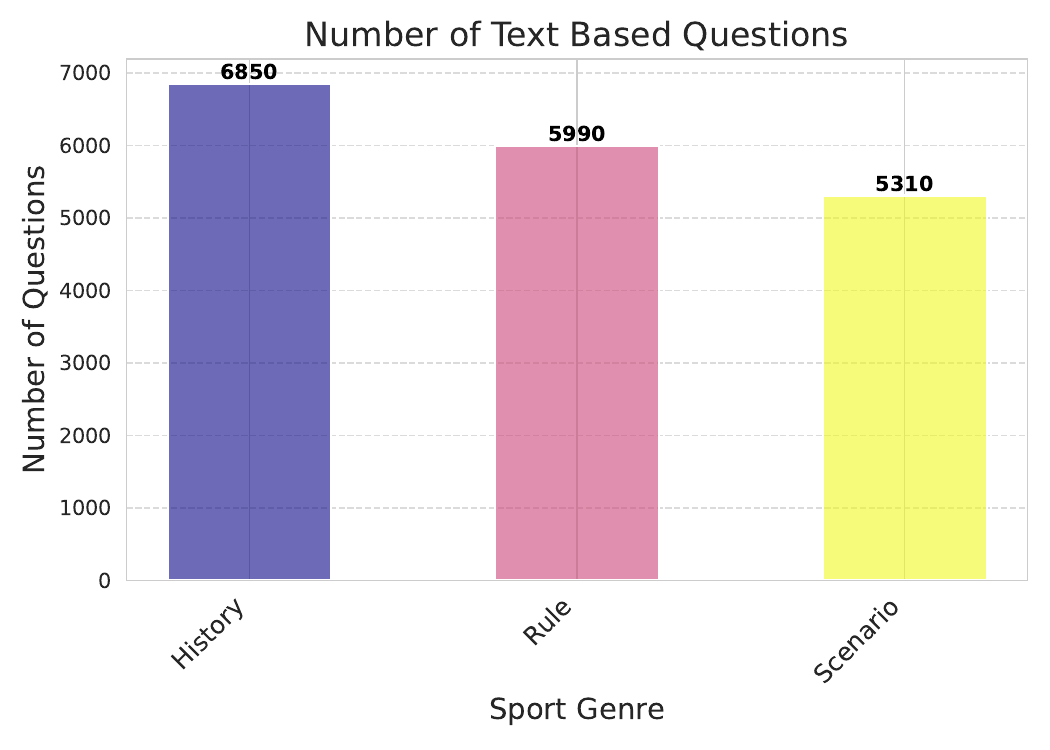}
        \caption{Text-based questions}
        \label{fig:total_text}
    \end{subfigure}
    \hfill
    \begin{subfigure}{0.23\textwidth}
        \centering
        \includegraphics[width=\linewidth]{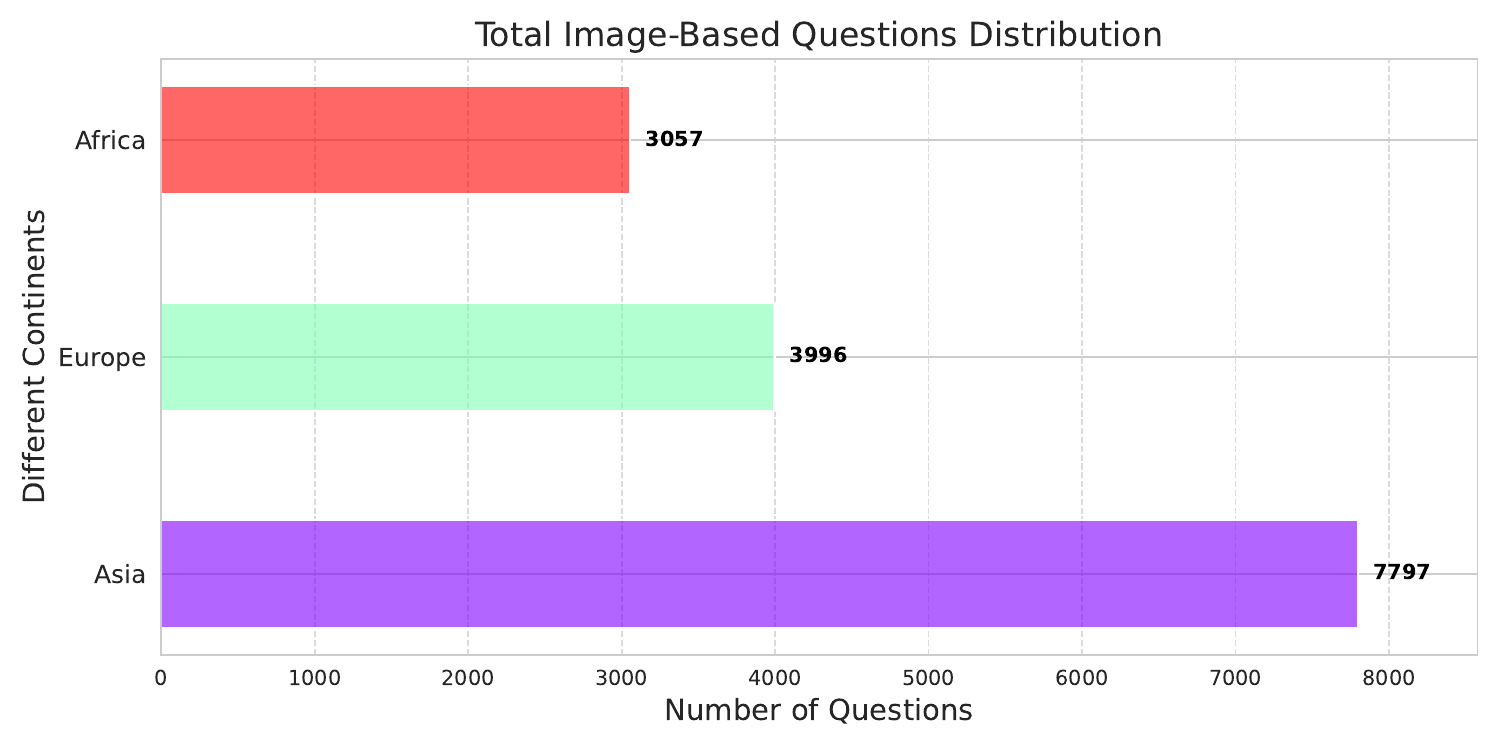}
        \caption{Image-based distribution by continents}
        \label{fig:continent_image}
    \end{subfigure}
    \hfill
    \begin{subfigure}{0.23\textwidth}
        \centering
        \includegraphics[width=\linewidth]{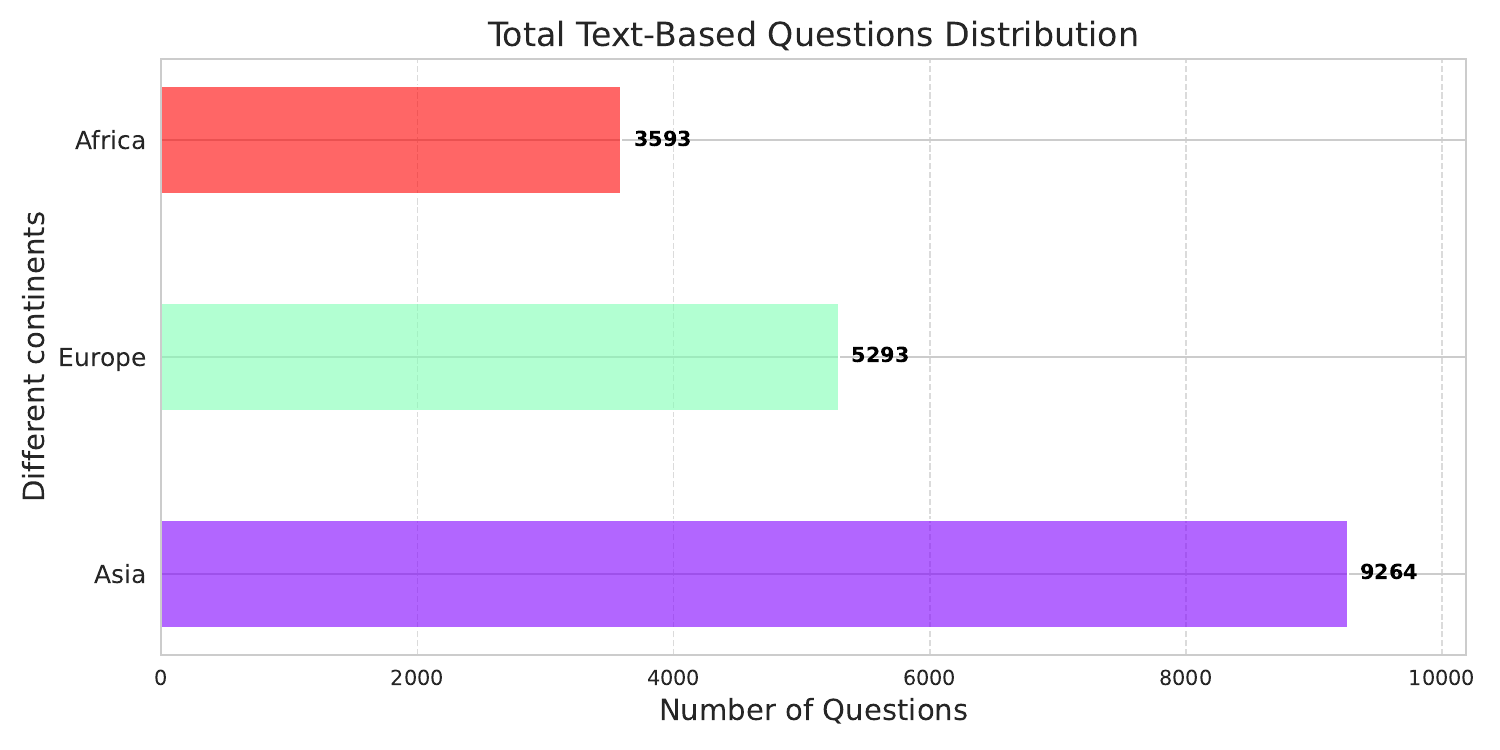}
        \caption{Text-based distribution by continents}
        \label{fig:continent_text}
    \end{subfigure}

    \caption{ Distribution of image-based and text-based questions across different question types and continents.}
    \label{fig:combined_statistics}
\end{figure*}

\subsection{Annotation Process}
We outline four main steps in annotation below:

\textbf{1. Team Structure.}
 The annotation team consisted of experienced experts with deep cultural knowledge and fluency in their respective countries' languages, ensuring both linguistic clarity and cultural authenticity. To promote diversity and thoroughness, we aimed to hire at least three annotators from each country. Within each team, two-thirds of the annotators were responsible for creating questions based on provided guidelines, leveraging their knowledge of traditional sports. The remaining annotator was tasked with validating and filtering out questions that failed to meet quality standards.\par

\textbf{2. Question Formation.}
For each selected textual passage or image, the annotator’s first task was to verify whether the content aligned with the cultural sport associated with the country. Content unrelated to the regional sport was immediately rejected. If the content was relevant, the annotator created questions focusing on rules, history, and scenarios. Each question had to be complete, self-contained, and understandable without additional context. The questions followed a multiple-choice format, consisting of four options, with only one correct answer. The final annotated format included the source passage, a relationship attribute indicating the question’s context, the type of question (e.g., history-based, rule-based, or scenario-based), and the four answer options. After a question is constructed it is translated to the regional language that the annotator is familiar with.\footnote{The annotators were paid at the rate between 0.10 dollar to 0.50 dollar per example, depending on country exchange rate and difficulty of annotation.}

\textbf{3. Training and Guidelines.}
Annotators underwent comprehensive training that included objectives of the \textbf{\textit{CultSportQA}} dataset, definitions and examples of question types, and best practices for maintaining consistency and cultural sensitivity. Detailed guideline documents provided templates, metadata tagging standards, and examples of appropriate cultural representation. Additional language and cultural training sessions emphasized the use of local terminologies and traditions. \textit{The complete process has been shown in the Appendix.} 

\textbf{4. Quality Assurance and Cross-Validation.}
A rigorous quality assurance process was conducted through multi-step validations. Each question-answer pair underwent cross-validation by at least one annotator, who understood the basic requirement to be qualified to be included in the dataset, and also the translation quality. Image-based questions were reviewed for proper alignment between visual elements and textual prompts. Spot checks and random sampling were performed by quality analysts to maintain clarity and consistency. Bias mitigation measures ensured a balanced representation of sports across regions and question types, with cultural sensitivity reviews eliminating stereotypes or offensive content. \textit{The complete validation guidelines are added in the Appendix}.

\section{Statistical Analysis of \textit{CultSportQA}}
The \textbf{\textit{CultSportQA}} dataset shows a balanced mix of text-based and image-based questions, with a slight dominance of text-based questions comprising 18,150 questions over the visual ones comprising 14,850. Image-based questions focus more on sports scenarios, testing practical understanding, while text-based questions lean toward sports history, highlighting knowledge recall. Regionally, Asia has the highest question count in both types, reflecting a strong focus on Asian sports themes. Figure-\ref{fig:combined_statistics} shows the distribution of text- and image-based questions across question categories and continents. \textit{More statistical analysis is shown in the Appendix.}

\begin{table*}[]
\renewcommand{\arraystretch}{1.5} 
\scalebox{0.6}{
\begin{tabular}{|l|l|l|l|l|l|l|}
\hline
Dataset                                  & Number of Samples             & Number of Sports          & Cultural Aspects   & Number of Language        & Modalities                            & Type Of Questions                                                             \\ \hline

SportQA \cite{xia2024sportqa}            & 70,000                        & 36                        & No                 & 1                         & Text                                  & \begin{tabular}[c]{@{}l@{}}MCQ\end{tabular} \\ \hline
SPORTU \cite{xia2024sportu}              & 12,948                        & 7                         & No                 & 1                         & Text+video                           & MCQ                                                \\ \hline
BoolQ \cite{clark2019boolq}              & 15,942                        & Not specified             & No                 & 1                         & No                                   & YES/NO                                                          \\ \hline
Sports-QA \cite{li2024sports}            & \textbf{94,000}               & 4                         & No                 & 1                         & No                                   & Descriptive                                                      \\ \hline

\textbf{\textit{CultSportQA (ours)}}     & 33,000                        & \textbf{84}               & \textbf{Yes}       & \textbf{11}               & Both Text and Text+Image              & \textbf{MCQ}                                                       \\ \hline
 
\end{tabular}
}
\caption{Comparison of our dataset with other Sports datasets. The metadata
includes Number of Samples (number of questions), Number of Sports, Cultural Aspects  (whether the data considers cultural nuances), Number of Languages, and Modalities 
(whether the data includes multimodal questions), and Type Of Questions.}
\end{table*}

\begin{figure*}[hbt!]
    \centering
    \begin{subfigure}[b]{0.32\textwidth}
        \centering
        \includegraphics[height=5.2cm, width=5.5cm]{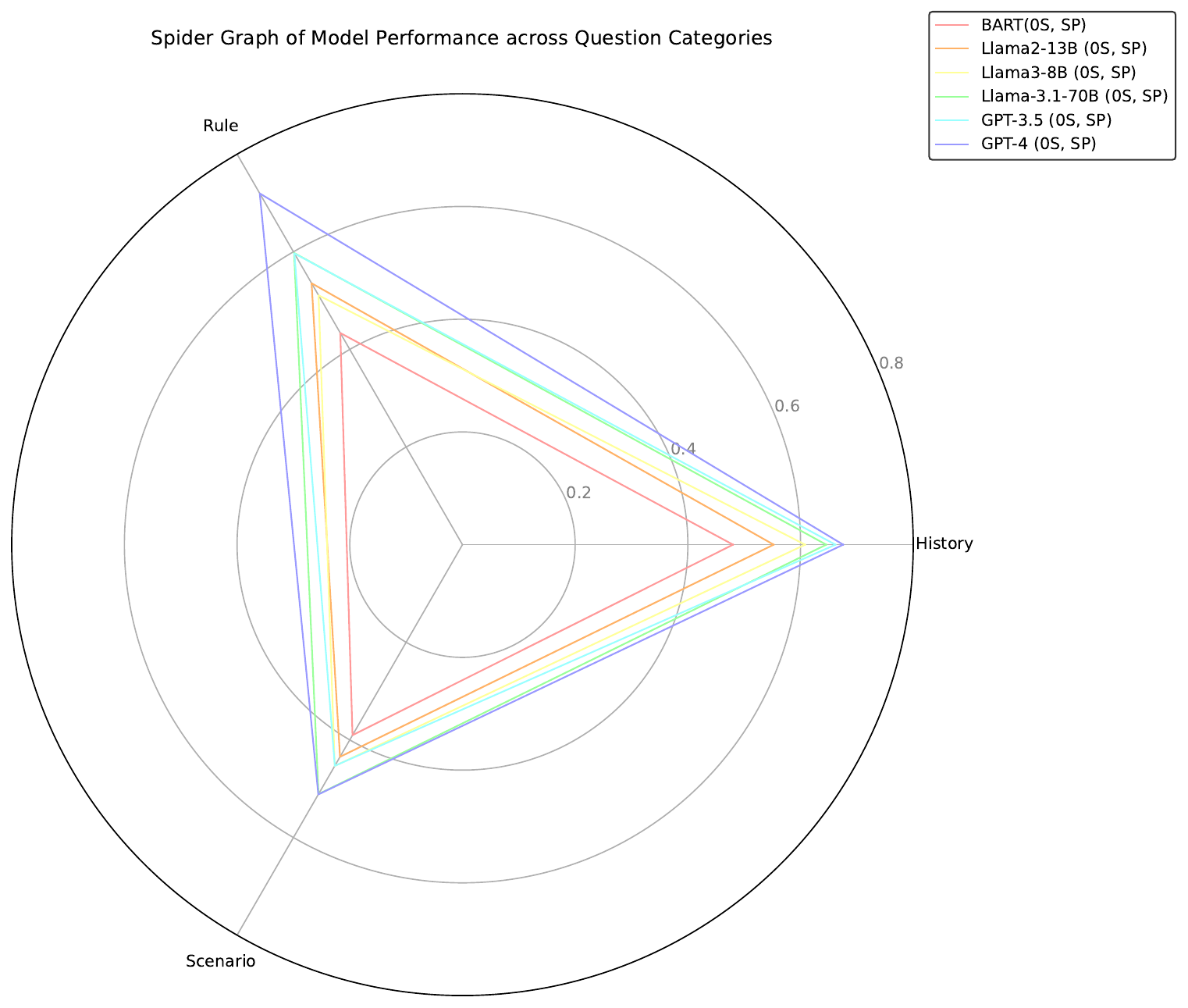}
        \caption{Zero-shot results of LLMs}
        \label{fig:czero_llm}
    \end{subfigure}
    \hfill
    \begin{subfigure}[b]{0.32\textwidth}
        \centering
        \includegraphics[height=5.2cm, width=5.5cm]{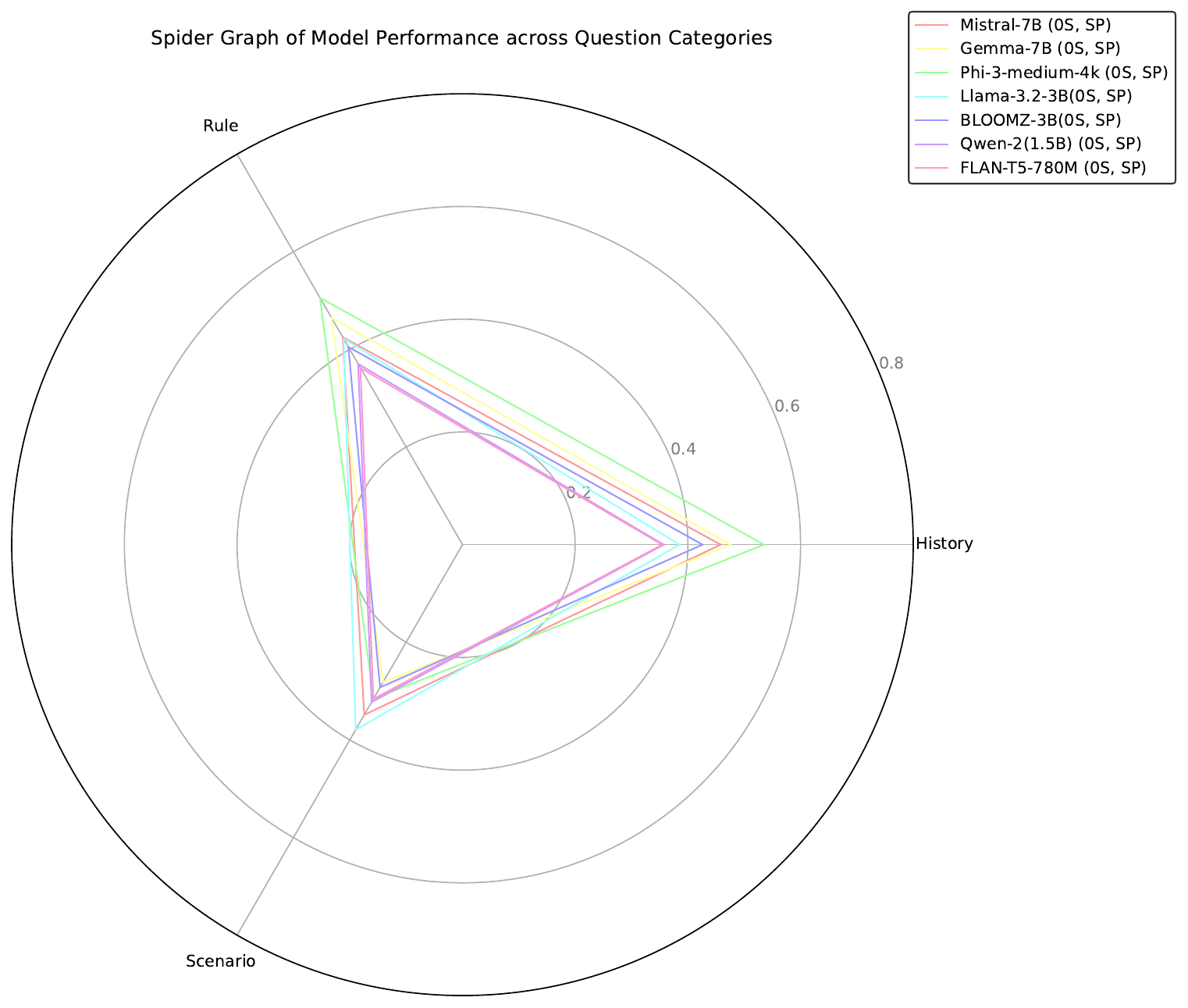}
        \caption{Zero-shot results of SLMs}
        \label{fig:category_slm}
    \end{subfigure}
    \hfill
    \begin{subfigure}[b]{0.32\textwidth}
        \centering
        \includegraphics[height=5.2cm, width=5.5cm]{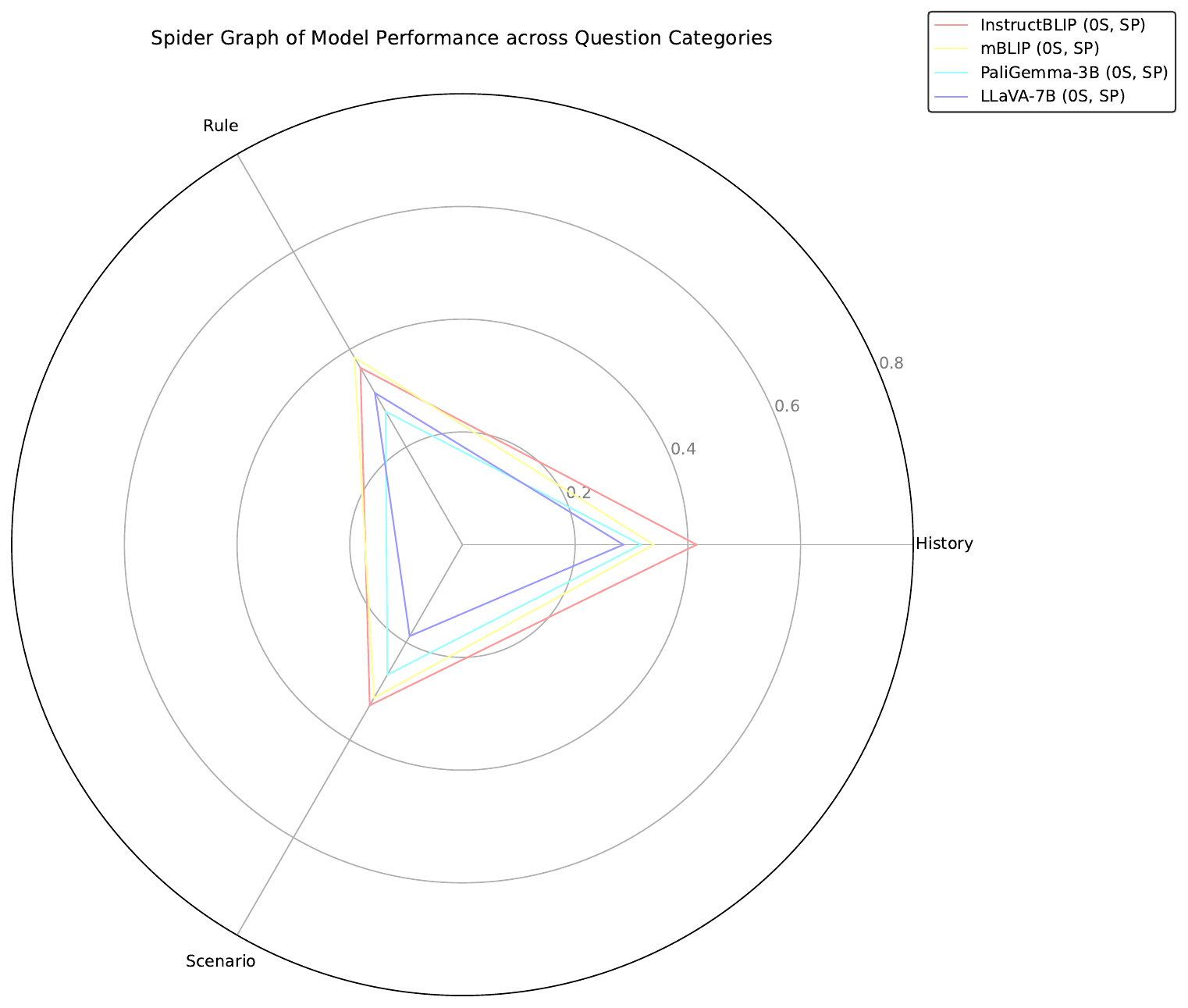}
        \caption{Zero-shot results of MLLMs}
        \label{fig:category_zero_VLM}
    \end{subfigure}
    
    \caption{Zero-shot results of language models across different question types.}
    \label{fig:sanskriti1_results}
\end{figure*}

\vspace{-0.3cm}

\begin{figure*}[hbt!]
    \centering
    \begin{subfigure}[b]{0.32\textwidth}
        \centering
        \includegraphics[width=\textwidth]{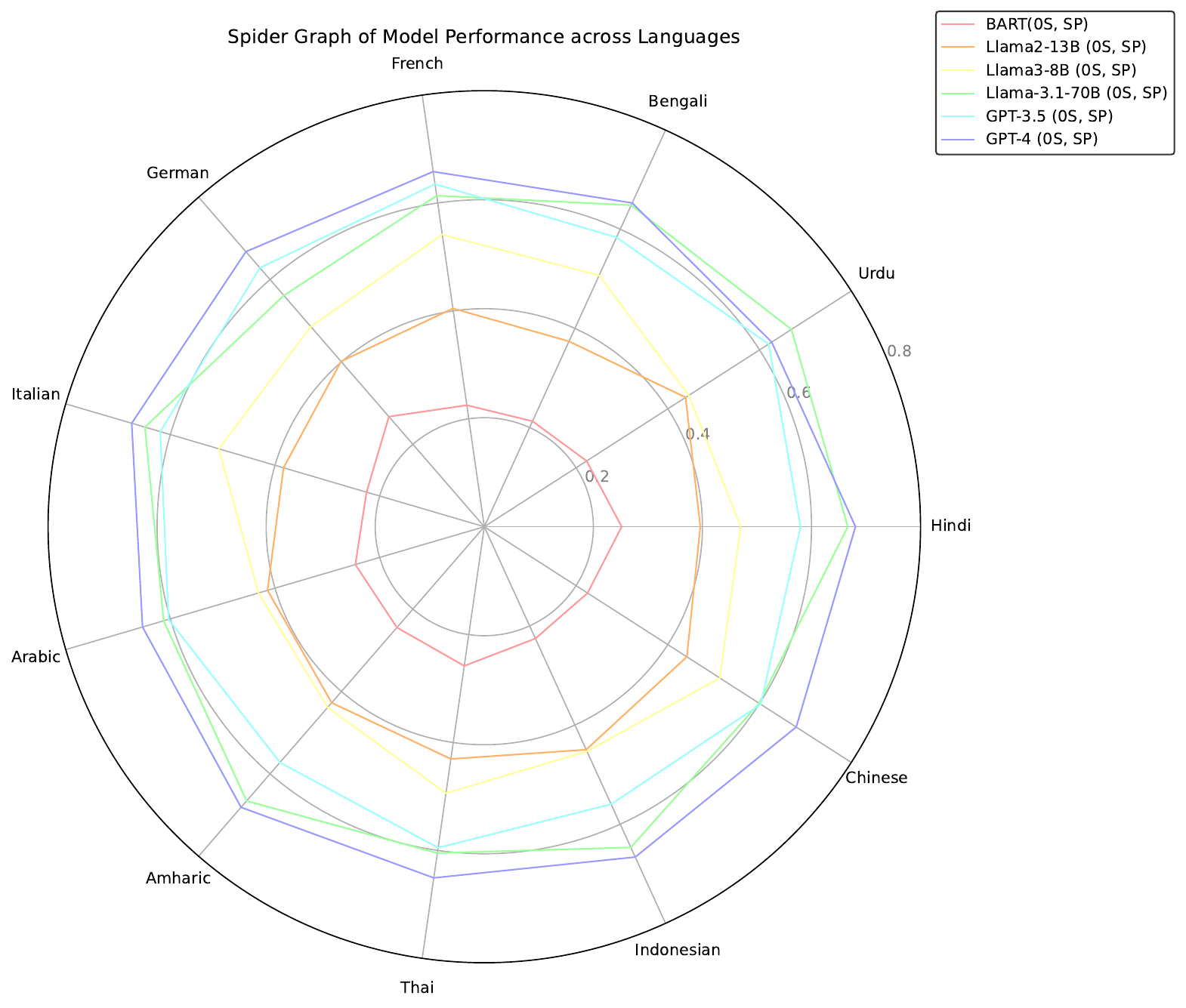}
        \caption{Performance of LLMs across languages}
        \label{fig:lzero_llm}
    \end{subfigure}
    \hfill
    \begin{subfigure}[b]{0.32\textwidth}
        \centering
        \includegraphics[width=\textwidth]{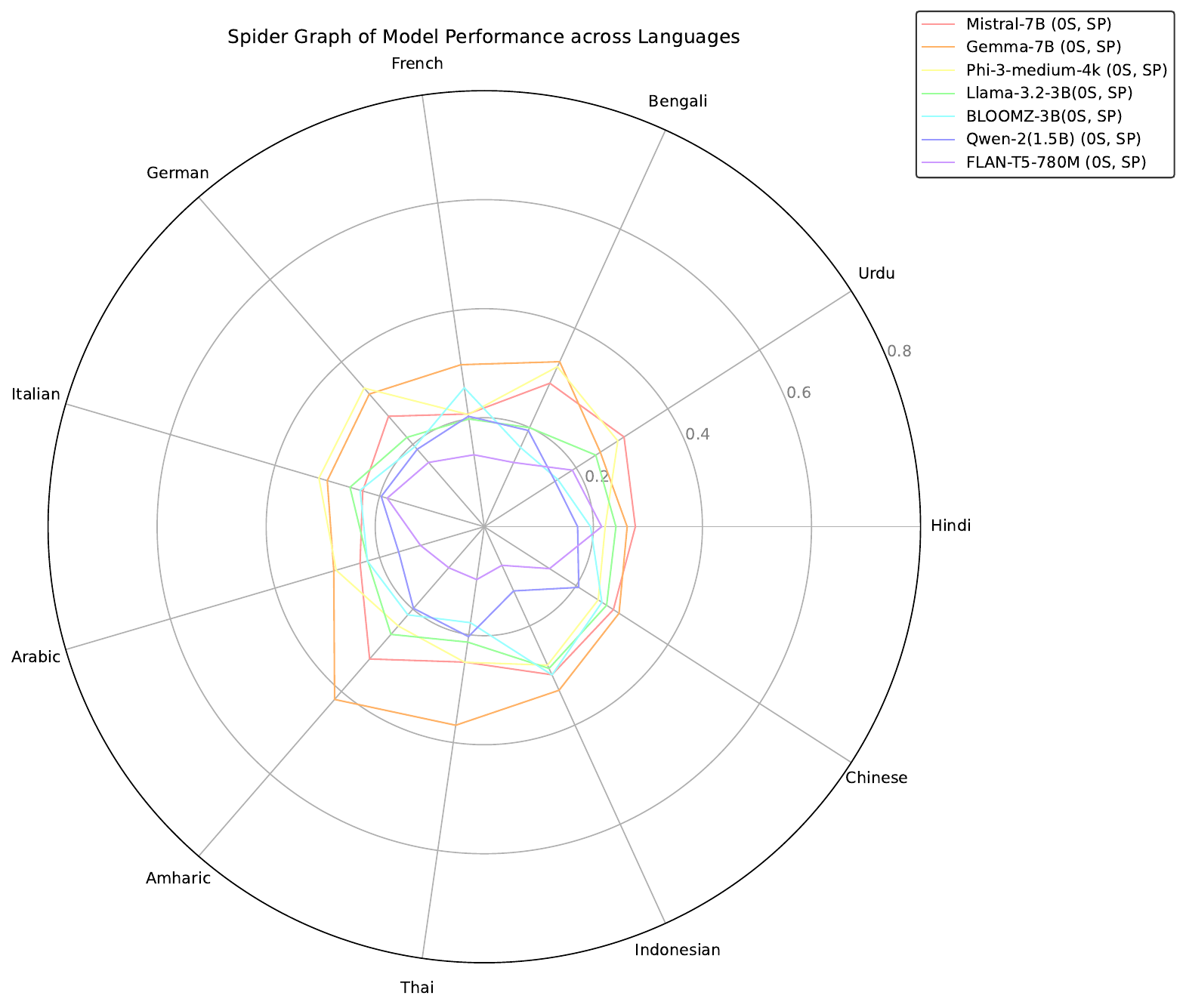}
        \caption{Performance of SLMs across languages}
        \label{fig:Lang_zero_SLM}
    \end{subfigure}
    \hfill
    \begin{subfigure}[b]{0.32\textwidth}
        \centering
        \includegraphics[width=\textwidth]{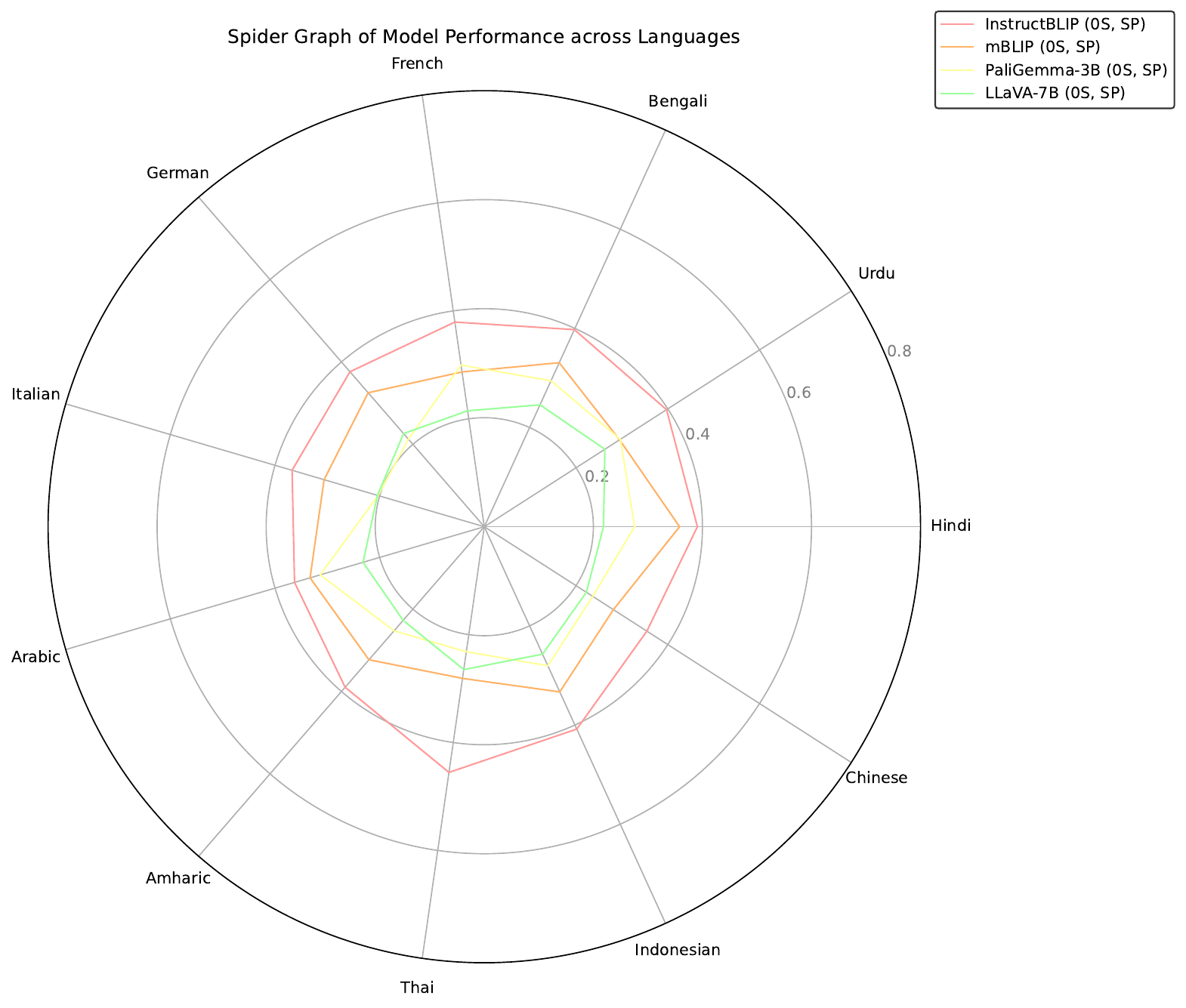}
        \caption{Performance of MLLMs across languages}
        \label{fig:Lang_ZERO_VLM}
    \end{subfigure}
    
    \caption{Average results of language models on the \textit{\textbf{CultSportQA}} dataset classified on the basis of languages.}
    \label{fig:Lang_ZERO_results}
\end{figure*}

\section{Experimental Setup}
\subsection{Models}
To conduct a comprehensive evaluation of our benchmark, \textbf{\textit{CultSportQA}}, we carried out an extensive assessment across a diverse range of language models across different modalities of text and image. For text, our evaluation encompassed leading LLMs, including Llama2-13B \cite{dubey2024llama}, Llama3-8B \cite{dubey2024llama}, Llama-3.1-70B-Instruct \cite{dubey2024llama}, and GPT-3.5. Additionally, we tested several SLMs such as Mistral-7B \cite{jiang2023mistral}, Gemma-7B \cite{team2024gemma}, Phi-3-medium-4k \cite{abdin2024phi}, Llama-3.2-3B \cite{dubey2024llama} \cite{touvron2023llama}, BLOOMZ-3B \cite{muennighoff2022crosslingual} Qwen-2 (1.5B) \cite{bai2023qwen}, and FLAN-T5-780M \cite{chung2024scaling} and BART \cite{lewis2019bart}.
Beyond text-based models, we also evaluated a range of MLLMs to test the reasoning of sports across multilingual settings. In this category, we assessed InstructBLIP \cite{panagopoulou2023x} and mBLIP—a BLIP-2-based model \cite{geigle2023mblip} supporting 96 languages—where we tested two variations: PaliGemma-3B \cite{beyer2024paligemma} and LLaVA-7B \cite{liu2023visual}. Finally, for a more holistic comparison, we incorporated the proprietary model GPT-4o into our evaluation.

\begin{table*}[htp]
\renewcommand{\arraystretch}{1.5} 
\scalebox{0.50}{
\begin{tabular}{|l|l|l|l|l|l|l|l|l|l|l|l|l|l|}
\hline
Model     & BART  & Llama2-13B & Llama3-8B & Llama-3.1-70B & GPT-3.5 & GPT-4o & Mistral-7B & Gemma-7B & Phi-3-medium & Llama-3.2-3B & BLOOMZ-3B & Qwen-2(1.5B) & FLAN-T5-780M \\ \hline
Zero-shot & 24.24 & 41.47      & 48.09     & 62.07         & 59.99   &  \textbf{66.29} & 23.71      & 30.56    & 31.45        & 24.67        & 21.67     & 18.60        & 15.45        \\ \hline
Few-shot   & 27.24 & 43.72      & 51.87     & 64.54         & 63.93   & \textbf{69.53} & 33.73      & 33.78    & 34.67        & 27.08        & 25.11     & 21.98        & 18.90        \\ \hline
CoT        & 31.93 & 46.24      & 54.90     & 69.34         & 67.18   & \textbf{74.51} & 40.98      & 38.98    & 39.56        & 30.23        & 28.79     & 26.29        & 23.76        \\ \hline
\end{tabular}
}
\caption{Performance comparison of various LLMs and SLMs in the text-based questions of \textbf{\textit{CultSportQA}}}
\label{tab:llm_task_comparison}
\end{table*}

\begin{table}[htp]
\scalebox{0.67}{
\begin{tabular}{|l|l|l|l|l|}
\hline
Model     & InstructBLIP & mBLIP & PaliGemma-3B & LLaVA-7B \\ \hline
Zero-shot & \textbf{38.90}        & 32.90 & 27.33        & 24.95    \\ \hline
Few-shot   & \textbf{44.83}        & 36.22 & 31.67        & 29.02    \\ \hline
 CoT   & \textbf{49.45}        & 40.85 & 37.90        & 35.37    \\ \hline
\end{tabular}

}
\caption{Performance comparison of various MLLMs in the image-based questions of \textbf{\textit{CultSportQA}}}
\label{tab:vlm1_task_comparison}
\end{table}

\subsection{Evaluation Setup}
We conducted a comprehensive evaluation of the \textbf{\textit{CultSportQA}} dataset, which includes text and image-based \textbf{Multiple-Choice Questions (MCQs)} grouped into three categories: \textbf{1. Cultural and Historical Knowledge}, \textbf{2. Rule Comprehension}, and \textbf{3. Scenario-Based Reasoning}. To assess model performance across languages and modalities, we used three prompting techniques: \textbf{zero-shot}, \textbf{few-shot (3-shot)}, and \textbf{Chain of Thought (CoT)}, with the \textbf{temperature parameter set to 0} for consistency. \textbf{Accuracy} was the sole evaluation metric. \textbf{Open-source models} used \textbf{16-bit floating-point precision} and \textbf{greedy decoding}, while \textbf{proprietary models} were accessed via APIs. Predictions were based on the highest output probability, ensuring a standardized evaluation process.

\section{Experimental Results}

\subsection{Main Results} 
The overall performances on the \textbf{\textit{CultSportQA}} dataset across various LLMs, SLMs, and MLLMs are presented in Table-\ref{tab:llm_task_comparison} and  Table-\ref{tab:vlm1_task_comparison}\par
 \textbf{Performance of LLMs:}  
LLMs consistently outperformed other model types, with GPT-4o .87 and Llama-3.1-70B .84 leading across all evaluation categories. These models demonstrated superior accuracy in language-based tasks, question-type performance, and continent-based variations. GPT-3.5 .81, while slightly behind the top two, remained highly competitive, particularly excelling in history-based (83.1\%) and rule-based (82.7\%) questions. 

The performance comparison shows \textbf{GPT-4o} as the strongest model, excelling in few-shot setting due to its superior ability in-context learning. GPT-3.5 performs well but lags behind GPT-4o, especially with more shots. Among LLaMA models, \textbf{LLaMA-3.1-70B} leads, benefiting from its larger size,  LLaMA2-13B performs moderately, while Pretrained Language Models (PLMs) \textbf{BART} ranks lowest, highlighting its limited capacity. Overall, larger models and few-shot learning drive the highest performance.

\textbf{Performance of SLMs:}
Among SLMs, \textbf{Mistral-7B} achieves the highest performance, particularly excelling with few-shot learning, highlighting its strong generalization capabilities. \textbf{Gemma-7B} and \textbf{Phi-3-medium} follow closely, showing competitive results with steady improvement across settings. \textbf{LLaMA-3.2-3B} performs moderately, benefiting from few-shot examples but trailing behind larger models. \textbf{BLOOMZ-3B, Qwen-2(1.5B), and FLAN-T5-780M} exhibit the lowest performance, with FLAN-T5 performing weakest, reflecting its limited capacity for complex reasoning. Overall, performance scales with model size and improves with few-shot learning, with Mistral and Gemma emerging as the strongest contenders among SLMs.

 \textbf{Performance of MLLMs:}
The performance comparison of MLLMs shows that \textbf{InstructBLIP} outperforms all other models, demonstrating strong reasoning and adaptability, especially in the fewshot setting. \textbf{mBLIP} performs well but trails behind \textbf{InstructBLIP}, indicating slightly weaker multimodal integration. \textbf{PaliGemma-3B} and \textbf{LLaVA-7B} show lower performance, with \textbf{LLaVA-7B} performing the weakest, highlighting its limitations in complex tasks. All models benefit from few-shot learning, with \textbf{InstructBLIP} showing the greatest improvement, underscoring its superior in-context learning capability.
 
\textbf{Performance across Languages:}
In the case of LLMs, GPT-4o leads in performance across all languages, closely followed by GPT-3.5, while LLaMA-3B (70B) outperforms LLaMA-2-13B in most languages; BART performs the weakest, showing significant gaps, especially in non-Latin scripts like Amharic and Thai. For SLMs, Mistral-7B and Gemma-7B lead across most languages, with Mistral-7B excelling particularly in Arabic and Italian; Phi-3-medium and LLaMA-3B show moderate performance, while BLOOMZ-3B, Qwen-2.5B, and FLAN-T5-780M lag, especially in non-Latin languages like Amharic and Thai. And finally, concerning MLLMs, InstructBLIP leads overall, excelling in Hindi, Chinese, and Arabic, while mBLIP performs well but falls behind in Urdu and German; PoliGemma-3B shows moderate performance, outperforming LLaVA-7B, which struggles across most languages.

\begin{figure*}[htp]
\includegraphics[height = 6.8 cm,width=1\textwidth]{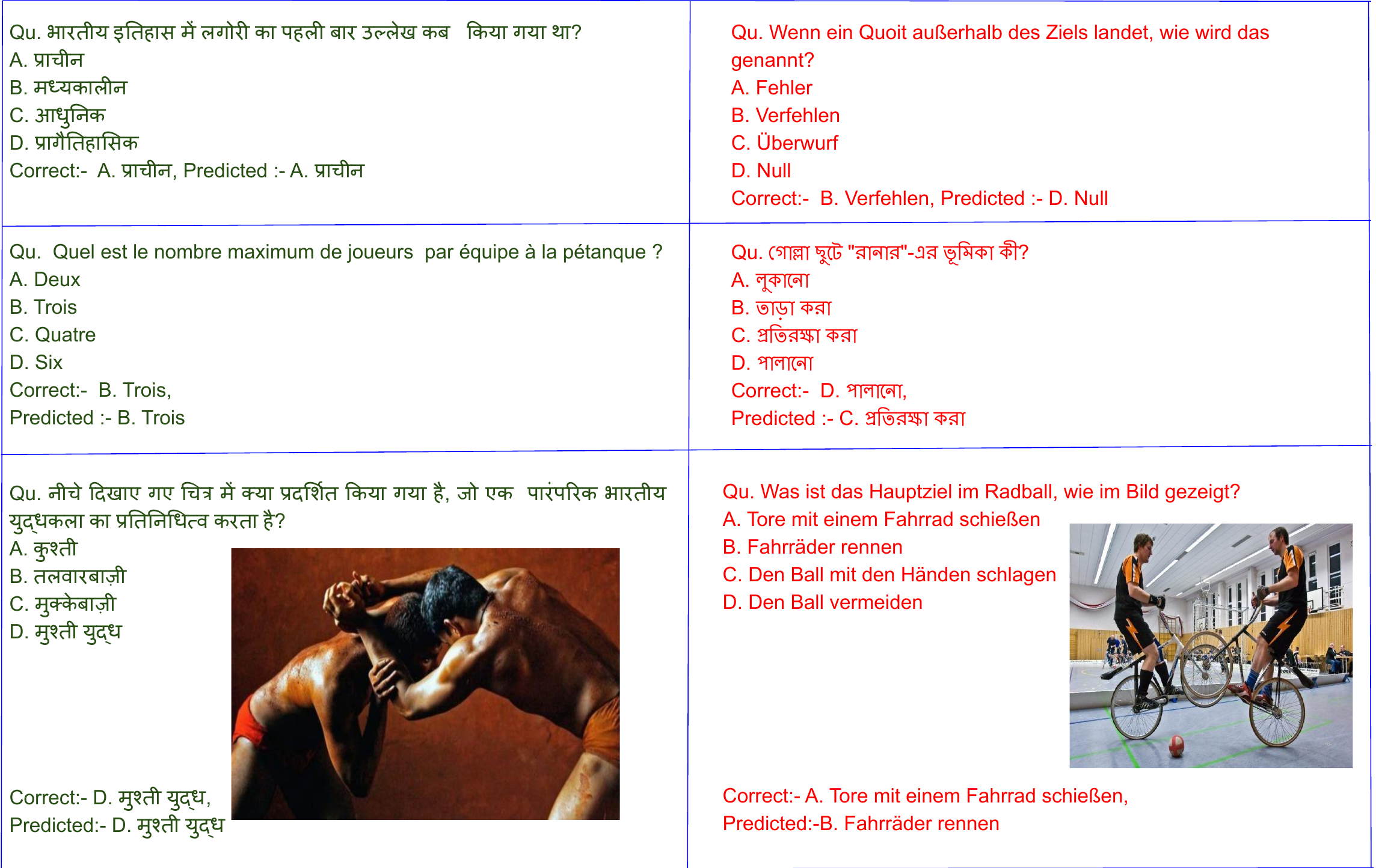}
\caption{The LHS displays the correctly answered questions, while the RHS highlights the incorrectly answered ones by the language models on the {\textbf{\textit{CultSportQA}} dataset. } }
\label{Error}
\end{figure*}
 
\textbf{Performance across  Question types:} Figure-\ref{fig:sanskriti1_results} analyzes the performance of LLMs, SLMs, and MLLMs in a zero-shot setup across various kinds of questions related to history, rule-based, and scenario-based of \textbf{\textit{CultSportQA}} dataset. With respect to LLMs, GPT-4o leads across all categories, with GPT-3.5 closely following, while LLaMA models show moderate performance, with the 70B variant surpassing other versions; BART performs the lowest across all categories. Among SLMs, Mistral-7B leads across all categories, especially in History and Rule, with Gemma-7B and LLaMA-3B showing competitive performance, while BLOOMZ-3B and FLAN-T5-780M lag behind, particularly in Scenario-based questions. Among MLLMs, InstructBLIP outperforms all models, particularly excelling in History and Scenario categories, while mBLIP and PoliGemma-3B show moderate performance across all categories, with LLaVA-7B trailing, especially in Scenario-based questions. \textit{Appendix contains the results of language models in  COT and few-shot across languages.}

\subsection{Error Analysis}
To evaluate the strengths and limitations of the best-performing models on the \textbf{\textit{CultSportQA}} dataset, we conducted an error analysis, grouping questions into correctly and incorrectly answered sets, as shown in Figure \ref{Error}. The analysis highlights key patterns of success and failure. On the left (LHS), correct predictions stem from strong keyword associations (e.g., “Lagori” and “Petanque”) and well-structured questions with distinct answer choices. On the right (RHS), errors are driven by limited knowledge of culturally nuanced sports (e.g., “Quoits”) and confusion caused by ambiguous or overlapping answer options (e.g., “Game of Skill” vs. “Strategic Team Game”). These issues point to gaps in cultural coverage and underscore the need for more diverse training data to improve model performance on sports-related queries.

\section{Conclusion}
In this work, we introduced \textbf{\textit{CultSportQA}}, a comprehensive benchmark designed to evaluate language models’ understanding of Asian, African, and European traditional sports. The dataset, consisting of 33,000 curated question-answer pairs from 11 countries, covers key aspects such as rules, cultural significance, and historical context. Evaluations with leading models revealed notable gaps in answering traditional sport-specific questions, highlighting biases likely caused by training data limitations. \textbf{\textit{CultSportQA}}, built for quality and cultural sensitivity, advances inclusive AI research. Future expansions will add more languages and traditional sports to enhance its impact.

\section{Limitations}
While this study represents one of the most comprehensive evaluations of language models in the context of traditional sports and cultural knowledge, several notable limitations must be acknowledged:

\textbf{(1) Limited Geographic Scope:}
The dataset and analysis are focused solely on the regional sports of 11 countries spanning across 3 continents. While these regions provide valuable insights, the dataset can still be extended to other countries across different continents. In the future, we will expand the dataset to include regional sports from additional countries, which could offer a broader understanding and uncover more diverse trends.

\textbf{(2) Limited Representation of Traditional Sports:}
Although the study covers 84 traditional sports (46 from Asia, 25 from Europe, and 13 from Africa), which is the largest sports cultural data set, the data set may not fully represent the rich tapestry of traditional sports across these continents. Future iterations could expand the dataset to include a wider range of sports and introduce diverse question-answering tasks, such as True/False questions, adversarial questions, and scenario-based reasoning.

\textbf{(3) Limited Language and Cultural Coverage:}
The \textbf{\textit{CultSportQA}} dataset spans 11 languages from 11 different countries, providing a valuable initial benchmark for evaluating language models. However, expanding the dataset to include more low-resource languages would enhance its diversity and inclusivity. Such an expansion would not only promote traditional sports at the grassroots level but also enable more comprehensive assessments of language models across diverse linguistic and cultural contexts.

\textbf{(4) Limited scope  of  Modalities:}
The dataset includes only text and image modalities, lacking other potential modalities. It specifically focuses on multimodal combinations that require reasoning across multiple modalities simultaneously to answer queries effectively. The complexity involved in creating multimodal questions is high, but we remain committed to continually updating and expanding the dataset to enhance its scope and depth.
\section{Ethics Statement}
\textbf{Data Collection and Bias Mitigation:} The data used in the development of \textbf{\textit{CultSportQA}} was collected from publicly accessible platforms, as outlined in Section \ref{subsec:preprocessing}. These platforms were carefully selected to ensure authenticity, making \textbf{\textit{CultSportQA}} a significant milestone in establishing a standardized and inclusive benchmark for evaluating Asian, European, and African traditional sports. The dataset sources were thoroughly verified by annotators through multiple rounds of group discussions. Following the collection process, annotators curated the dataset by extracting portions suitable for question generation and discarding irrelevant metadata. To prevent language bias, the dataset comprises 3,000 data points for each of the 11 selected languages, ensuring balanced representation.

\textbf{Human Annotation:}
Human annotators were key in creating, checking, and translating questions to make sure the dataset truly reflects cultural and sports contexts. The team included 42 experts from 11 countries, with backgrounds in sports, linguistics, and related fields. Most were native or bilingual speakers with over 15 years of sports experience, aged between 30 and 50. They received training on the dataset’s goals, question types, and sports-specific guidelines. To maintain quality, a separate sub-team cross-checked the work. Throughout the process, fairness and inclusivity were emphasized, avoiding stereotypes and ensuring cultural diversity was respected.

\section*{Acknowledgments}

This work was partially funded by the Geo-R2LLM CHIST-ERA project. The experiments presented were partially conducted using the OCCIDATA platform administered by IRIT (CNRS/University of Toulouse).

\if 0\fi

\bibliography{Custom1}

\appendix

\section{Appendix}

The Appendix includes information about discussion about models, information about annotators wage annotators distribution across countries, an annotation example of a data point, prompts for evaluation, additional statistical analysis of the dataset \textbf{\textit{CultSportQA}}, further results on languages (in few and CoT questions), four categories of questions (in few-shot and CoT questions), performance across continents, and many more qualitative examples from our benchmark \textbf{CultSportQA}.

\begin{table*}[ht]
\centering
\caption{\textbf{Annotator Wages and Annotator Counts by Country}}
\begin{minipage}[t]{0.48\linewidth}
\centering
\textbf{(a) Average Wages per 100 Samples}\\[2pt]
\begin{tabular}{|l|l|l|}
\hline
\textbf{Language} & \textbf{Country} & \textbf{Wage (US\$)} \\ \hline
Hindi      & India       & 3   \\ \hline
Urdu       & Pakistan    & 2   \\ \hline
Bengali    & Bangladesh  & 1.5 \\ \hline
Thai       & Thailand    & 3   \\ \hline
Indonesian & Indonesia   & 5   \\ \hline
Chinese    & China       & 4   \\ \hline
French     & France      & 13  \\ \hline
German     & Germany     & 28  \\ \hline
Italian    & Italy       & 12  \\ \hline
Amharic    & Ethiopia    & 2   \\ \hline
Arabic     & Sudan       & 1.5 \\ \hline
\end{tabular}
\end{minipage}
\hfill
\begin{minipage}[t]{0.48\linewidth}
\centering
\textbf{(b) Number of Annotators per Country}\\[2pt]
\begin{tabular}{|l|l|l|}
\hline
\textbf{Country} & \textbf{Language} & \textbf{Annotators} \\ \hline
India       & Hindi       & 5 \\ \hline
Pakistan    & Urdu        & 5 \\ \hline
Bangladesh  & Bengali     & 4 \\ \hline
Thailand    & Thai        & 5 \\ \hline
Indonesia   & Indonesian  & 3 \\ \hline
China       & Mandarin    & 5 \\ \hline
France      & French      & 3 \\ \hline
Germany     & German      & 3 \\ \hline
Italy       & Italian     & 3 \\ \hline
Ethiopia    & Amharic     & 5 \\ \hline
Sudan       & Arabic      & 5 \\ \hline
\end{tabular}
\end{minipage}
\end{table*}

\section{Discussion about Models}

To ensure a holistic evaluation, our study includes models from diverse categories—Large Language Models (LLMs, >7B parameters), Small Language Models (SLMs, <=7B parameters), and Multimodal Large Language Models (MLLMs), the latter being essential for addressing the visual question answering component of our benchmark.

\subsection{Large Language Models (LLMs)}

\begin{itemize}
    \item \textbf{Meta's LLaMA Series}:
    \begin{itemize}
        \item \textbf{LLaMA 2 13B} and \textbf{LLaMA 3 8B}: Open-source models optimized for general-purpose language tasks.
        \item \textbf{LLaMA 3.1 70B Instruct}: Supports a 128K token context window and multilingual capabilities across eight languages, suitable for complex reasoning and enterprise applications.
    \end{itemize}
    
\end{itemize}

\subsection{Small Language Models (SLMs)}

\begin{itemize}
    \item \textbf{Mistral 7B (Mistral AI)}: Employs grouped-query and sliding window attention mechanisms, offering efficient inference and handling of longer sequences, ideal for deployment in resource-constrained environments.
    \item \textbf{Gemma 7B (Google DeepMind)}: Demonstrates strong performance in code generation and mathematical problem-solving tasks, outperforming similar-sized models in these domains.
    \item \textbf{Phi-3 Medium (Microsoft)}: With 14 billion parameters and a 128K token context window, Phi-3 Medium is designed for demanding computational tasks, offering a balance between performance and efficiency.
    \item \textbf{BART (Facebook AI)}: A denoising autoencoder combining bidirectional and autoregressive transformers, excelling in text generation and comprehension tasks such as summarization and translation.
\end{itemize}

\subsection{Multimodal Large Language Models (MLLMs)}

\begin{itemize}
    \item \textbf{InstructBLIP (Salesforce)}: An instruction-tuned vision-language model built upon BLIP-2, excelling in zero-shot performance across various multimodal tasks, including image captioning and visual question answering.
    \item \textbf{mBLIP}: A multilingual extension of BLIP, designed to handle vision-language tasks across multiple languages, enhancing cross-cultural understanding and accessibility.
    \item \textbf{LLaVA 7B}: Integrates visual and textual information, enabling tasks that require understanding and generating content from both modalities.
    \item \textbf{GPT-4o (OpenAI)}: A multimodal model capable of processing and generating text, images, and audio, offering advanced capabilities in tasks that require integrating information across different modalities.
\end{itemize}

\section{Sources for Dataset Collection}

\textbf{India:} The dataset for India was compiled from publicly accessible sources including \url{https://www.wikipedia.org}, \url{https://www.traditionalsports.org/}, \url{https://indianexpress.com/section/sports/}, \url{https://sports.ndtv.com/}, \url{https://www.cnbc.com/sport/}, and \url{https://www.traditionalsportsgames.org/}.

\textbf{Pakistan:} Relevant data for Pakistan was gathered from \url{https://www.wikipedia.org}, \url{https://www.traditionalsports.org/}, \url{https://www.cnbc.com/sport/}, \url{https://www.pakistantoday.com.pk/}, \url{https://tsgpakistan.com}, and \url{https://www.traditionalsportsgames.org/}.

\textbf{Bangladesh:} Sources for Bangladesh include \url{https://www.wikipedia.org}, \url{https://www.traditionalsports.org/}, \url{https://www.cnbc.com/sport/}, and \url{https://www.traditionalsportsgames.org/}.

\textbf{Thailand:} The Thai dataset was created using data from \url{https://www.wikipedia.org}, \url{https://www.traditionalsports.org/}, \url{https://www.cnbc.com/sport/}, and \url{https://www.traditionalsportsgames.org/}.

\textbf{Indonesia:} For Indonesia, sources include \url{https://www.wikipedia.org}, \url{https://www.traditionalsports.org/}, \url{https://www.cnbc.com/sport/}, and \url{https://www.traditionalsportsgames.org/}.

\textbf{China:} The dataset covering China was built using information from \url{https://www.wikipedia.org}, \url{https://www.traditionalsports.org/}, \url{https://www.cnbc.com/sport/}, \url{https://www.traditionalsportsgames.org/}, and \url{https://chcp.org/Games}.

\textbf{France:} Data for France was collected from \url{https://www.wikipedia.org} and \url{https://www.traditionalsports.org/}.

\textbf{Germany:} The German dataset is based on content from \url{https://www.wikipedia.org} and \url{https://www.traditionalsports.org/}.

\textbf{Italy:} Italy’s dataset draws from \url{https://www.wikipedia.org} and \url{https://www.traditionalsports.org/}.

\textbf{Ethiopia:} Relevant materials for Ethiopia were obtained from \url{https://www.wikipedia.org} and \url{https://www.traditionalsports.org/}.

\textbf{Sudan:} For Sudan it was sourced from \url{https://www.wikipedia.org}



**Source text for Kho-Kho "Sports" From "India"**

खो-खो मैदानी खेलों के सबसे प्राचीनतम रूपों में से एक है जिसका उद्भव प्रागैतिहासिक भारत में माना जा सकता है। मुख्य रूप से आत्मरक्षा, आक्रमण व प्रत्याक्रमण के कौशल को विकसित करने के लिए इसकी खोज हुई थी।

खो-खो का जन्मस्थान पुण कहा जाता है। यह खेल महाराष्ट्र, गुजरात और मध्य प्रदेश आदि प्रदेशों में अधिक खेला जाता है, किंतु भारत के अन्य प्रदेशों में भी इसका प्रचार अब बढ़ रहा है। यह खेल सरल है और इसमें कोई खतरा नहीं है। पुरुष और महिलाएँ दोनों समान रूप से इस खेल को खेल सकते हैं।खो-खो खेल में न किसी गेंद की आवश्यकता होती है, न बल्ले की। इसके लिये केवल १११ फुट लंबे और ५१ फुट चौड़े मैदान की आवश्यकता होती है। दोनों और दस-दस फुट स्थान छोड़कर चार चार फुट ऊँचे, लकड़ी के दो खंभे गाड़ दिए जाते हैं और इन खंभों के बीच की दूरी आठ बराबर भागों में इस प्रकार विभाजित कर दी जाती है कि दोनों दलों के खिलाड़ी एक दूसरे की विरुद्ध दिशाओं की ओर मुँह करके अपने अपने नियत स्थान पर बैठ जाते हैं। प्रत्येक दल को एक-एक पारी के लिए सात सात मिनट दिए जाते हैं और नियत समय में उस दल को अपनी पारी समाप्त करनी पड़ती है। दोनों दलों में से एक-एक खिलाड़ी खड़ा होता है, पीछा करने वाले दल का खिलाड़ी विपक्षी दल के खिलाड़ी को पकड़ने के लिए सीटी बचाते ही दौड़ता है। विपक्षी दल का खिलाड़ी पंक्ति में बैठे हुए खिलाड़ियों का चक्कर लगाता है। जब पीछा करने वाला खिलाड़ी उस भागने वाले खिलाड़ी के निकट आ जाता है, तब वह अपने ही दल के खिलाड़ी के पीछे जाकर 'खो' शब्द का उच्चारण करता है तो वह उठकर भागने लगता है और पीछा करने वाला खिलाड़ी पहले को छोड़कर दूसरे का पीछा करने लगता है।पहले इस खेल का कोई व्यवस्थित नियम न था। खेल की लोकप्रियता के साथ इसके नियम बनते-बिगड़ते रहे। १९१४ ई. में पहली बार पूना के डकन जिमखाना ने अनेक मैदानी खेलों के नियम लिपिबद्ध किए और उनमें खो-खो भी था। तब से उसके बनाए नियम के अनुसार, थोड़े स्थानीय हेर-फेर के साथ यह खेल खेला जाता है।खो-खो की पहली प्रतियोगिता पूना के जिमखाने में १९१८ ई॰ में हुई। फिर सन् १९१९ में बड़ौदा के जिमखाने में भारतीय स्तर पर प्रतियोगिता का आयोजन हुआ। तब से समय-समय पर इस खेल की अखिल भारतीय स्तर पर प्रतियोगिताएँ होती रहती हैं।खेल का मैदान खो-खो का क्रीड़ा क्षेत्र आयताकार होता है। यह 27 X 16 मीटर होता है। मैदान के अंत में दो मुक्त आयताकार क्षेत्र होते हैं। आयताकार की भुजा 16 मीटर और दूसरी भुजा 1.50 मी॰ होती है। इन दोनों आयताकारों के मध्य में दो लकड़ी के स्तम्भ होते हैं। केन्द्रीय गली 24 मी॰ लम्बी और 30 सैंटीमीटर चौड़ी होती है।

**Questions formulated using above source text and the rationale behind it**

| Question | Option A | Option B | Option c | Option D |
|---|---|---|---|---|
| खो-खो किस उद्देश्य के लिए | केवल मनोरंजन के | आत्मरक्षा, आक्रमण और | शारीरिक सौंदर्य बढ़ाने के लिए | मानसिक विकास के लिए |

| विकसित किया गया था? | लिए | प्रत्याक्रमण के कौशल विकसित करने के लिए | | |
| --- | --- | --- | --- | --- |
| खो-खो का जन्मस्थान कहाँ माना जाता है? | पुणे | मुंबई | दिल्ली | कोलकाता |
| निम्नलिखित में से कौन से प्रदेशों में खो-खो खेल का प्रमुख रूप से आयोजन होता है? | पंजाब, हरियाणा और राजस्थान | महाराष्ट्र, गुजरात और मध्य प्रदेश | उत्तराखंड, हिमाचल प्रदेश और जम्मू-कश्मीर | पश्चिम बंगाल, बिहार और ओडिशा |
| खो-खो खेलने के लिए किस वस्तु की आवश्यकता नहीं होती? | गेंद | बल्ला | दोनों गेंद और बल्ला | सीटी |
| खो-खो के मैदान का आकार (मीटर में) क्या है? | 30 x 20 मीटर | 27 x 16 मीटर | 25 x 15 मीटर | 20 x 10 मीटर |
| भारतीय स्तर पर खो-खो प्रतियोगिता का पहला आयोजन कब किया गया? | 1914 | 1918 | 1919 | 1920 |

# Prompt Templates for CultSportQA

## Text-Based Prompts

### 1. Zero-Shot Text-Based Prompt:

You are a multilingual sports expert. Answer the following multiple-choice question:

Question: [Insert text-based question here]

Options:

A) [Option A]

B) [Option B]

C) [Option C]

D) [Option D]

Select the correct option from A, B, C, or D.

---

### 2. Few-Shot (3-Shot) Text-Based Prompt:

You are a multilingual sports expert. Answer the following multiple-choice question based on the provided examples.

Example 1:

Q: [Example question 1]

A: [Correct answer]


Example 2:

Q: [Example question 2]

A: [Correct answer]

Example 3:

Q: [Example question 3]

A: [Correct answer]

Now, answer this question:

Question: [Insert text-based question here]

Options:

A) [Option A]

B) [Option B]

C) [Option C]

D) [Option D]

Select the correct option from A, B, C, or D.

---

**3. Chain-of-Thought (CoT) Text-Based Prompt:**

You are a multilingual sports expert. Answer the following question step-by-step:

Question: [Insert text-based question here]

Options:

A) [Option A]

B) [Option B]

C) [Option C]

D) [Option D]

Think carefully step-by-step before selecting the final answer:

1. Analyze the question.

2. Recall relevant sports knowledge.

3. Eliminate incorrect options.

4. Choose the most suitable answer.

Final Answer: [Provide answer as A, B, C, or D]

---

**Image-Based Prompts**

**4. Zero-Shot Image-Based Prompt:**

You are a multilingual sports expert. Analyze the following image and answer the multiple-choice question:

[Insert Image Here]

Question: [Insert image-based question here]

Options:

A) [Option A]

B) [Option B]

C) [Option C]

D) [Option D]

Select the correct option from A, B, C, or D.

---

**5. Few-Shot (3-Shot) Image-Based Prompt:**

You are a multilingual sports expert. Analyze the following examples and answer the multiple-choice question.

Example 1:

[Image 1]

Q: [Example question 1]

A: [Correct answer]

Example 2:

[Image 2]

Q: [Example question 2]

A: [Correct answer]

Example 3:

[Image 3]

Q: [Example question 3]

A: [Correct answer]

Now answer this:

[Insert Image Here]

Question: [Insert image-based question here]

Options:

A) [Option A]

B) [Option B]

C) [Option C]

D) [Option D]

Select the correct option from A, B, C, or D.

---

## 6. Chain-of-Thought (CoT) Image-Based Prompt:

You are a multilingual sports expert. Analyze the following image and answer the question with reasoning.

[Insert Image Here]

Question: [Insert image-based question here]

Options:

A) [Option A]

B) [Option B]

C) [Option C]

D) [Option D]

Think carefully step-by-step before selecting the final answer:

1. Describe what you observe in the image.

2. Analyze any sports-related cues (e.g., equipment, uniforms).

3. Eliminate unlikely options.

4. Select the most suitable answer.

Final Answer: [Provide answer as A, B, C, or D]

# Traditional Games by Continent

## Games in Asia

1. Kho-Kho
2. Mallakhamb
3. Mushti Yuddha
4. Lagori
5. Kalarippayattu
6. Nadan Panthu Kali
7. Kabbadi
8. Gilli Danda
9. Pambaram
10. Buffalo Race
11. Carrom
12. Malakhra
13. Buzkashi
14. Gatka
15. Tent Pegging
16. Chaturanga
17. Oonch Neech
18. Karapan Sapi
19. Pencak Silat
20. Peresean
21. Sepak Takraw
22. Tarung Derajat
23. Egrang
24. Panjat Pinang
25. Karapan Ayam
26. Lari Balok
27. Chon Wua
28. Kaeng Ruer
29. Len Wow
30. Muay Thai
31. Dern Kala
32. Boli Khela
33. Ha-Du-Du
34. Lathi Khela
35. Nouka-Baich
36. Golla-Chut
37. Latim
38. Kanamachi
39. Cuju
40. Dragon Boat Racing
41. Go
42. Jianzi
43. Tai Chi
44. Liubo
45. Dariabandha (Bangladesh)
46. Pasha Khela

### Games in Europe

1. Joutes
2. Tambourin
3. Pelota
4. Camargue
5. Petanque
6. Boule Lyonnaise
7. Course Camarguaise
8. Balle au Poing
9. Boules Carrées
10. Bosseln
11. Klootschießen
12. Radball
13. Quoits
14. Eisstockschießen
15. Fingerhakeln (Finger Wrestling)
16. Morra
17. Calcio Storico
18. Ruzzola
19. Pallanuoto
20. Pallone col Bracciale
21. Palla Tamburello
22. Corsa con Sacco (Sack Race)
23. Schleuderball (Germany)
24. Water Jousting (France)

### Games in Africa

1. Nuba Wrestling
2. Camel Racing
3. Hyena
4. Boruboru
5. Cow Fighting
6. Genna
7. Gebeta
8. Yeferas Guks
9. Sidama
10. Afalula
11. Alemungula
12. Dala (Herding the Cows)
13. Nubian Stick Fighting

# Annotation Guidelines for Evaluating Questions

## Objective

The purpose of this guideline is to ensure consistent and accurate evaluation of questions in terms of their correct area, their relation to sports, and their relation to a specific country. Annotators must assess whether each question aligns with these criteria based on the provided definitions and examples.

## Annotation Criteria

**1. Evaluating Whether the Question is Related to Traditional Sports of their country and its origin is from that country**
● The question must be directly related to a sport, sporting event, or activity.
It should involve aspects such as rules, history, equipment, famous
players, strategies, or tournaments.

✅ **Valid Example :**
*"Which country hosts the annual Sepak Takraw World Championship?"*
(✅ The question clearly relates to a specific sport.)

❌ **Invalid Example :**
*"What is the capital of Ethiopia?"*
(❌ This question is about geography, not sports.)

**2. Evaluating Whether the Question is Related to a Country**
● The question should explicitly or implicitly refer to a specific
country.If the country is not directly mentioned, it should still be clear from the
context.Questions asking about sports in general or in multiple countries
should not be marked as country-specific.

✅ **Valid Example :**
*"Which traditional Ethiopian game is played with wooden sticks and
involves tactical movements?"*
(✅ The question is sports-related and explicitly mentions Ethiopia.)

❌ **Invalid Example :**
*"What is the national dish of Sudan?"*
(❌ The question is country-related but not about sports.)

# Question Types & Their Annotation Guidelines

**1. Evaluating History-Based Questions :**
These questions focus on the origins, historical significance, or evolution of a sport. They must be fact-based and grounded in verifiable history.

✅ **Valid Example:**
*"When was Kho-Kho first included in the National Games of India?"*
(✅ This question relates to the historical development of the sport.)
❌ **Invalid Example:**
*"Who was the greatest Kabaddi player of all time?"*
(❌ This question is subjective and lacks historical specificity.)

**2. Evaluating Rule-Based Questions**
These questions focus on the rules, gameplay mechanics, or
official regulations of a sport. They should be objective and specific to a sport's gameplay.
✅ **Valid Example:**
*"How many players are allowed on a Kabaddi team during a match?"*
(✅ The question asks about official gameplay rules.)
❌ **Invalid Example:**
*"Which is the best strategy to win in Kho-Kho?"*
(❌ The question is subjective and not strictly about rules.)

**3. Evaluating Image-Based Questions**
These questions are accompanied by an image and require the annotator to check if the visual elements are correctly aligned with the textual question. They should involve identifying a sport, equipment, movement, or player role.

✅ **Valid Example:**
*"Based on the image, which traditional African sport involves the use of long sticks in combat?"*
(✅ The question asks for an identification based on the image.)
❌ **Invalid Example:**
*"How many goals did a player score in this match?"* (without a scoreboard or necessary context in the image)
(❌ The question cannot be answered using the given image alone.)

**4. Evaluating Scenario-Based Questions**
These questions assess the ability to analyze a situation within a sport and determine the correct decision or action. The scenario should be realistic and sport-specific.

✅ **Valid Example:**
*"In a Kabaddi match, if a player successfully touches an opponent and returns to their side without being tackled, what is the outcome?"*
(✅ The question describes an in-game scenario and requires an understanding of the sport's rules.)

❌ **Invalid Example:**
*"What is the best way to win a Kho-Kho game?"*
(❌ This question is too broad and lacks a structured scenario.)

# Annotation Labels Guidelines

**Each question should be assigned a label based on the following**

**categories:**
**1. Area Correctness**
Correct (✔): The question belongs to the relevant dataset domain area.
Incorrect (❌): The question does not align with the intended.

**2. Sports Relevance**

Sports-related (✔): The question is directly about a sport or sporting event.
Not sports-related (❌): The question is about unrelated topics.

**3. Country Relevance**

Country-specific (✔): The question is explicitly tied to a country.
Not country-specific (❌): The question does not refer to a specific country.

**4. Question TypeEach question should be categorized into one of the following types:**

History-Based (H)
Rule-Based (R)
Scenario-Based (S)

## Final Notes for Annotators
If a question is incorrectly categorized, mark it appropriately and provide a suggested correction. If a question is ambiguous, flag it for review by a senior annotator.Ensure that culturally sensitive and respectful language is maintained in all annotations.

# Validation Guidelines
The purpose of this guideline is to ensure that annotated questions meet the defined quality standards for correct area classification, sports relevance, country relevance, and question type categorization. This validation process helps maintain accuracy, consistency, and cultural sensitivity across the dataset.

**Validation Criteria** :
1. Verifying Whether the Question is Related to Traditional Sports of its Country
Ensure that the question directly relates to a traditional sport that originates from the specified country.

2. The sport should be recognized as a culturally significant activity and should not be confused with globally popular sports. The question should cover historical aspects, rules, equipment, tournaments, or well-known figures in the sport.

**✅ Valid Example:**
*"Which country hosts the annual Sepak Takraw World Championship?"*
(✅ The question correctly relates to a traditional sport and references a country.)

**❌ Invalid Example:**
*"Which country won the last FIFA World Cup?"*
(❌ FIFA World Cup is not a traditional or indigenous sport.)

**Validation Process:**

Check authenticity: Verify the sport's origin and cultural relevance using reliable sources. Reject incorrect sports: If the question refers to modern or globally commercialized sports, mark it as incorrect.

### 2. Verifying Whether the Question is Related to a Country

The question must explicitly or implicitly mention a country related to the sport. If the country is not mentioned, the context should still make it clear. If the sport is played across multiple countries, it should not be marked as country-specific.

**✅ Valid Example:**
*"Which traditional Ethiopian game is played with wooden sticks and involves tactical movements?"*
(✅ The question is clearly related to Ethiopia and mentions a traditional game.)

**❌ Invalid Example:**
*"What is the national dish of Sudan?"*
(❌ The question is country-related but not about sports.)

**Validation Process:**
Verify country relevance using credible sources. Ensure specificity: The question should be linked to one country unless multiple origins are explicitly referenced.

### Validation by Question Type
Each question type has specific validation steps to ensure clarity, accuracy, and relevance.

### 3. History-Based Questions Validation

These questions should focus on the historical significance, origin, or evolution of a traditional sport. They must be fact-based, verifiable, and free from subjective opinions. Ensure the historical reference is accurate and properly framed.

✅ **Valid Example:**
*"When was Kho-Kho first included in the National Games of India?"*
(✅ Verifiable historical event.) ❌ **Invalid Example:**
*"Who was the greatest Kabaddi player of all time?"*
(❌ Subjective question, not grounded in history.)

**Validation Process:**

**Cross-check the historical claim** with multiple sources. **Reject vague or opinion-based questions** that lack factual grounding.

### 4. Rule-Based Questions Validation

Questions must focus on specific rules, regulations, or gameplay mechanics of a sport. They should be objective, measurable, and universally recognized within the sport.
Avoid opinion-based or strategy-based questions.

✅ **Valid Example:**
*"How many players are allowed on a Kabaddi team during a match?"*
(✅ The question correctly asks about an official rule.)
❌ **Invalid Example:**
*"Which is the best strategy to win in Kho-Kho?"*
(❌ The question is subjective and not strictly about rules.)

**Validation Process:**

Check the accuracy of the rule from an official governing body or sports authority. Ensure the question refers to an actual rule rather than strategy or subjective gameplay aspects. 5. Image-Based Questions Validation The image should be clear, relevant, and accurately represent the sport being referenced in the question. The question must require the image for answering—if it can be answered without the image, it may not be valid. The image should be culturally appropriate and avoid offensive or misleading depictions.

✅ **Valid Example:**
*"Based on the image, which traditional African sport involves the use of long sticks in combat?"*
(✅ The image is essential for answering the question.)

❌ **Invalid Example:**
*"How many goals did a player score in this match?"*
(❌ If the image lacks a scoreboard or relevant context, the question is invalid.)

**Validation Process:**
**Check whether the image is necessary** for answering the question.
**Verify image authenticity**:-nsure it represents the correct sport.
**Ensure visual clarity**—blurry or unclear images should be flagged for replacement.

**6. Scenario-Based Questions Validation**
● These questions should describe a realistic in-game situation and ask for the appropriate action or rule-based outcome.
● They should be clear, structured, and have only one correct answer.
● Avoid speculative or open-ended questions.

✅ **Valid Example:**
*"In a Kabaddi match, if a player successfully touches an opponent and returns to their side without being tackled, what is the outcome?"*
(✅ The scenario is realistic and directly tied to a rule.)
❌ **Invalid Example:**
*"What is the best way to win a Kho-Kho game?"*
(❌ The question is too broad and lacks a structured scenario.)

**Validation Process:**
● Ensure the scenario is plausible and follows the actual gameplay mechanics. Check whether the answer is well-defined and not open to multiple interpretations.

**Validation Labels**
Each validated question should be categorized with the following labels:
1. **Area Correctness**
   ○ **Correct (✔)**: The question belongs to the relevant dataset domain.
   ○ **Incorrect (❌)**: The question does not align with the intended area.
2. **Sports Relevance**
   ○ **Sports-related (✔)**: The question is directly about a sport or sporting event.
   ○ **Not sports-related (❌)**: The question is about unrelated topics such as history, geography, or politics.
3. **Country Relevance**
   ○ **Country-specific (✔)**: The question is explicitly tied to a country.
   ○ **Not country-specific (❌)**: The question does not refer to a specific country.
4. **Question Type**
   ○ **History-Based (H)**

○ **Rule-Based (R)**
○ **Scenario-Based (S)**
○ **Image-Based (I)**

## Final Notes for Validators

**Provide feedback** if a question needs revision or clarification. **Flag questions** that contain **unclear language, incorrect information, or cultural insensitivity**. **Maintain consistency** in labeling and categorization across the dataset. **Consult subject matter experts** if uncertain about a question's accuracy or relevance. By following this validation guideline, we ensure that all questions meet **high-quality standards** for accuracy, clarity, and cultural sensitivity



\label{sec:appendix}

\begin{figure*}[htbp]
    \centering
    \begin{subfigure}{0.32\textwidth}
        \centering
        \includegraphics[height=6cm,width=\linewidth]{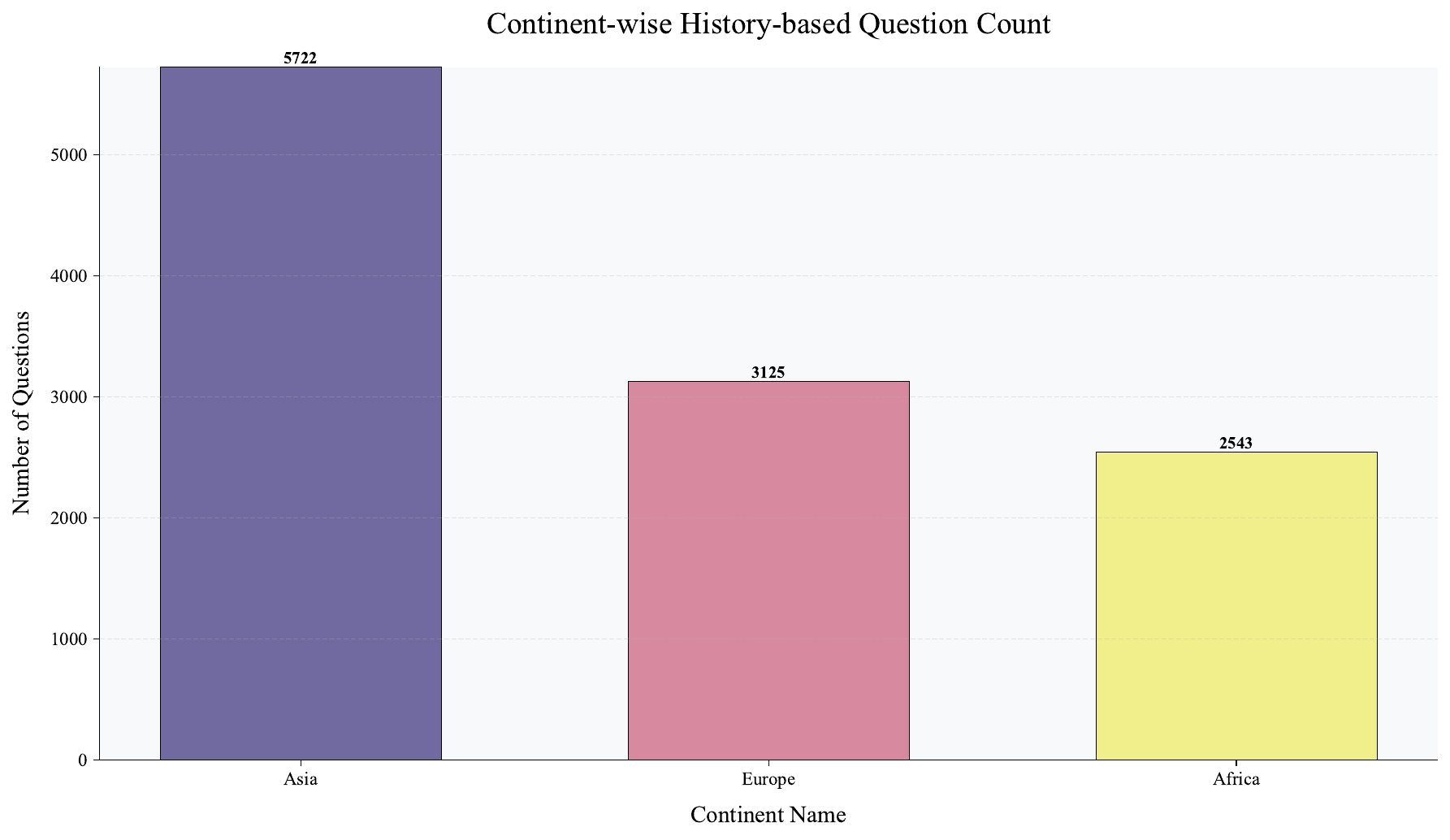}
        \caption{History-based}
        \label{fig:continent_history}
    \end{subfigure}
    \hfill
    \begin{subfigure}{0.32\textwidth}
        \centering
        \includegraphics[height=6cm,width=\linewidth]{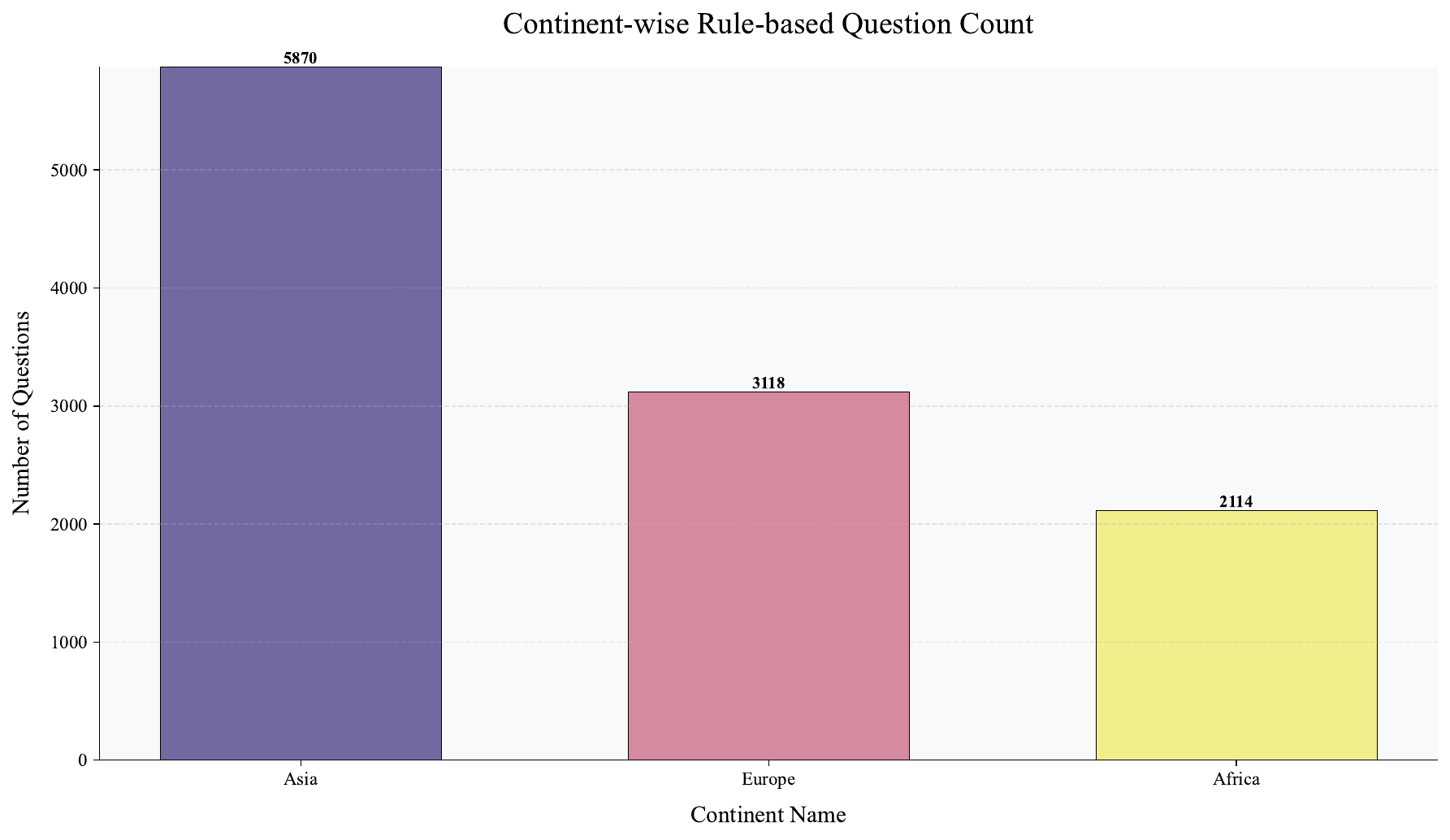}
        \caption{Rule-based}
        \label{fig:continent_rule}
    \end{subfigure}
    \hfill
    \begin{subfigure}{0.32\textwidth}
        \centering
        \includegraphics[height=6cm,width=\linewidth]{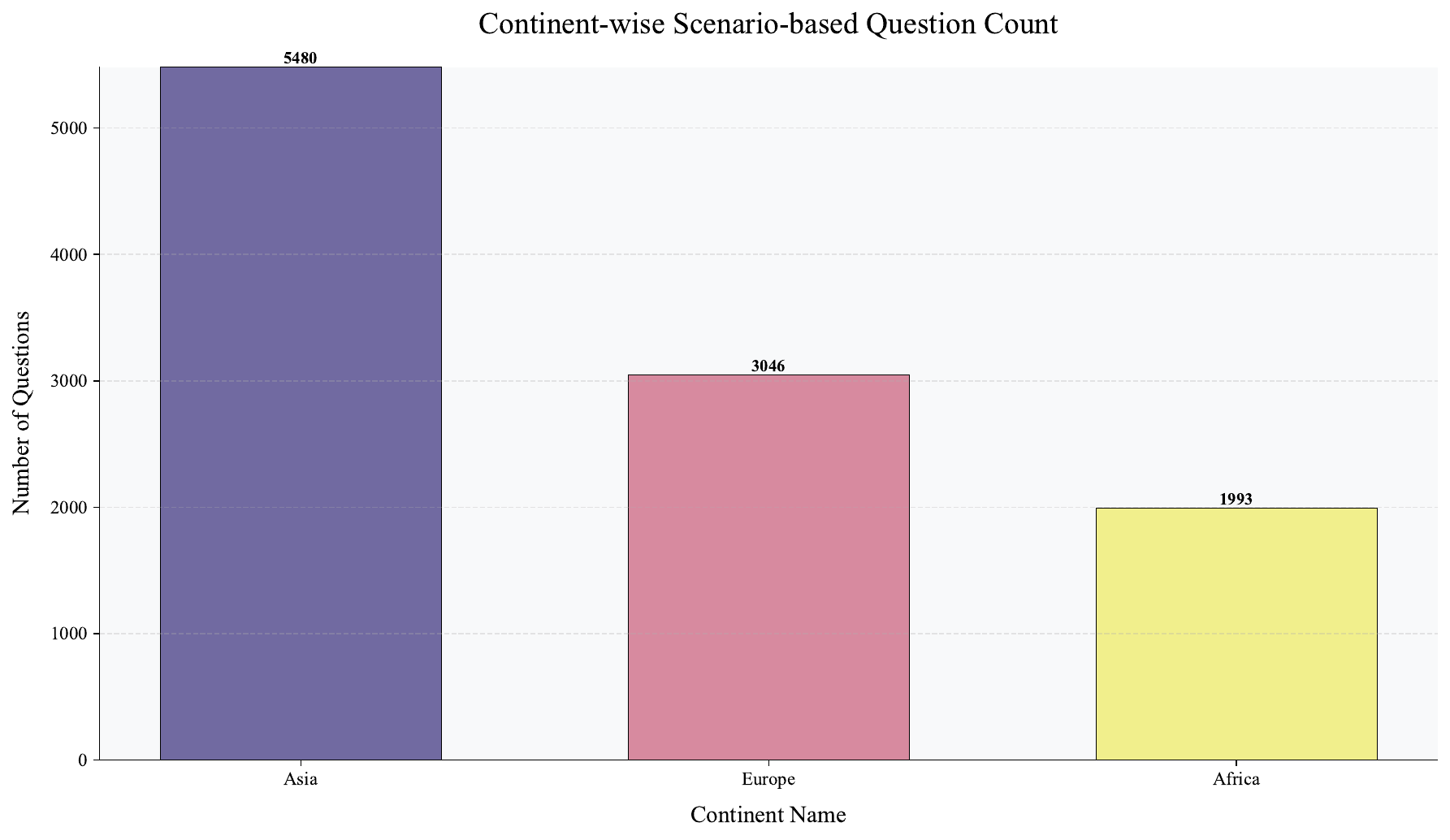}
        \caption{Scenario-based}
        \label{fig:continent_scenario}
    \end{subfigure}

    \caption{Statistics of history-based, rule-based, and scenario-based questions across continents}
    \label{fig:continent_comparison}
\end{figure*}


\begin{figure*}
\includegraphics[height = 6 cm,width=1\textwidth]{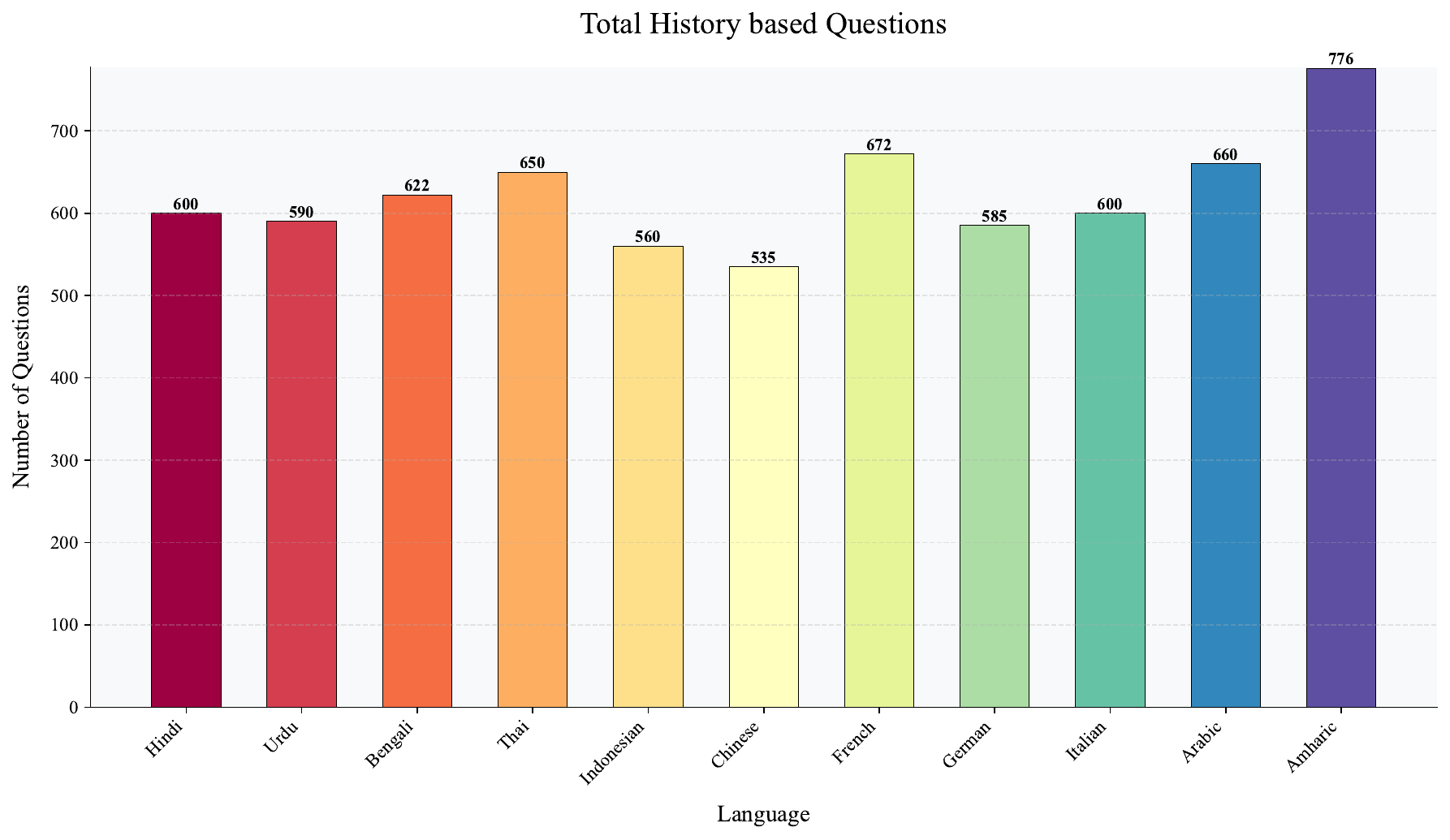}
\caption{Statistics of history-based questions across languages} 
\label{HBQL}
\end{figure*}

\begin{figure*}
\includegraphics[height = 6 cm,width=1\textwidth]{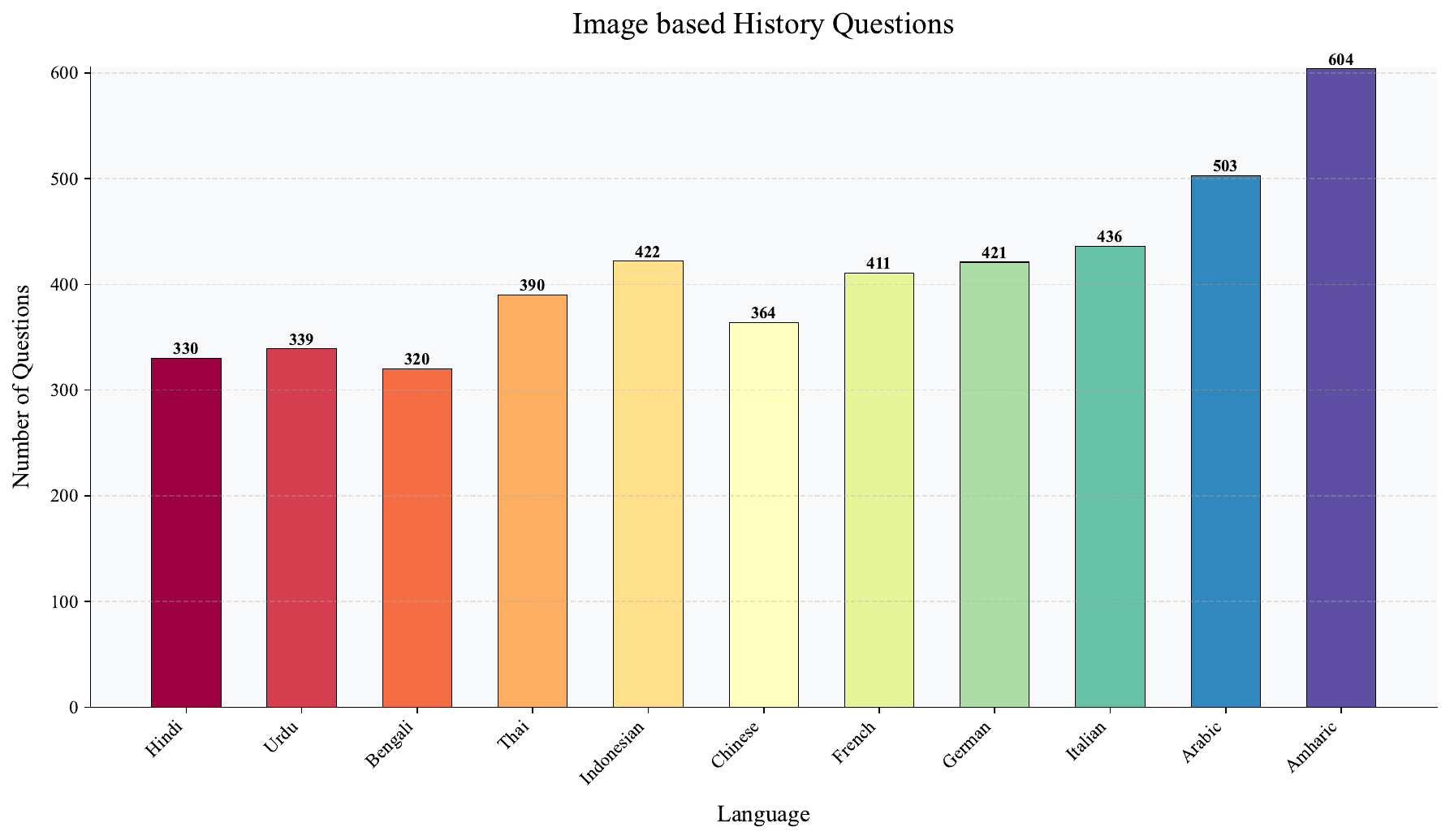}
\caption{Statistics of image-based history questions across languages} 
\label{IBHQL}

\end{figure*}
\begin{figure*}
\includegraphics[height = 6 cm,width=1\textwidth]{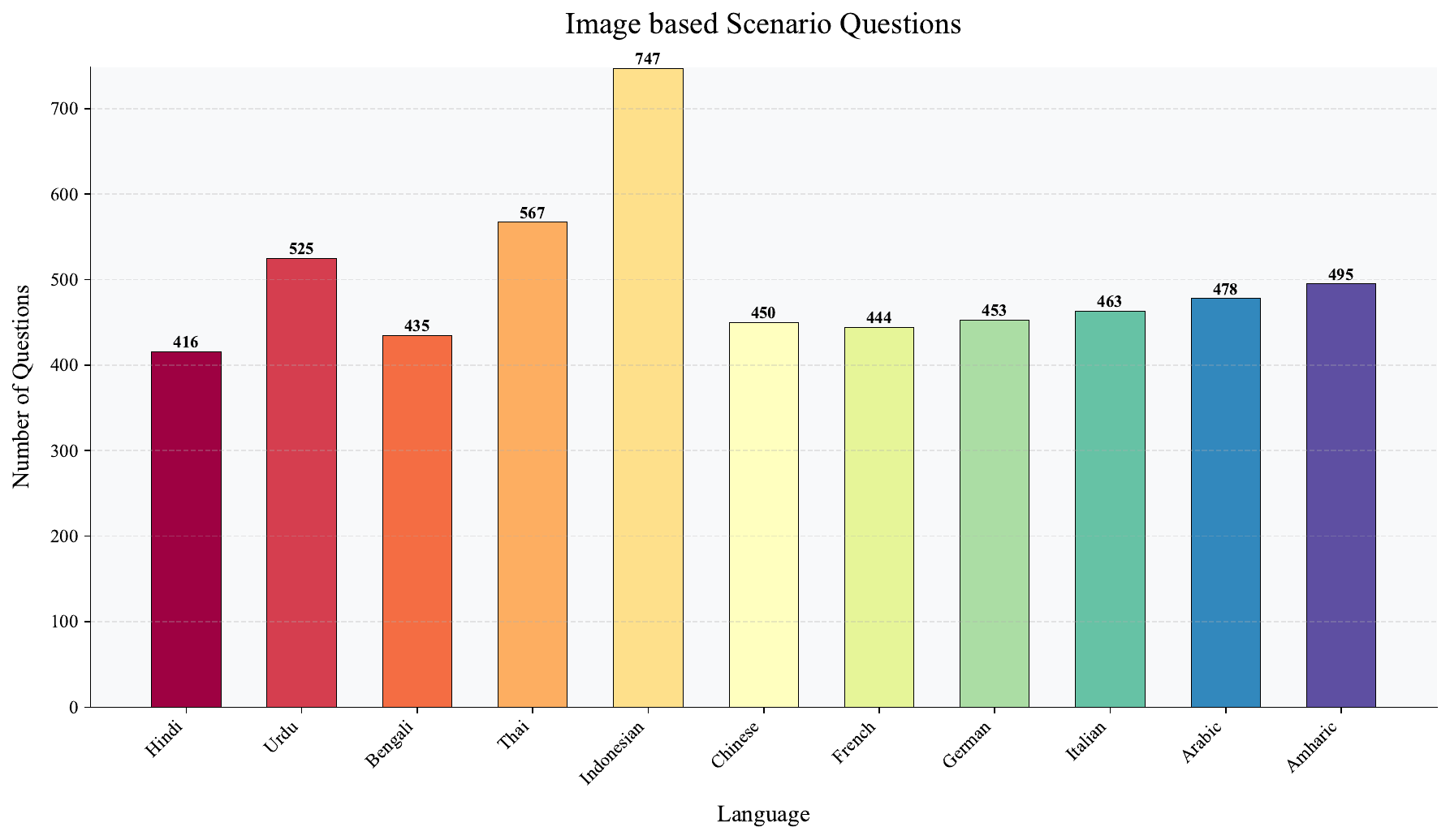}
\caption{Statistics of image-based scenario questions across languages} 
\label{IBSQL}
\end{figure*}

\begin{figure*}
\includegraphics[height = 6 cm,width=1\textwidth]{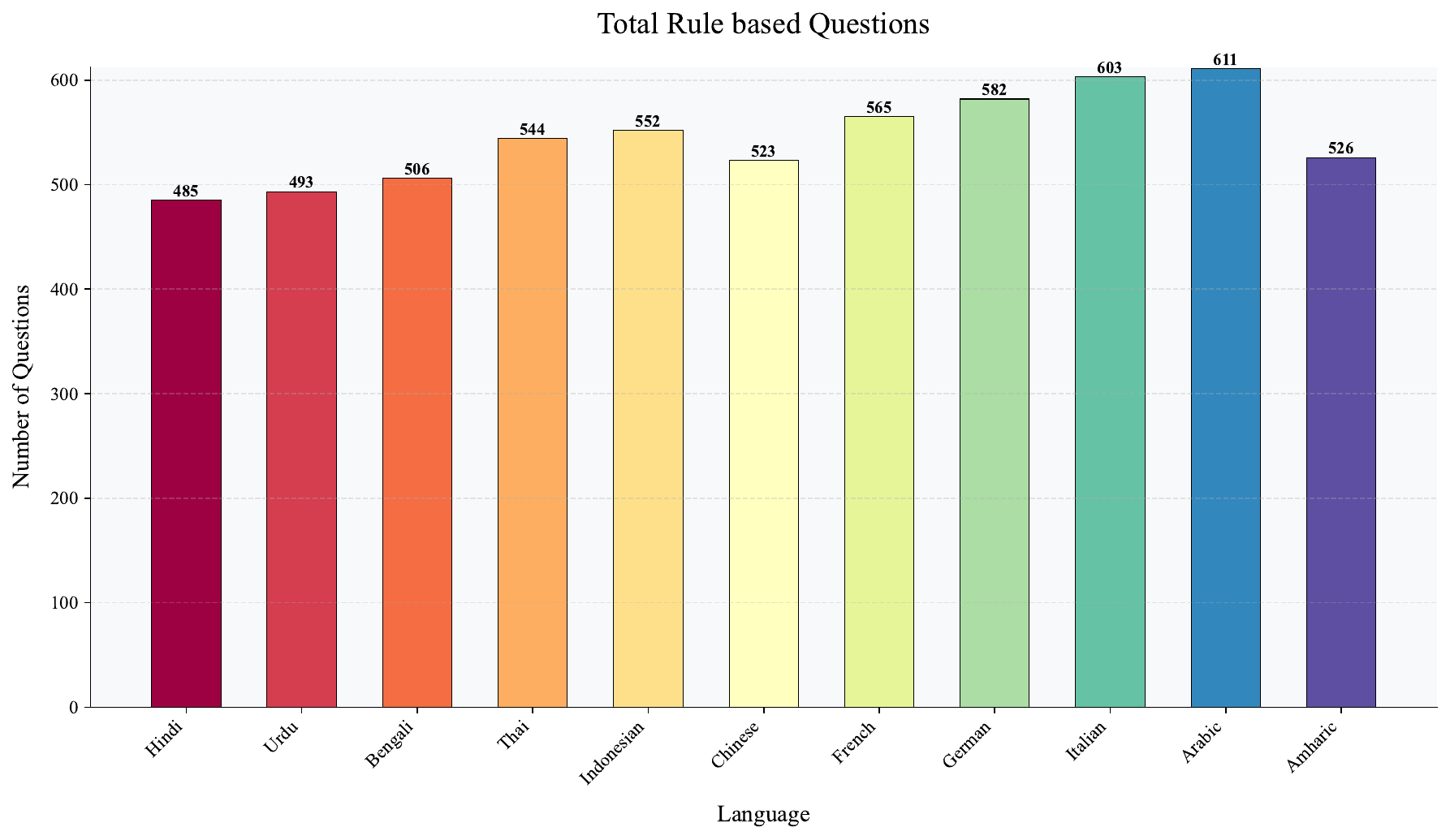}
\caption{Statistics of rule-based questions across languages} 
\label{RBQL}
\end{figure*}

\begin{figure*}
\includegraphics[height = 6 cm,width=1\textwidth]{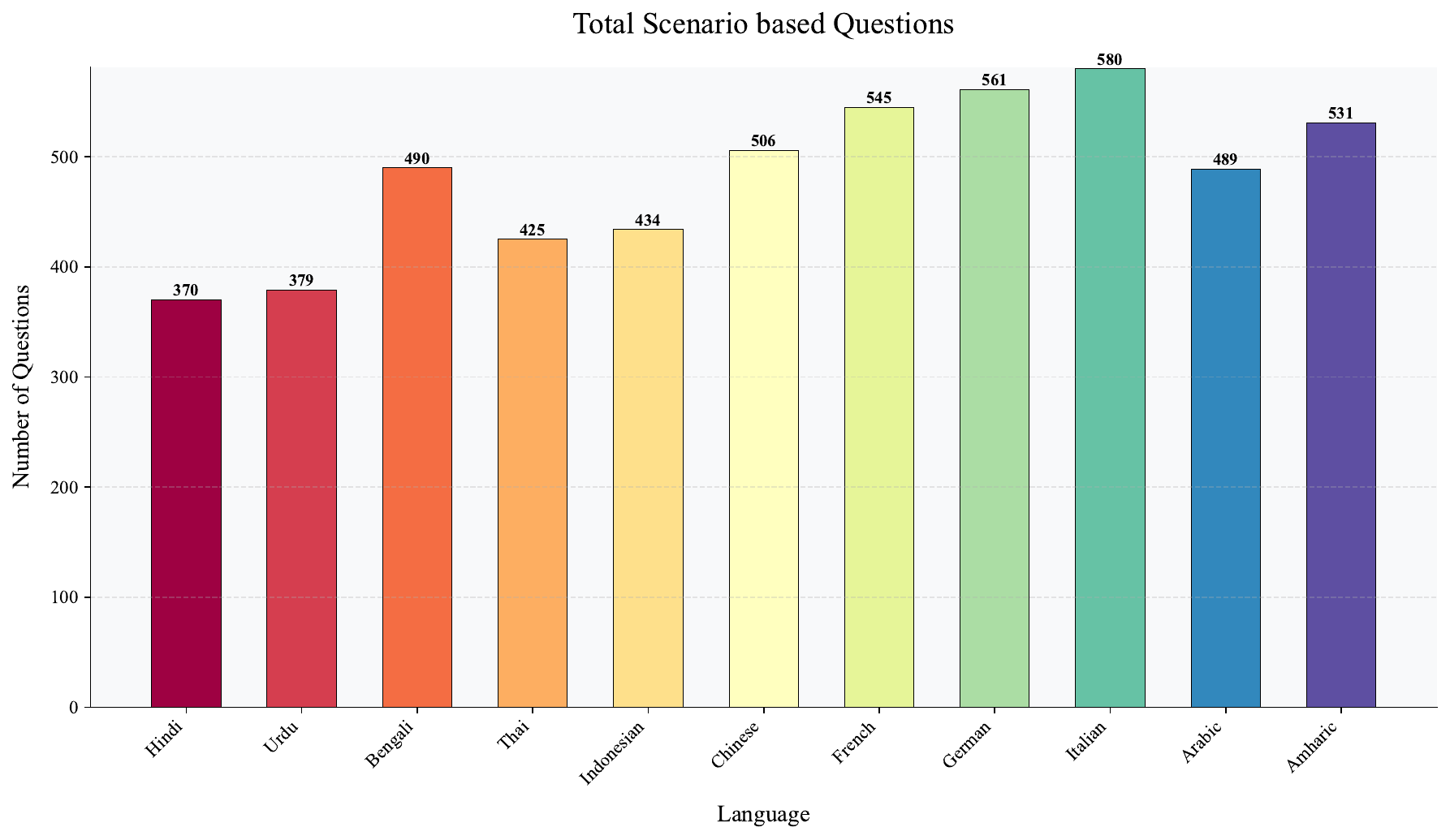}
\caption{Statistics of scenario-based questions across languages} 
\label{SBQL}
\end{figure*}

\begin{figure*}
\includegraphics[height = 6 cm,width=1\textwidth]{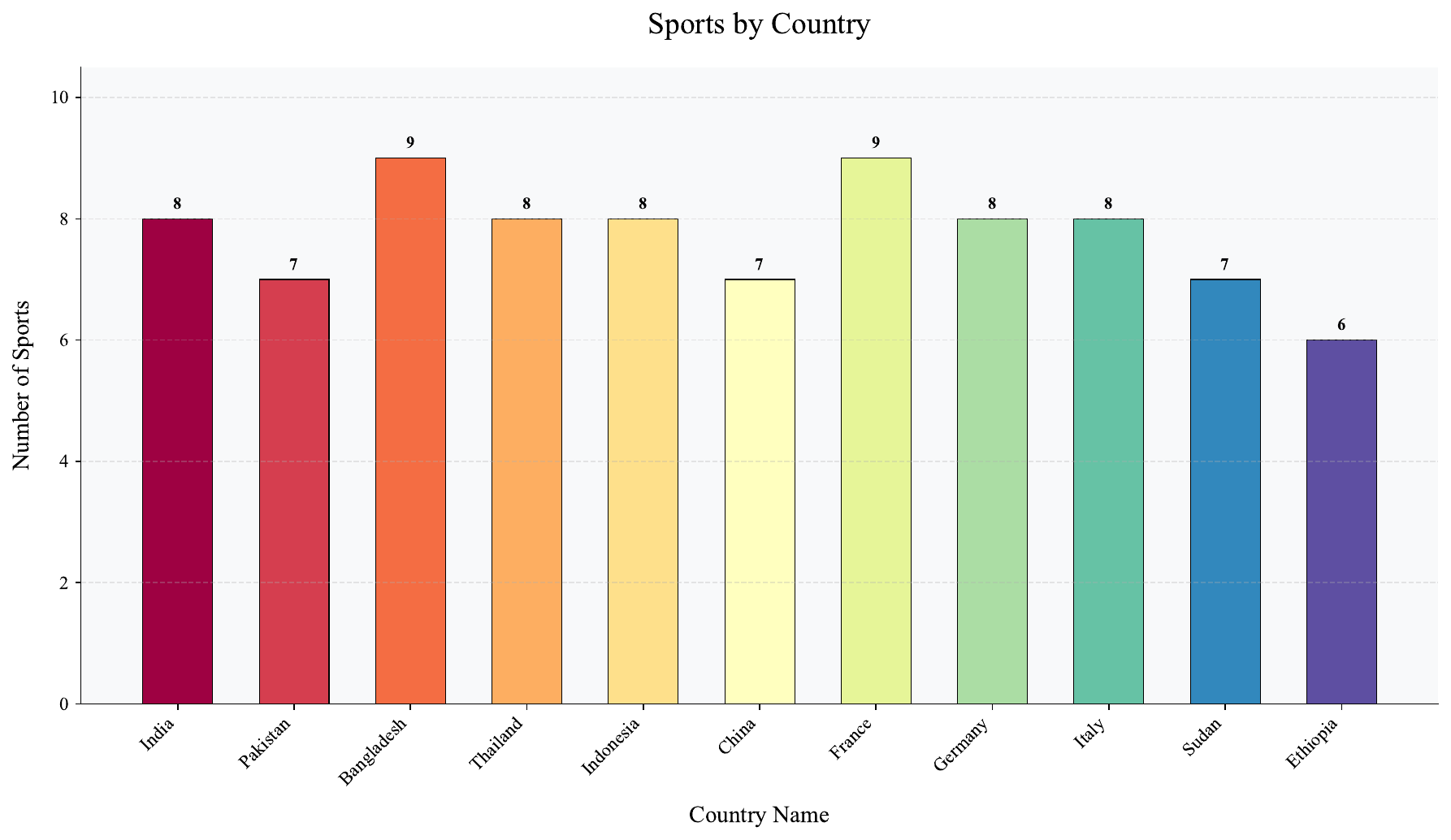}
\caption{Statistics of sports across countries} 
\label{SC}
\end{figure*}

\begin{figure*}
\includegraphics[height = 6 cm,width=1\textwidth]{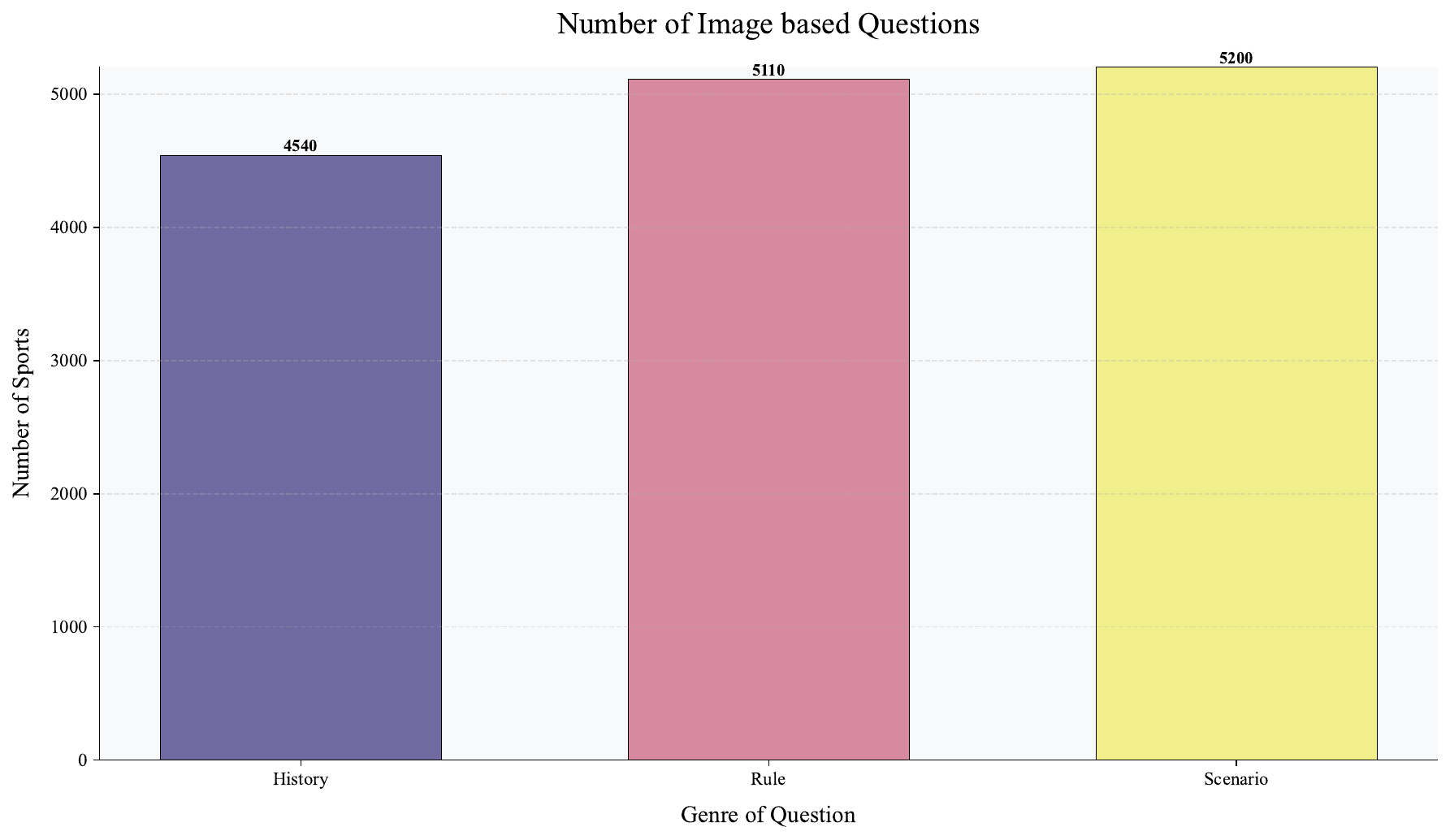}
\caption{Statistics of image-based questions across types} 
\label{IBQT}
\end{figure*}

\begin{figure*}
\includegraphics[height = 6 cm,width=1\textwidth]{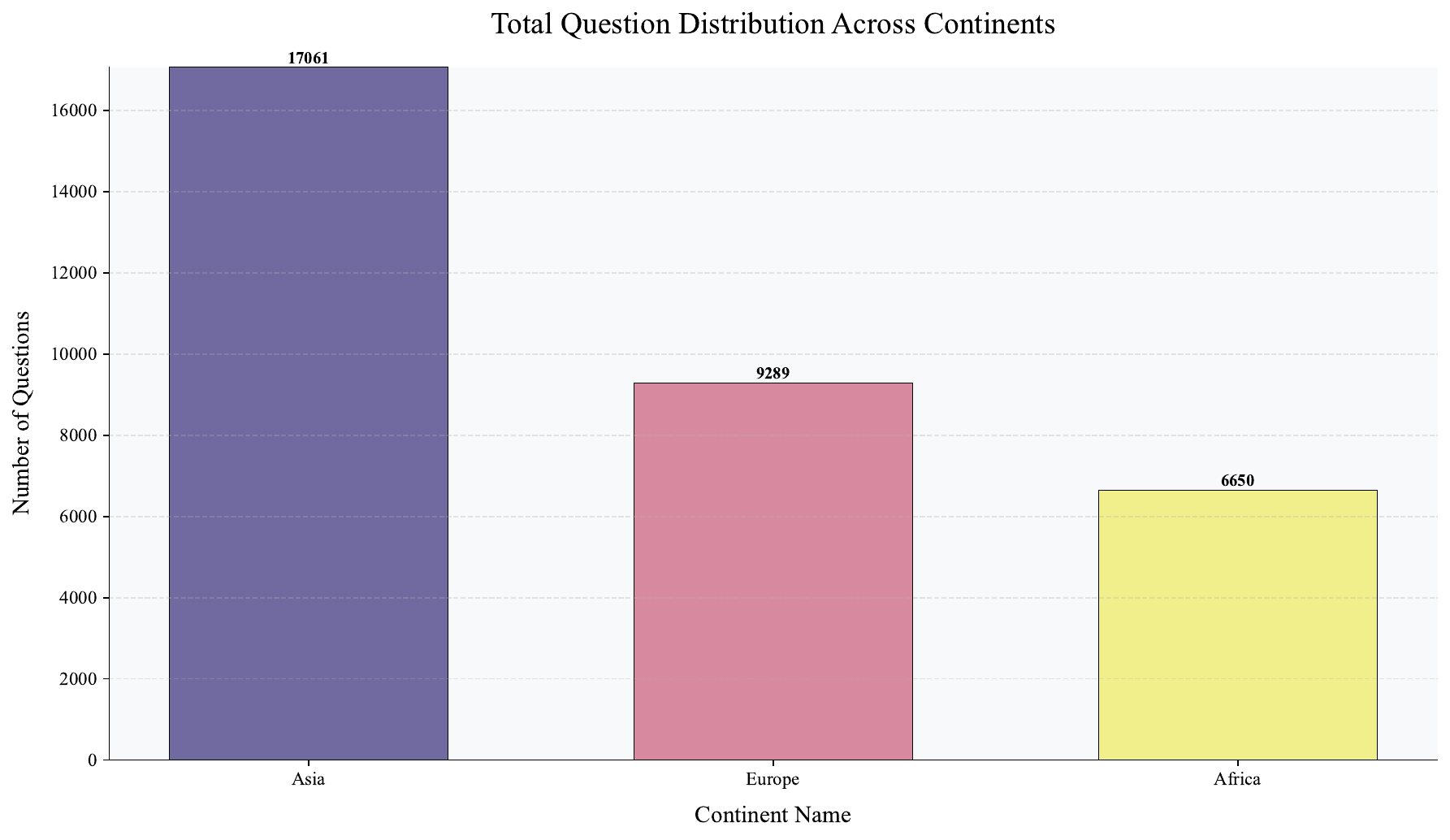}
\caption{Statistics of questions across continents} 
\label{QC}
\end{figure*}

\begin{figure*}[hbt!]
    \centering
    \begin{subfigure}[b]{0.32\textwidth}
        \centering
        \includegraphics[width=\textwidth]{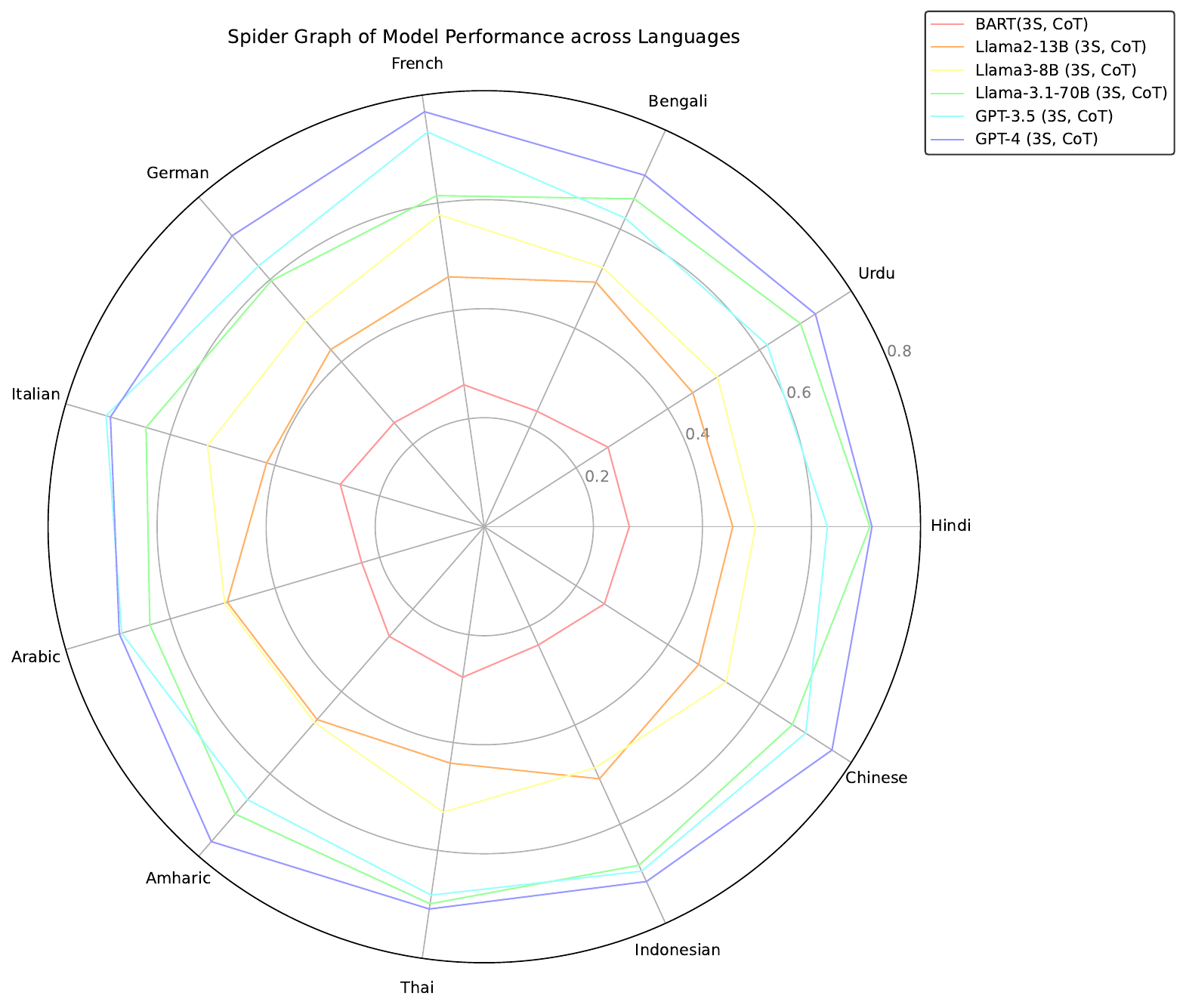}
        \caption{CoT-based results of LLMs across languages}
    \end{subfigure}
    \hfill
    \begin{subfigure}[b]{0.32\textwidth}
        \centering
        \includegraphics[width=\textwidth]{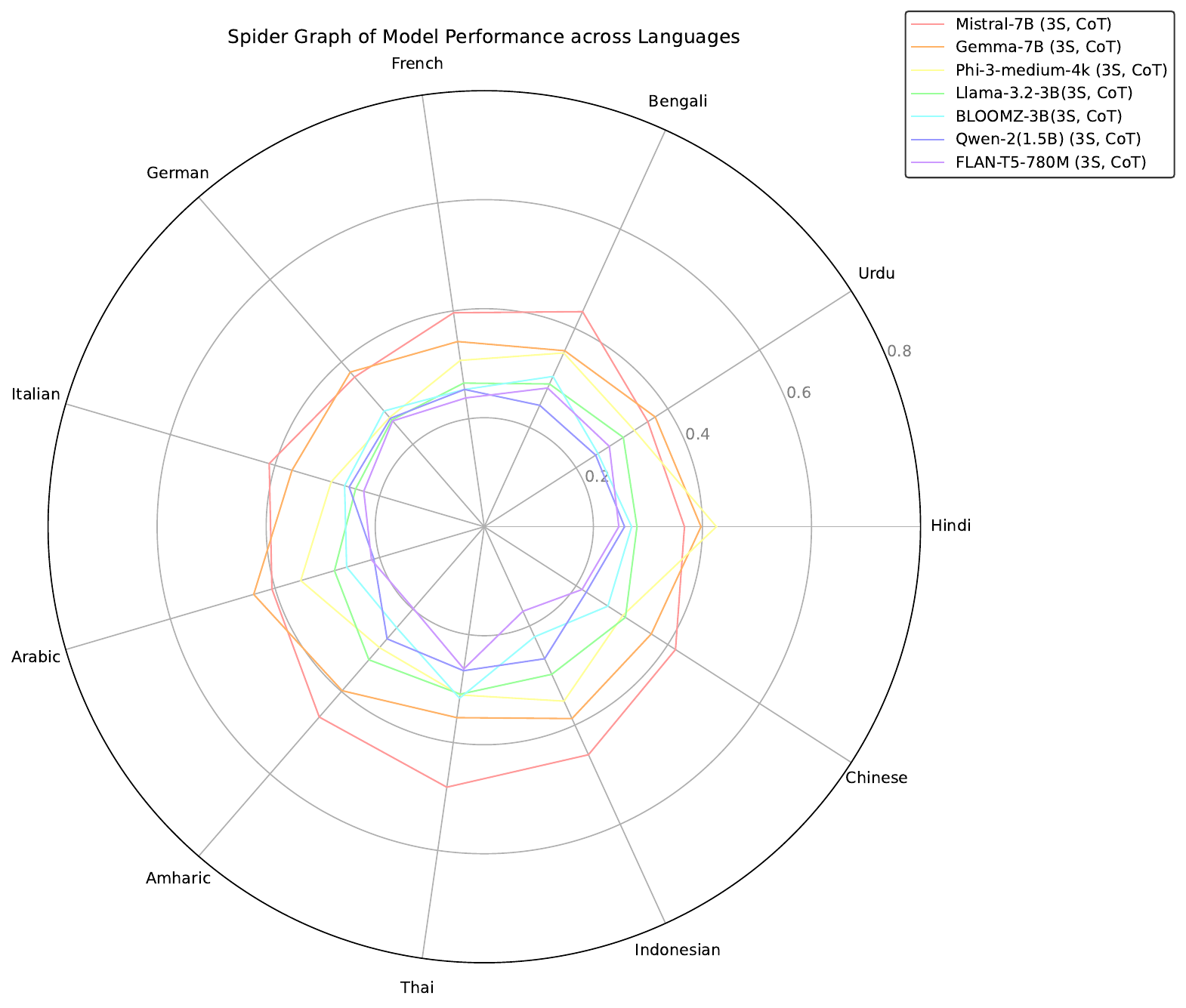}
        \caption{CoT-based results of SLMs across languages}
    \end{subfigure}
    \hfill
    \begin{subfigure}[b]{0.32\textwidth}
        \centering
        \includegraphics[width=\textwidth]{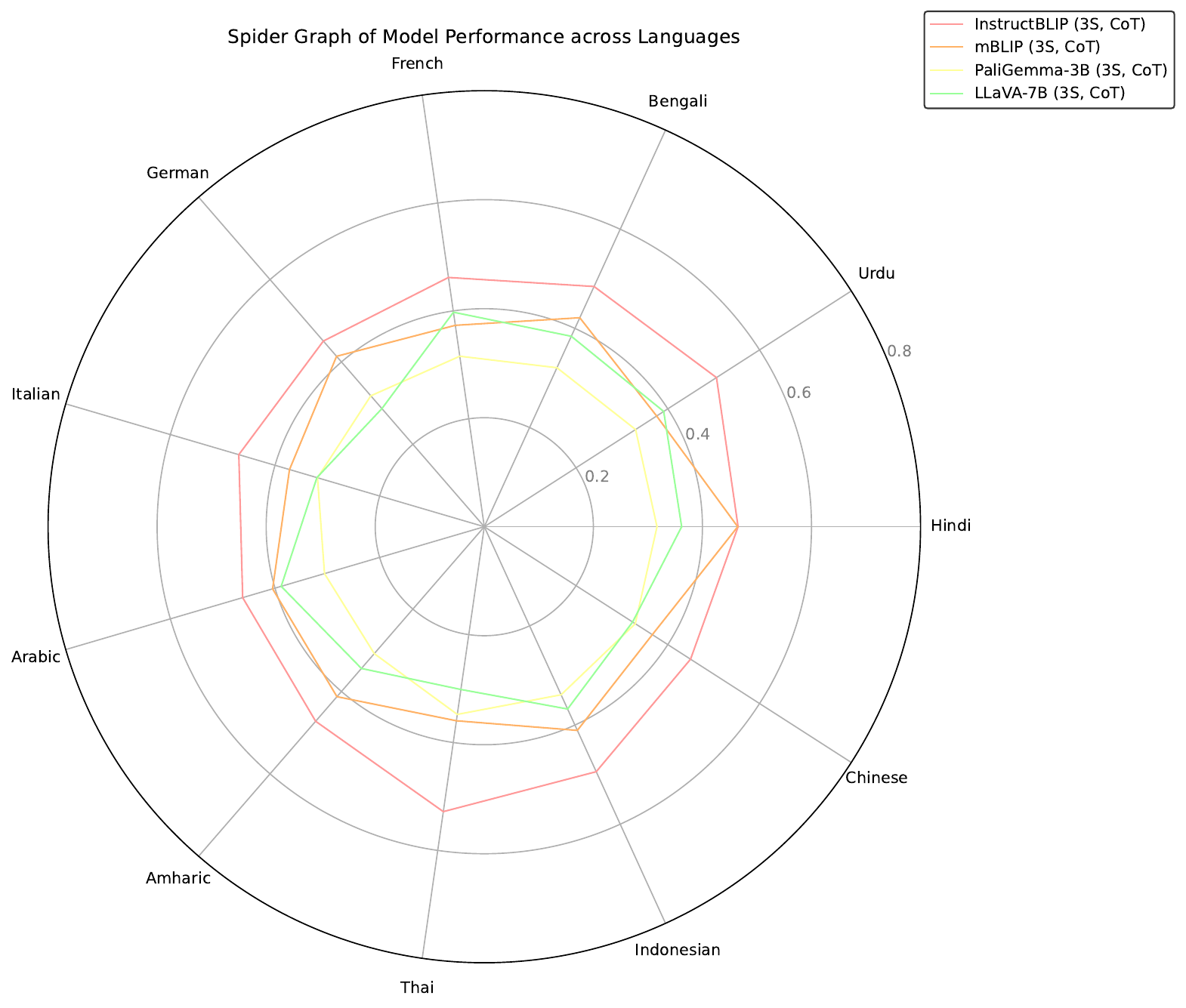}
        \caption{CoT-based results of MLLMs across languages}
        \label{fig:Lang_3S_COT_VLM}
    \end{subfigure}
    \caption{CoT-based results of models across languages}
\end{figure*}

\begin{figure*}[hbt!]
    \centering
    \begin{subfigure}[b]{0.32\textwidth}
        \centering
        \includegraphics[width=\textwidth]{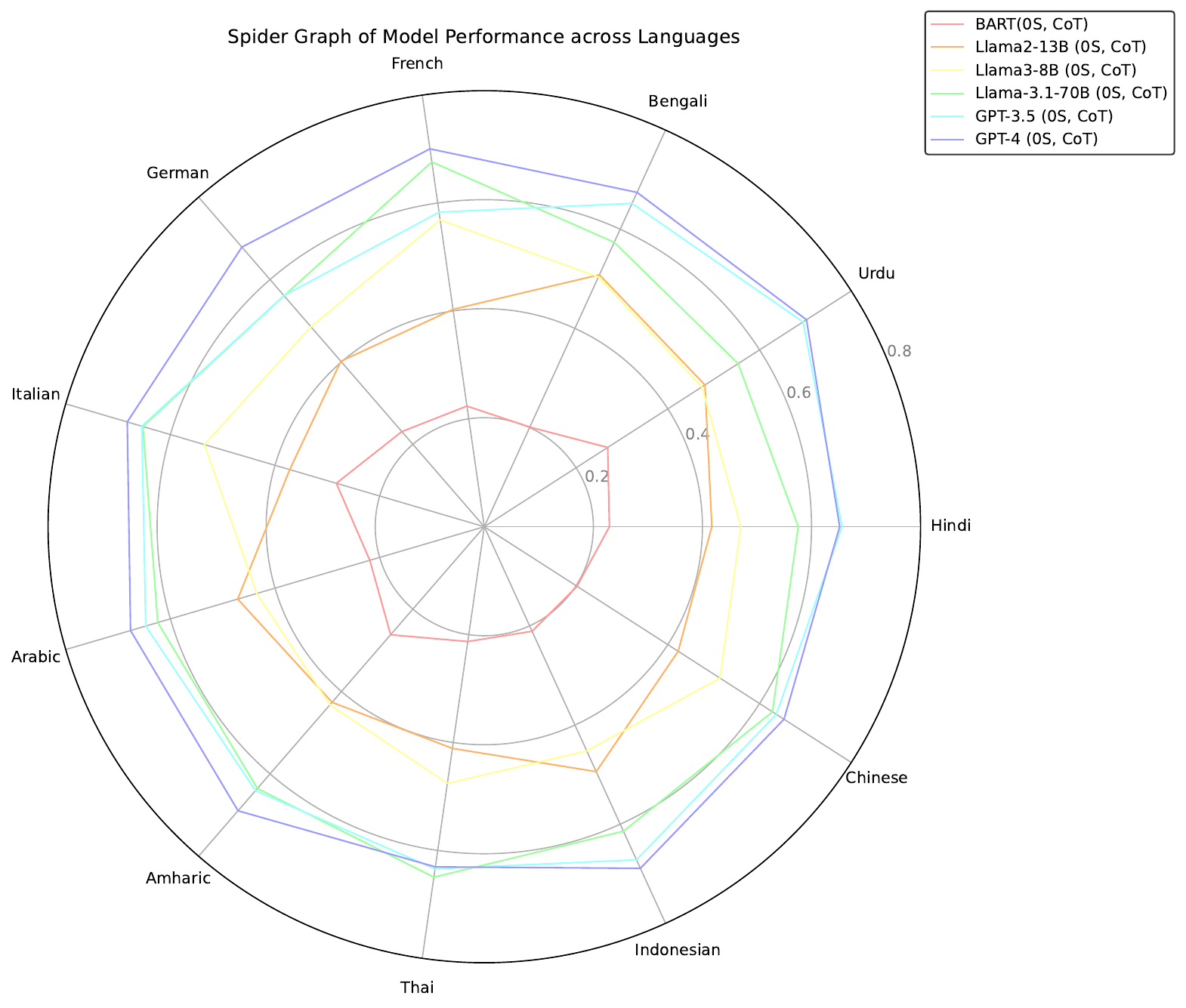}
        \caption{Few-shot-based results of LLMs across languages}
        \label{fig:lzero_cot_llm}
    \end{subfigure}
    \hfill
    \begin{subfigure}[b]{0.32\textwidth}
        \centering
        \includegraphics[width=\textwidth]{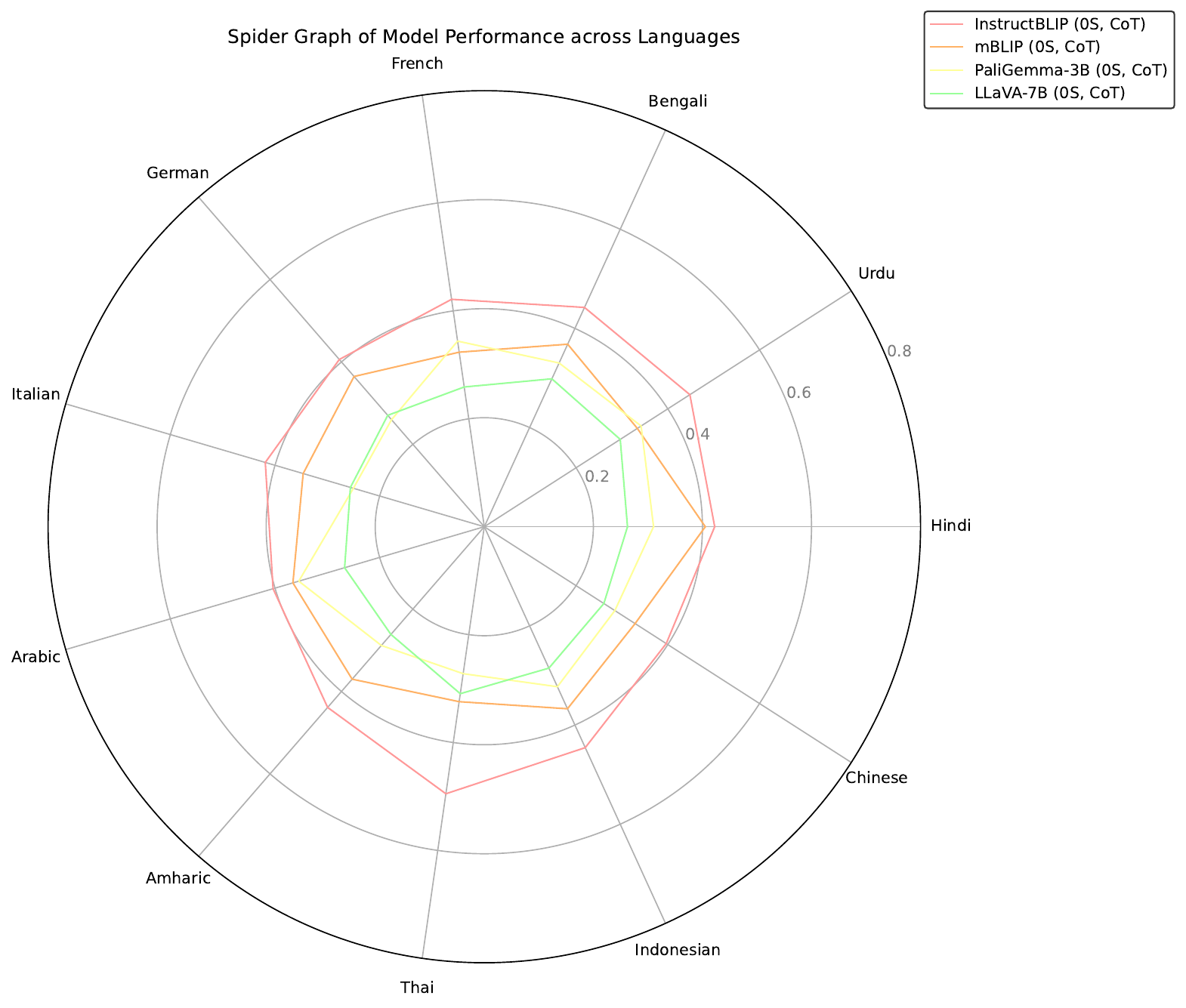}
        \caption{Few-shot-based results of MLLMs across languages}
        \label{fig:lang_resultsMLLm}
    \end{subfigure}
    \hfill
    \begin{subfigure}[b]{0.32\textwidth}
        \centering
        \includegraphics[width=\textwidth]{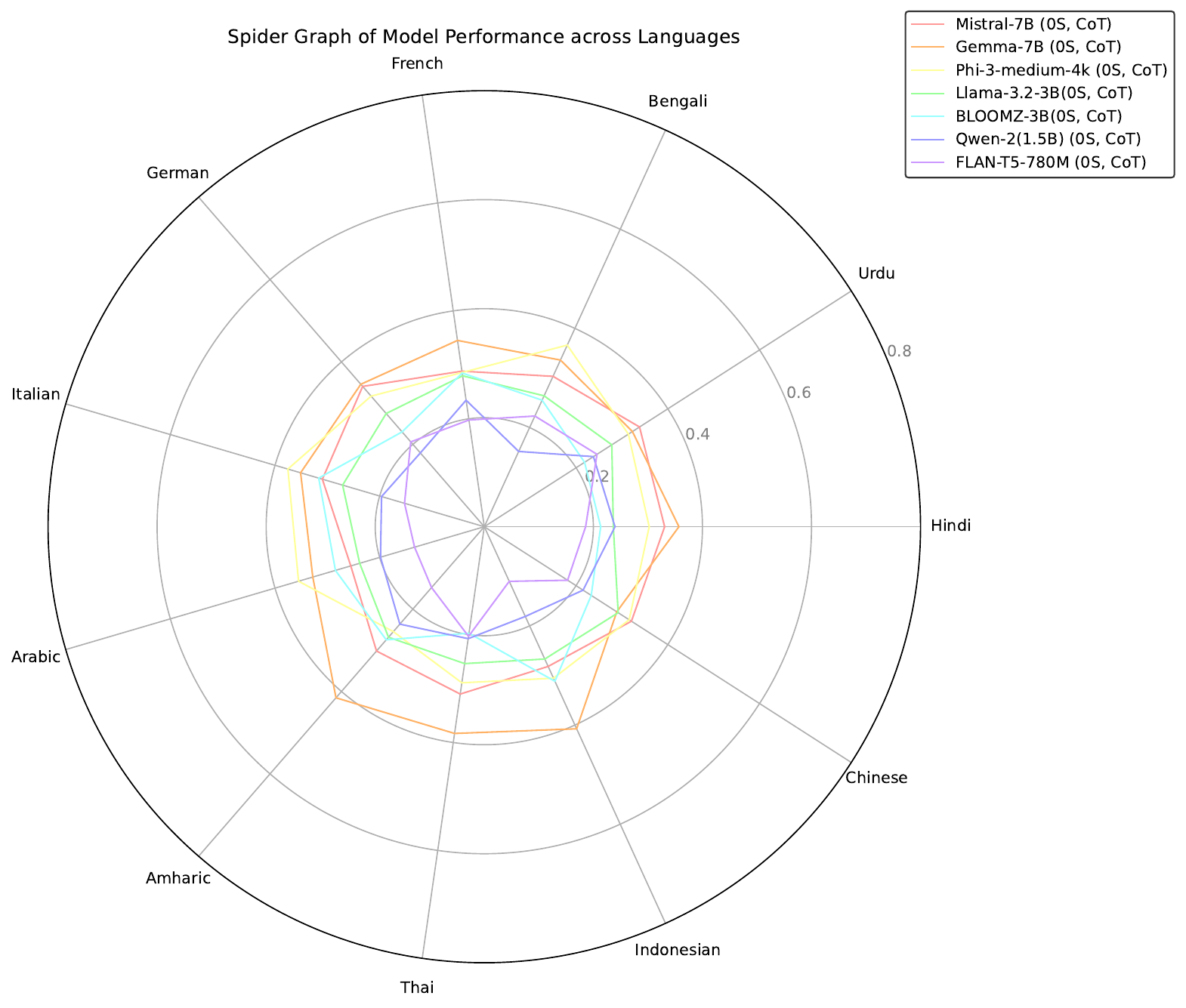}
        \caption{Few-shot-based Results of SLMs across languages}
        \label{fig:lzero_cot_slm}
    \end{subfigure}
    \caption{Few-shot-based results of models across languages\label{fig:lang_resultsLLMs}}
\end{figure*}

\begin{figure*}[hbt!]
    \centering
    \begin{subfigure}[b]{0.32\textwidth}
        \centering
        \includegraphics[width=\textwidth]{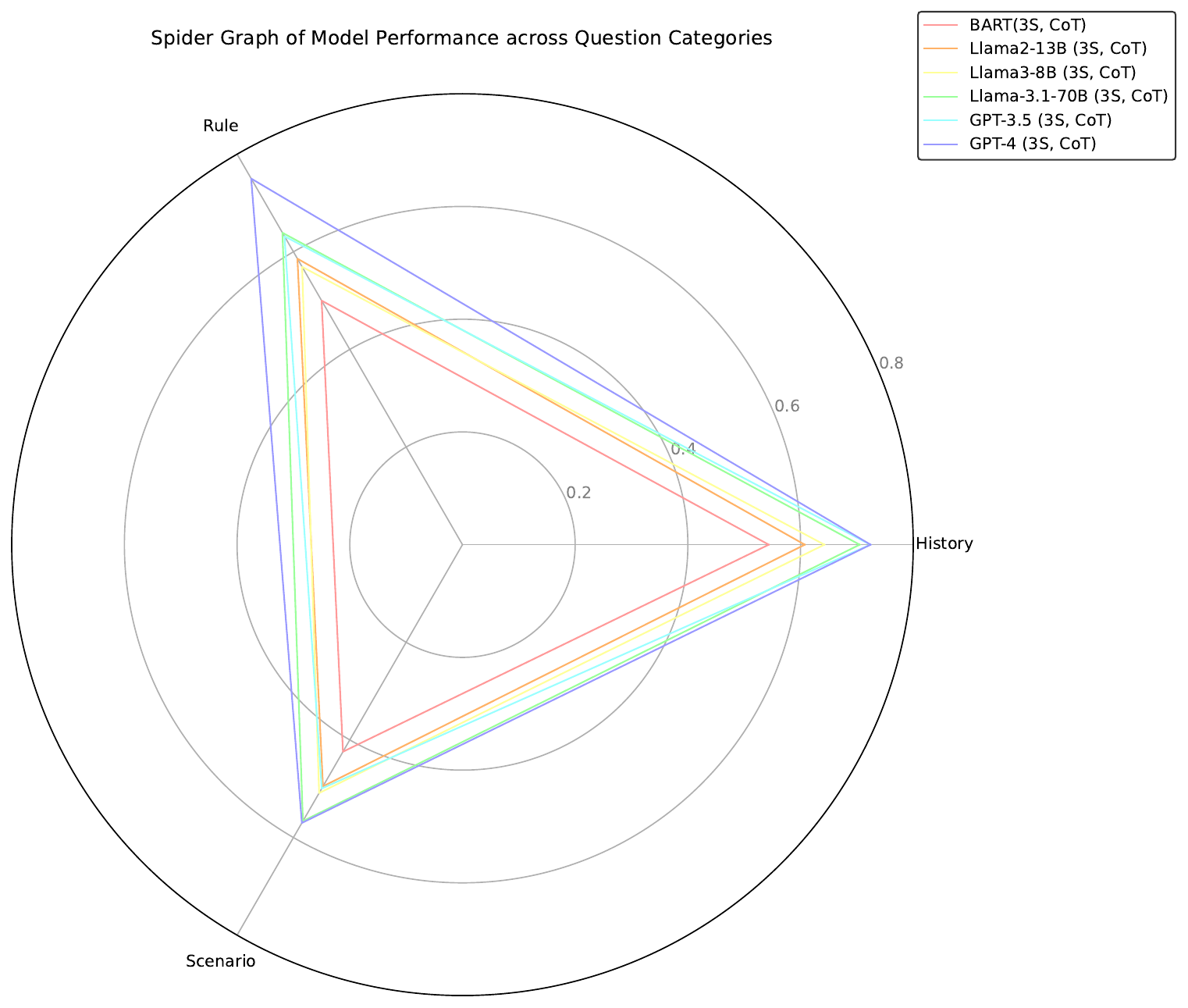} 
        \caption{CoT-based results of LLMs across types}
    \end{subfigure}
    \hfill
    \begin{subfigure}[b]{0.32\textwidth}
        \centering
        \includegraphics[width=\textwidth]{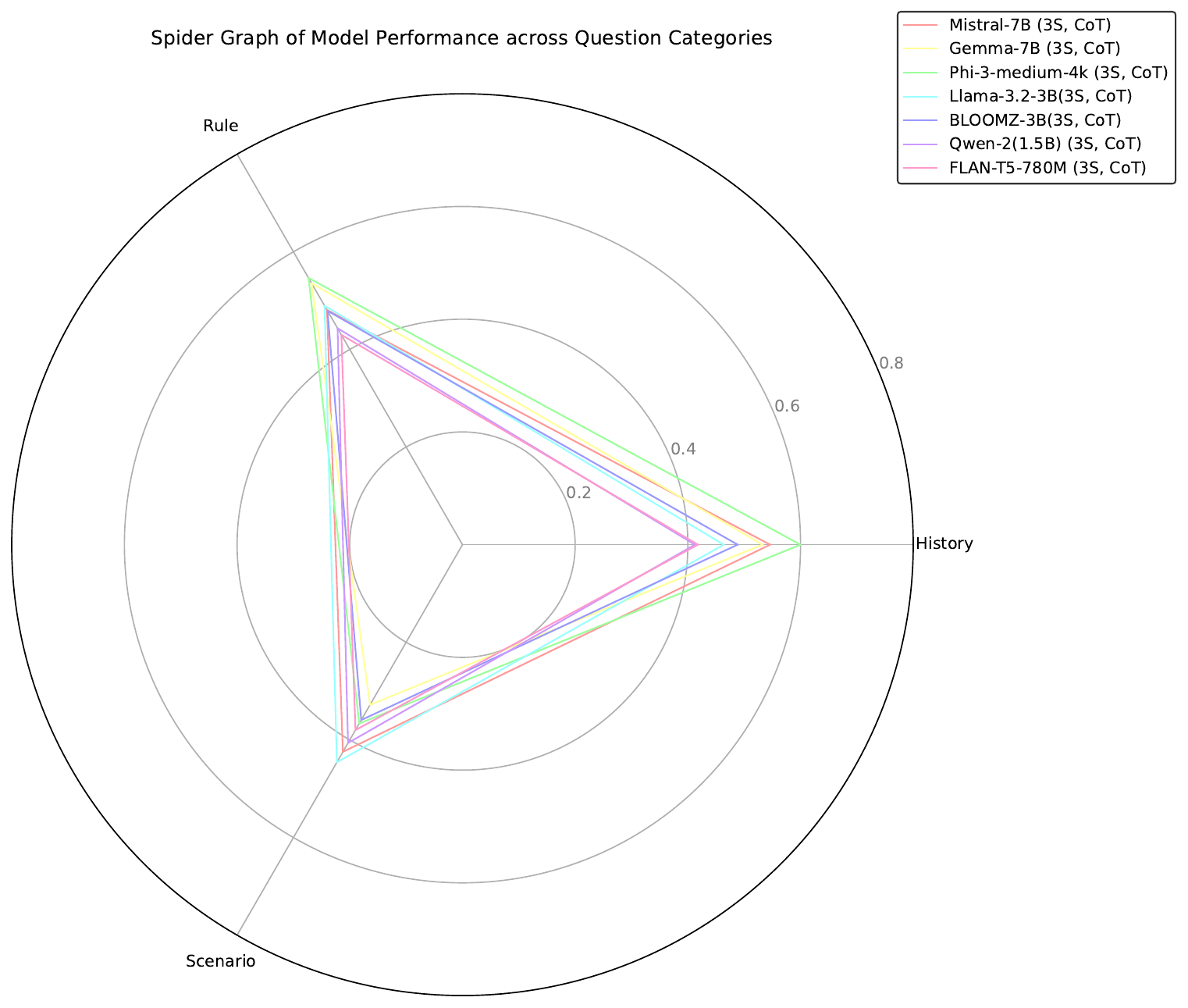} 
        \caption{CoT-based results of SLMs across types}
    \end{subfigure}
    \caption{CoT-based results of models across types}
    
\end{figure*}

\begin{figure*}[hbt!]
    \centering
    \begin{subfigure}[b]{0.32\textwidth}
        \centering
        \includegraphics[width=\textwidth]{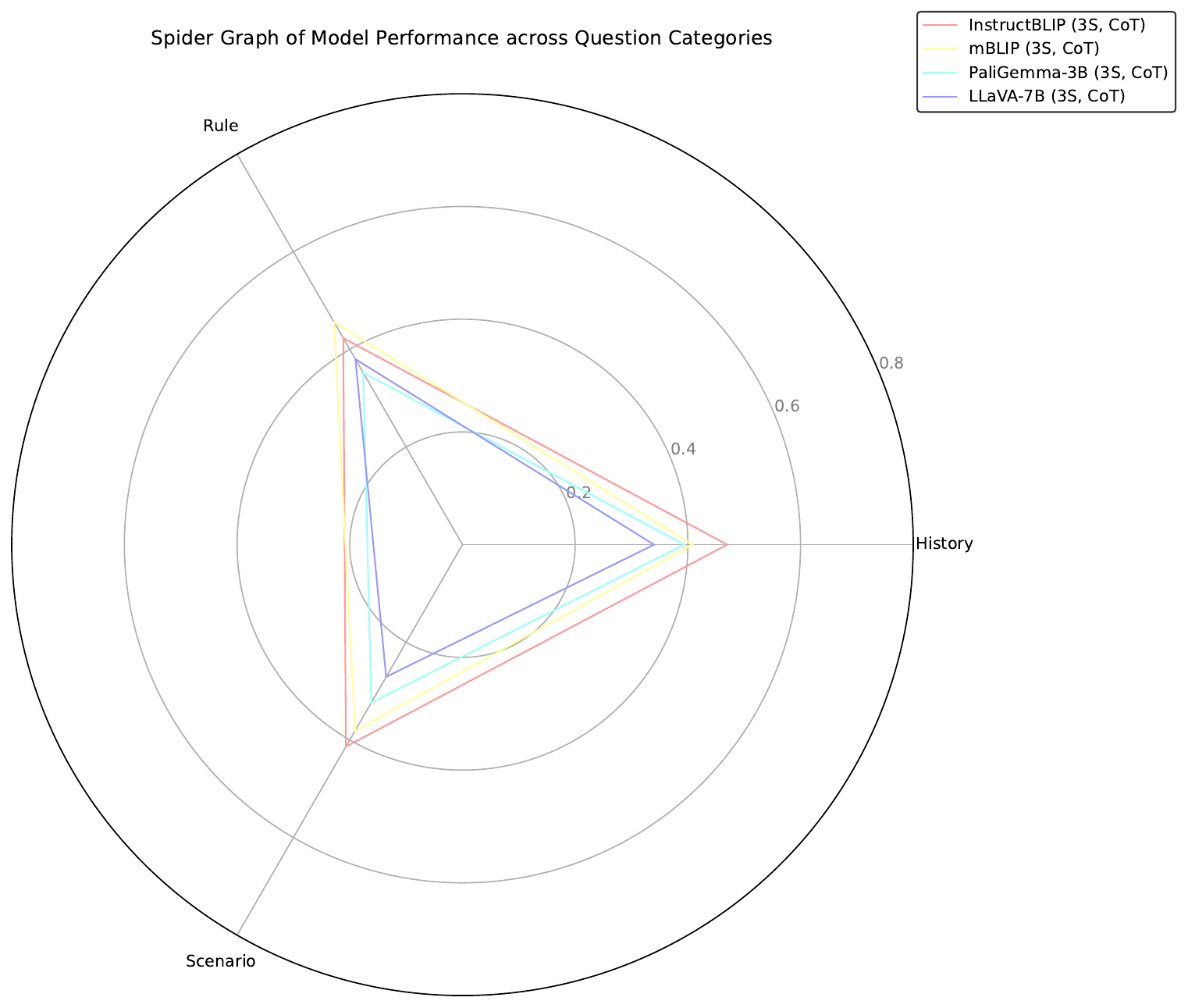}
        \caption{CoT-based results of MLLMs across types}
        \label{fig:c_3s_cot}
    \end{subfigure}
    \hfill
    \centering
     \hfill
    \begin{subfigure}[b]{0.32\textwidth}
        \centering
        \includegraphics[width=\textwidth]{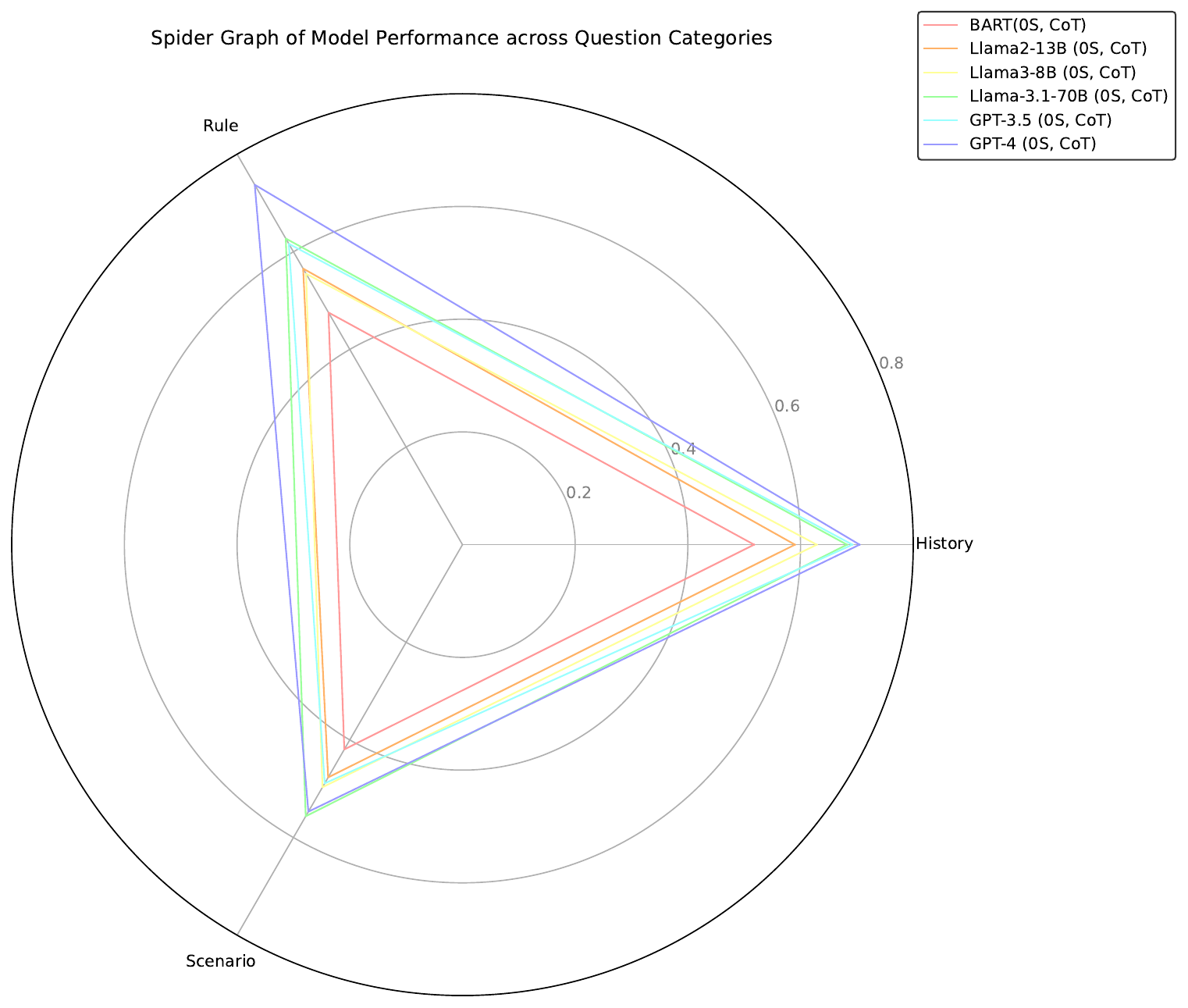}
        \caption{Few-shot based results of LLMs across types}
        \label{fig:category_zero_COt_LLm}
    \end{subfigure}
    \caption{Results of models across types
}
\end{figure*}

\begin{figure*}[hbt!]
    \centering
    \begin{subfigure}[b]{0.32\textwidth}
        \centering
        \includegraphics[width=\textwidth]{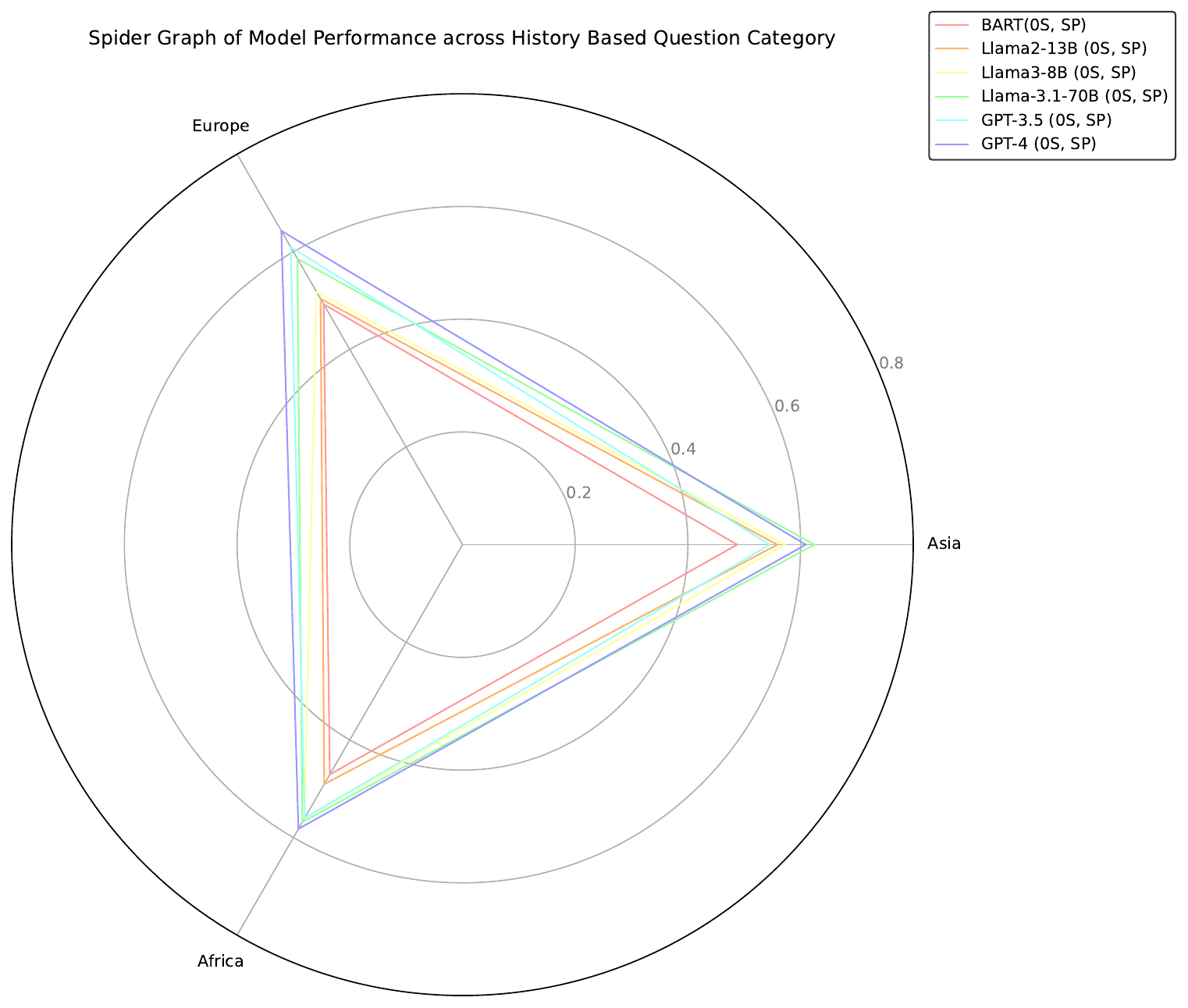}
        \caption{Zero-shot-based results of LLMs across continents}
        
    \end{subfigure}
    \hfill
    \begin{subfigure}[b]{0.32\textwidth}
        \centering
        \includegraphics[width=\textwidth]{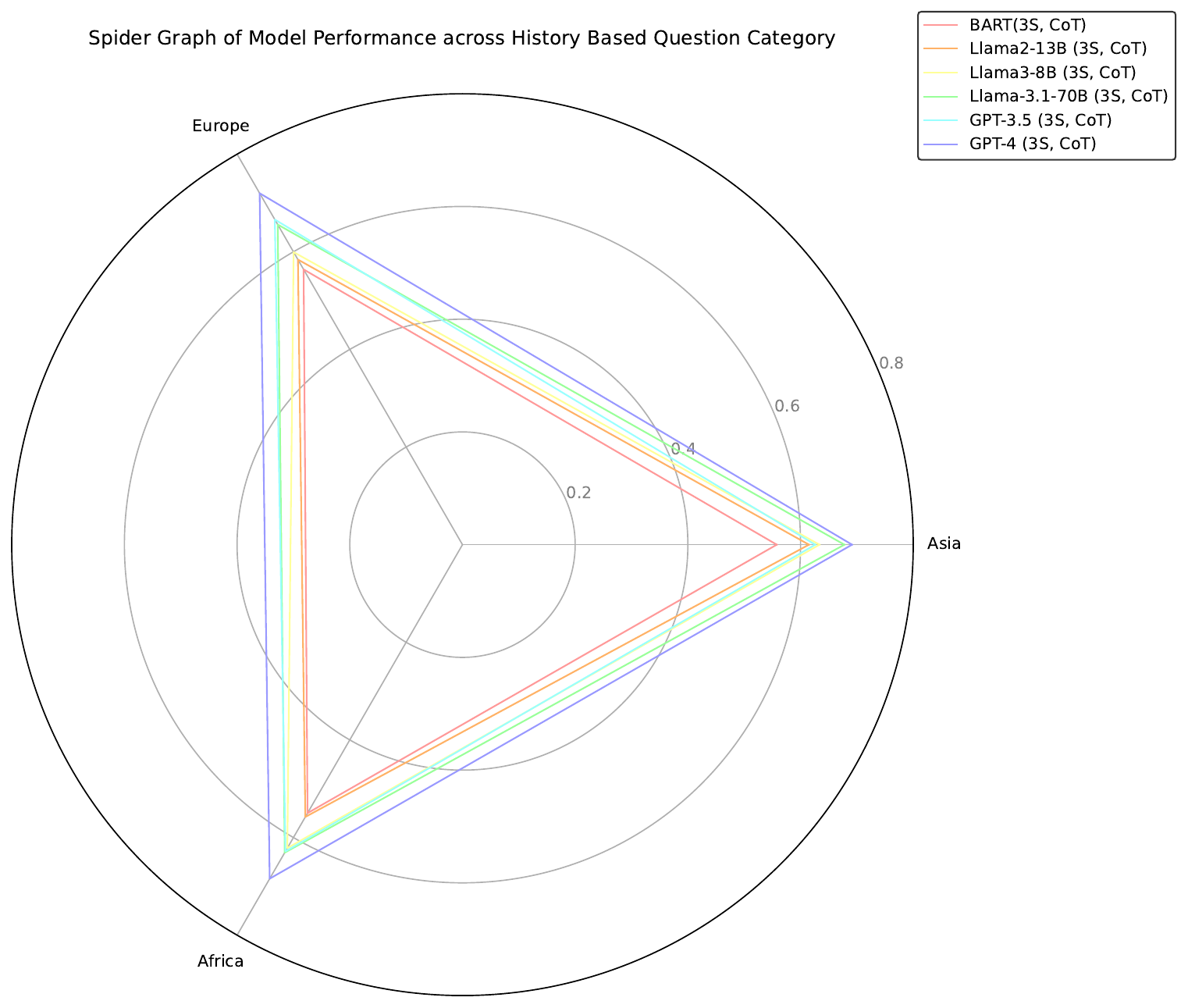}
        \caption{CoT-based results of LLMs across continents}
        
    \end{subfigure}
    \hfill
    \begin{subfigure}[b]{0.32\textwidth}
        \centering
        \includegraphics[width=\textwidth]{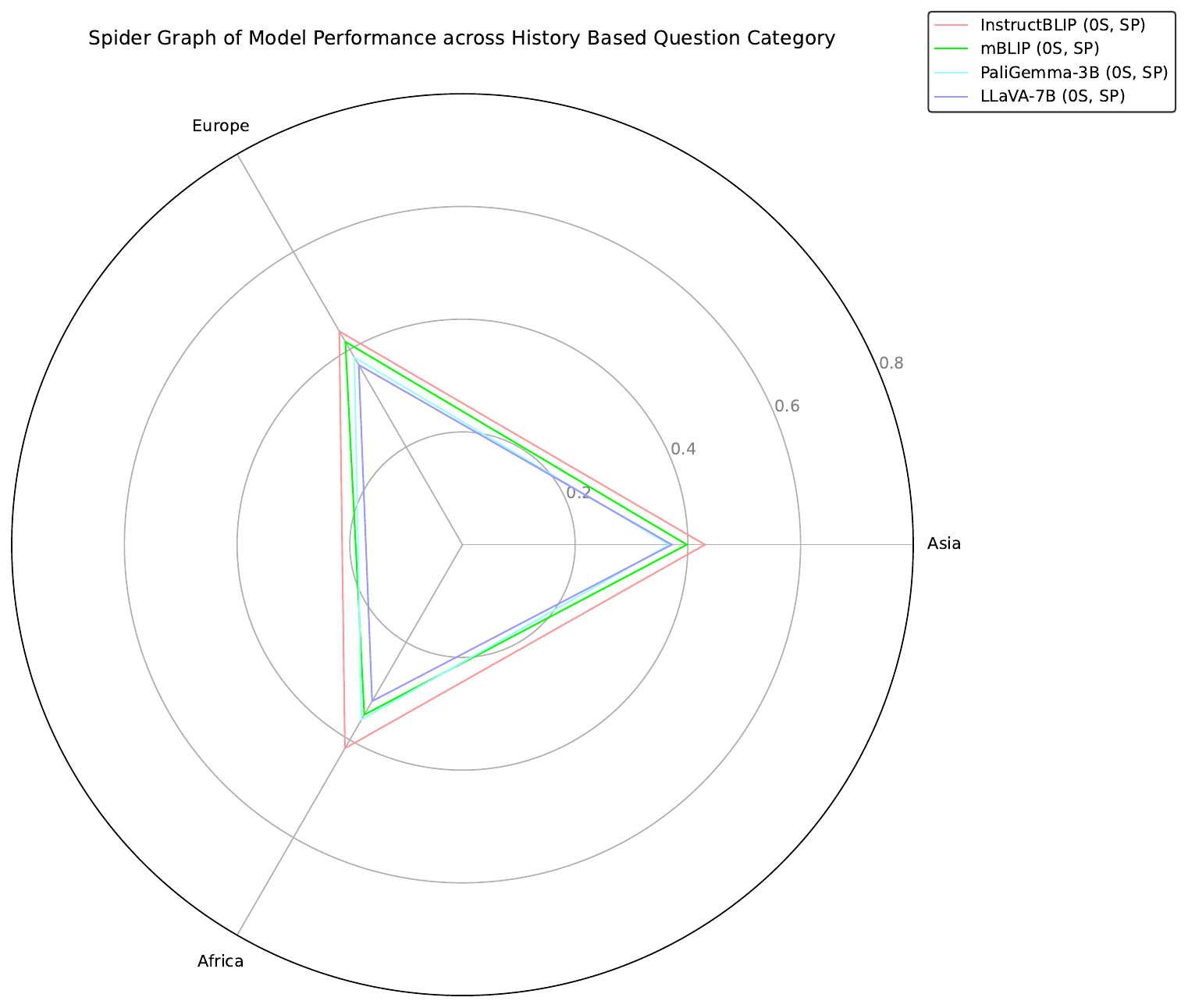}
        \caption{Zero-shot-based results of MLLMs across continents}
        
    \end{subfigure}
    \caption{Results across continents}
\end{figure*}

\begin{figure*}[hbt!]
    \centering
    \begin{subfigure}[b]{0.32\textwidth}
        \centering
        \includegraphics[width=\textwidth]{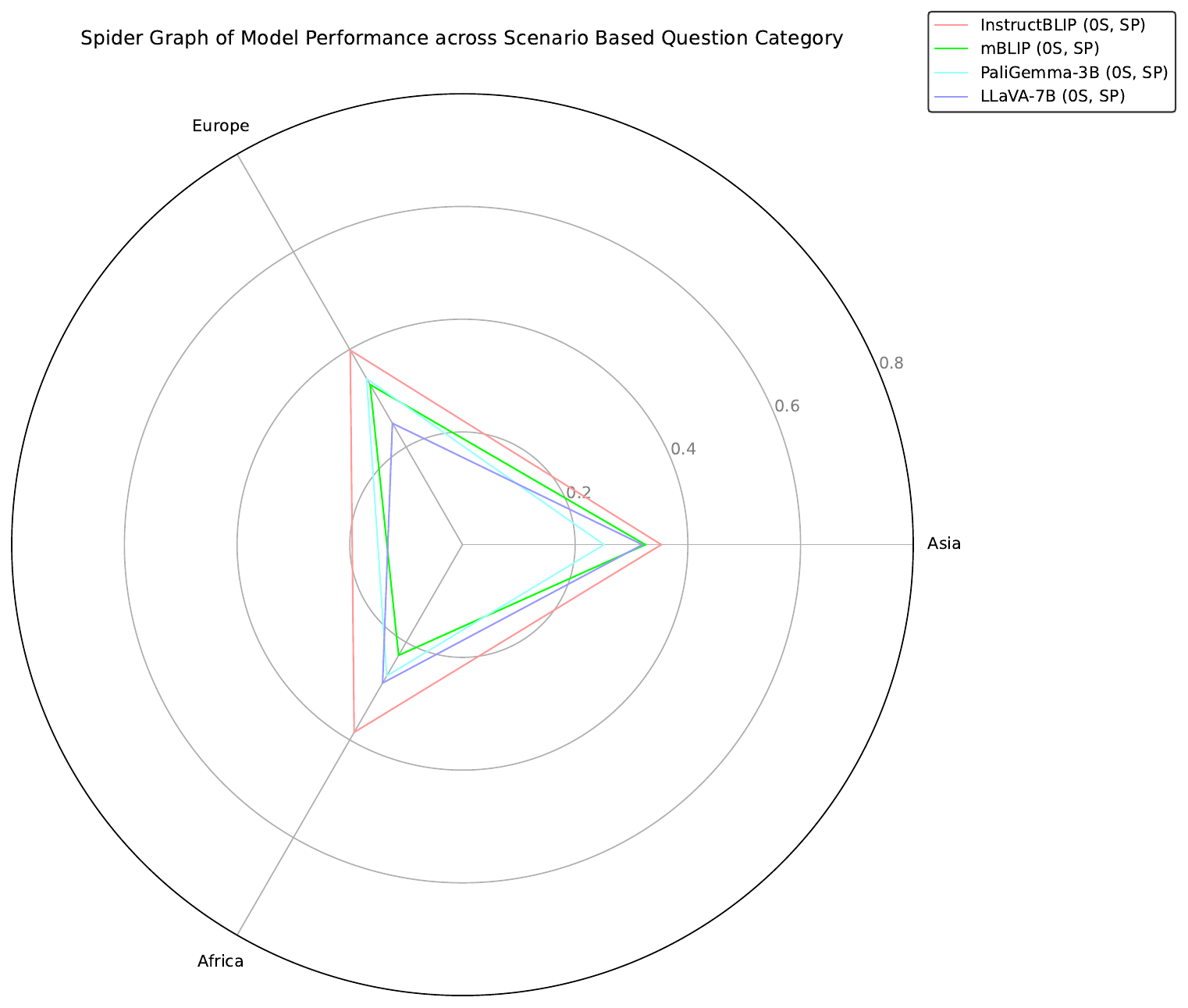}
        \caption{Zero-shot-based results of  MLLMs across continents}
        \label{fig:Comtient_Rule_ZERO_VLM}
    \end{subfigure}
    \hfill
    \begin{subfigure}[b]{0.32\textwidth}
        \centering
        \includegraphics[width=\textwidth]{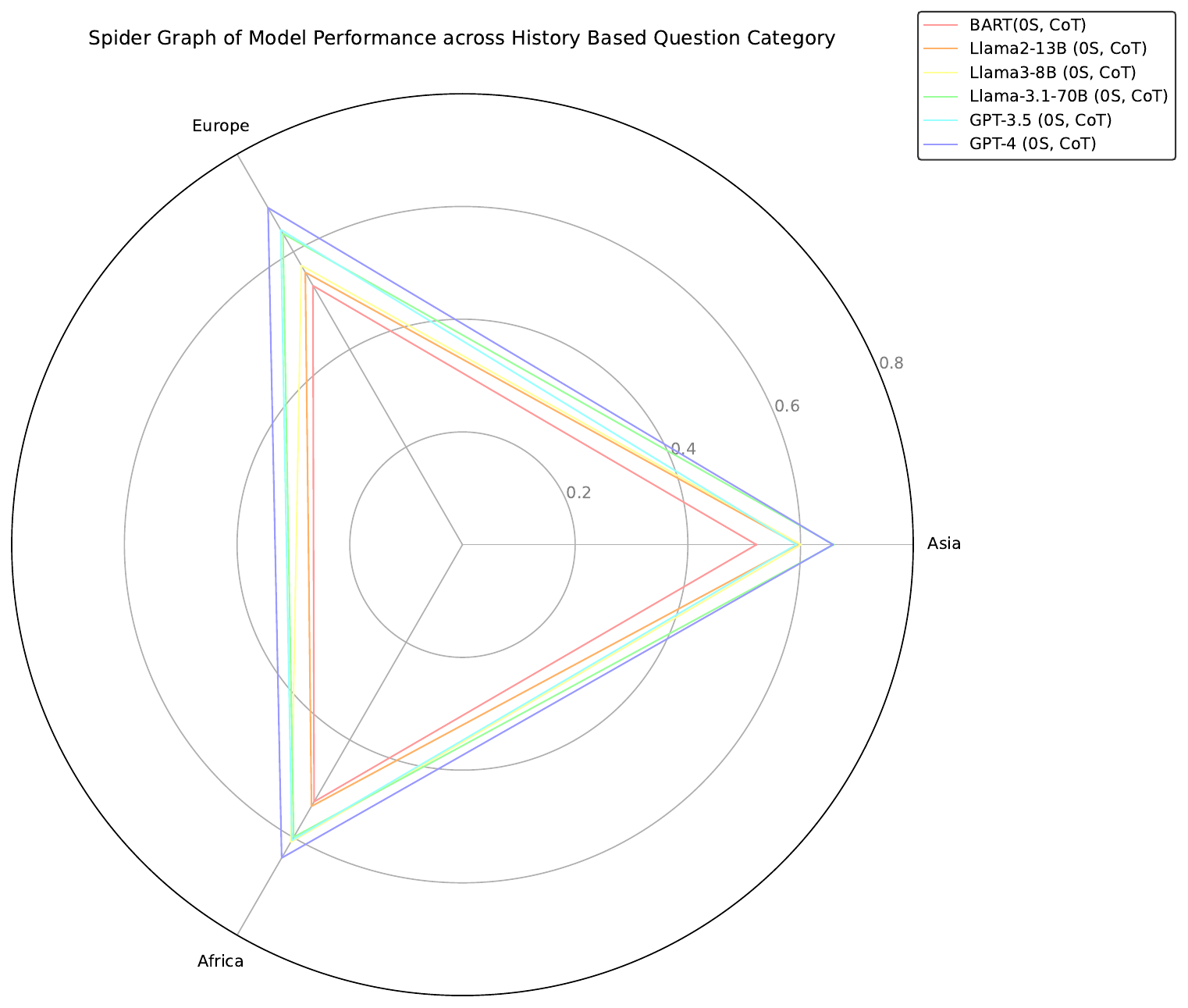}
        \caption{Few-shot-based results of LLMs across continents}
        \label{fig:Continent_Zero_COT_LLM}
    \end{subfigure}
    \hfill
    \begin{subfigure}[b]{0.32\textwidth}
        \centering
        \includegraphics[width=\textwidth]{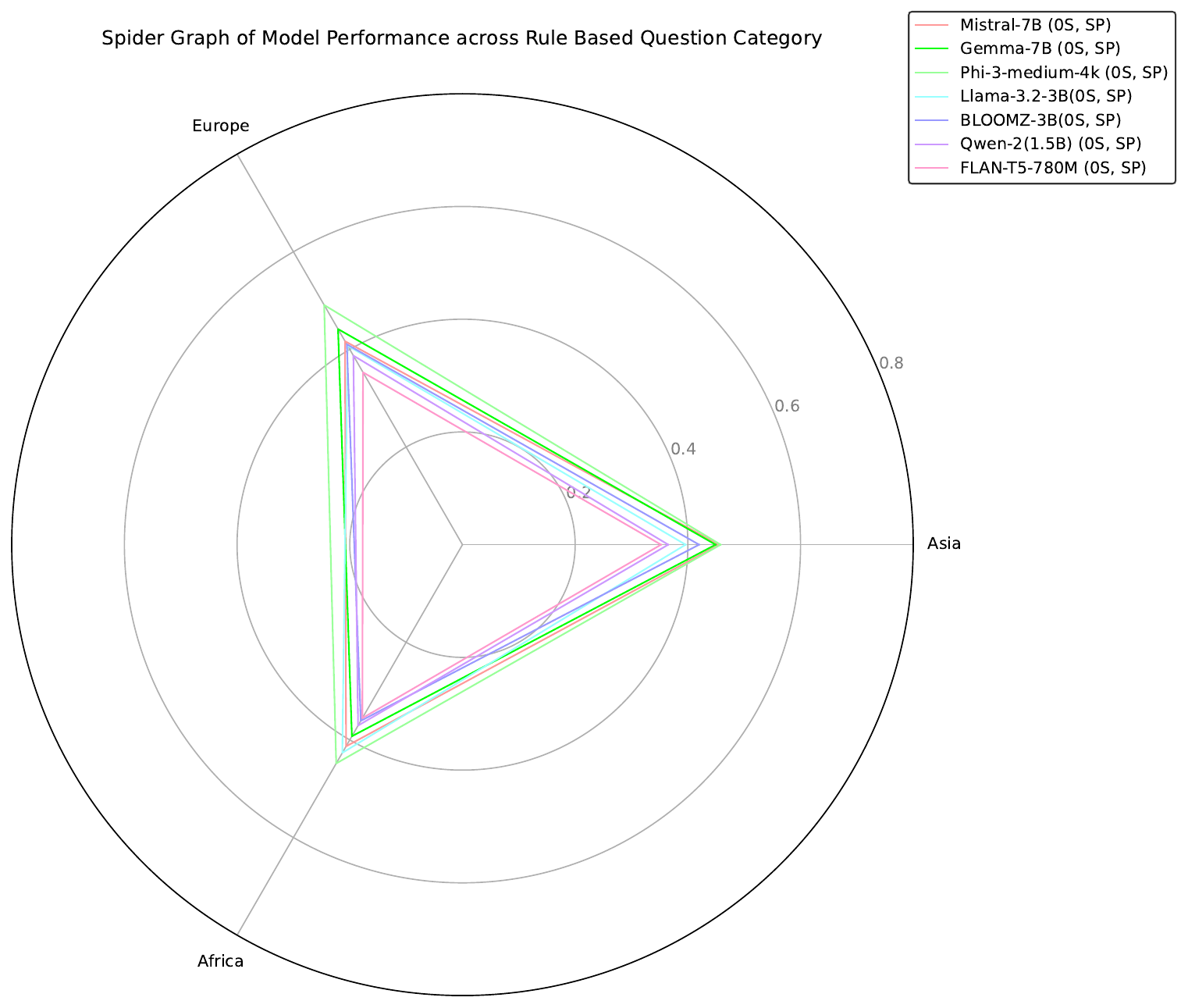}
        \caption{Zero-shot-based results of LLMs across continents}
        \label{fig:Continent_Rule_Zero_SLM}
    \end{subfigure}
    \caption{Results of models across continents}
    \label{fig:Continent_Rule_Zero_results}
\end{figure*}

\begin{figure*}[hbt!]
    \centering
    \begin{subfigure}[b]{0.32\textwidth}
        \centering
        \includegraphics[width=\textwidth]{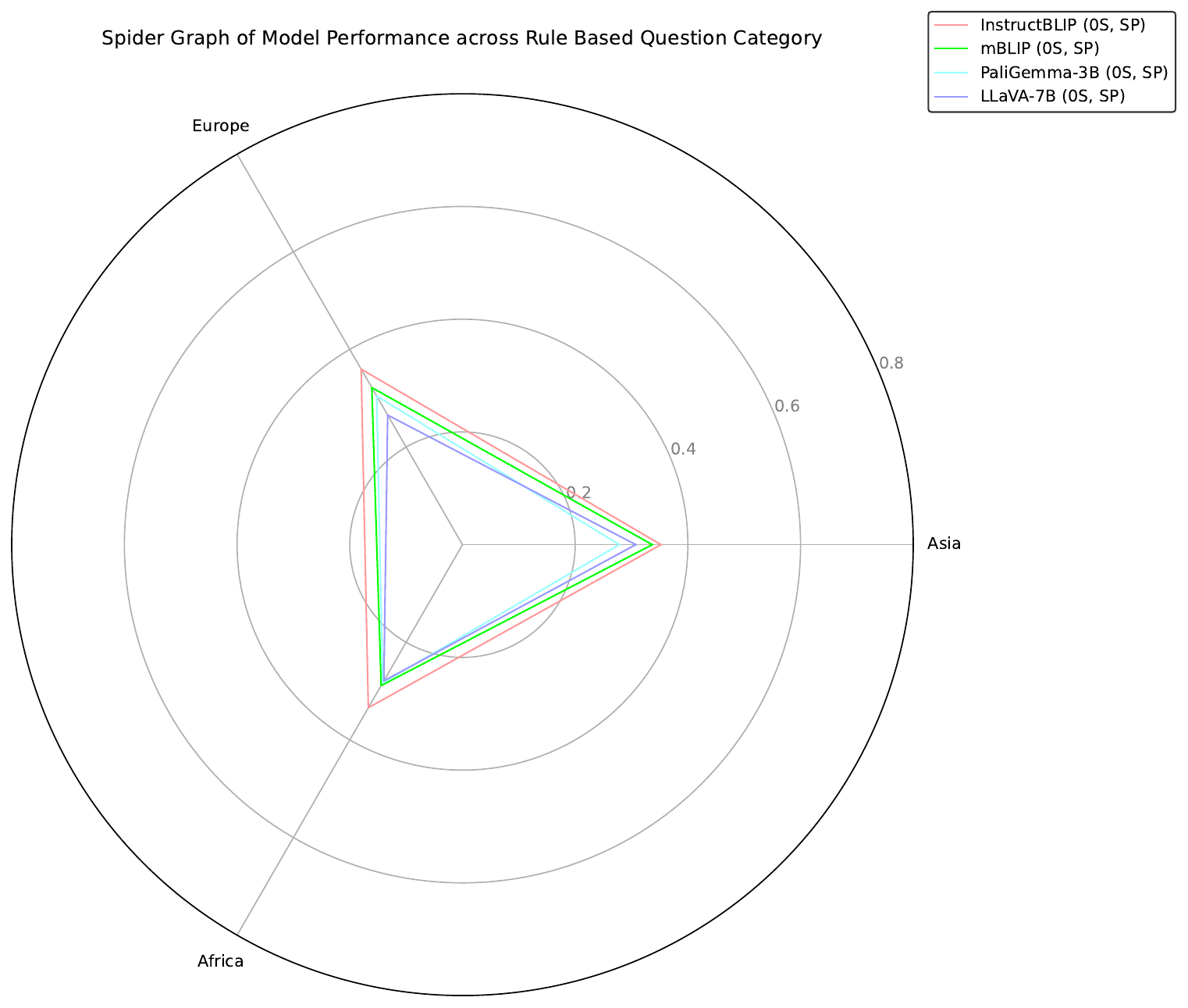}
        \caption{Zero-shot-based results of  MLLMs across continents}
        \label{fig:Continent_Rule_Zero_VLM}
    \end{subfigure}
    \hfill
    \begin{subfigure}[b]{0.32\textwidth}
        \centering
        \includegraphics[width=\textwidth]{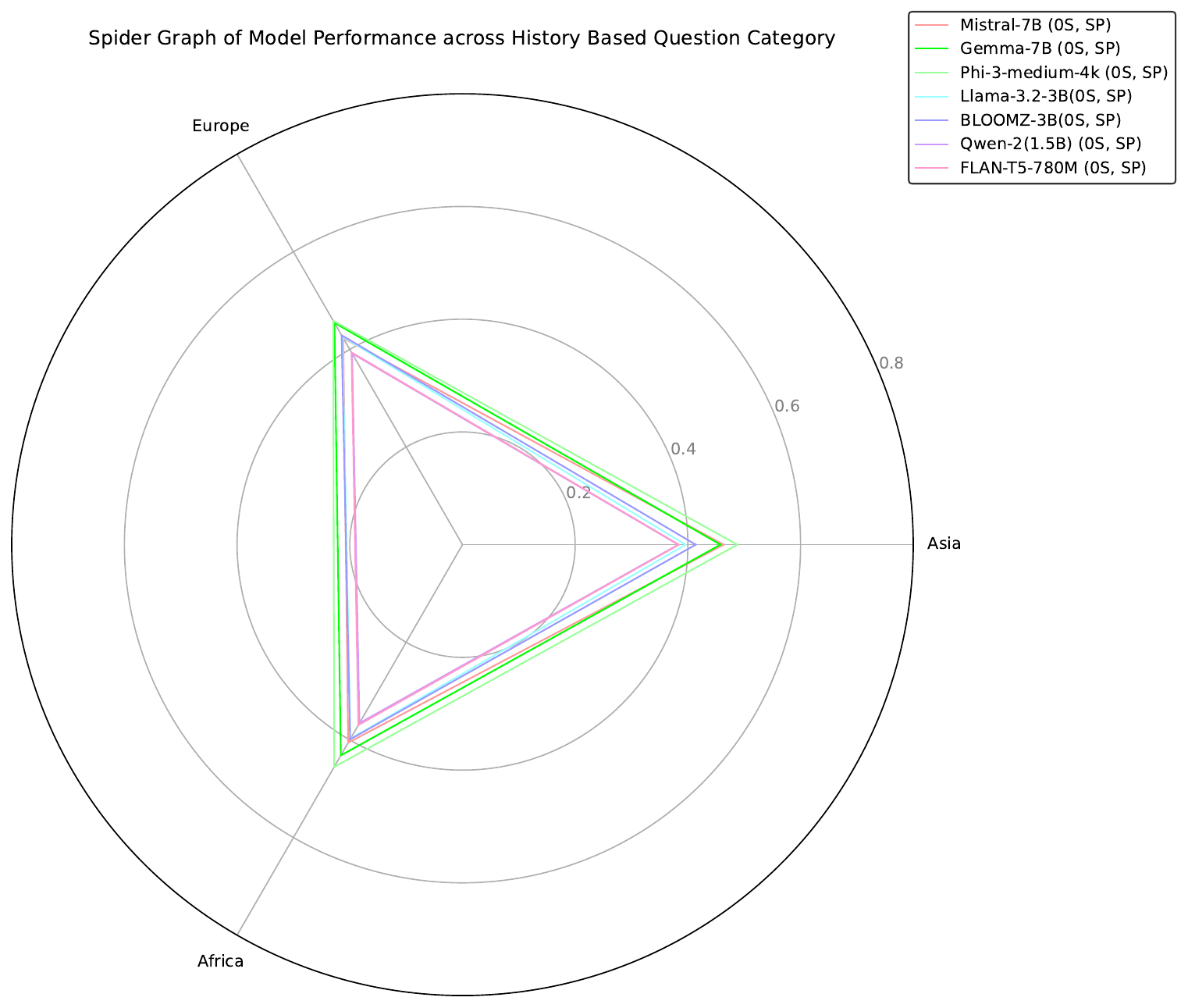}
        \caption{Zero-shot-based results of SLMs across continents}
        \label{fig:Continent_HistroyZERO_SLM}
    \end{subfigure}
    \hfill
    \begin{subfigure}[b]{0.32\textwidth}
        \centering
        \includegraphics[width=\textwidth]{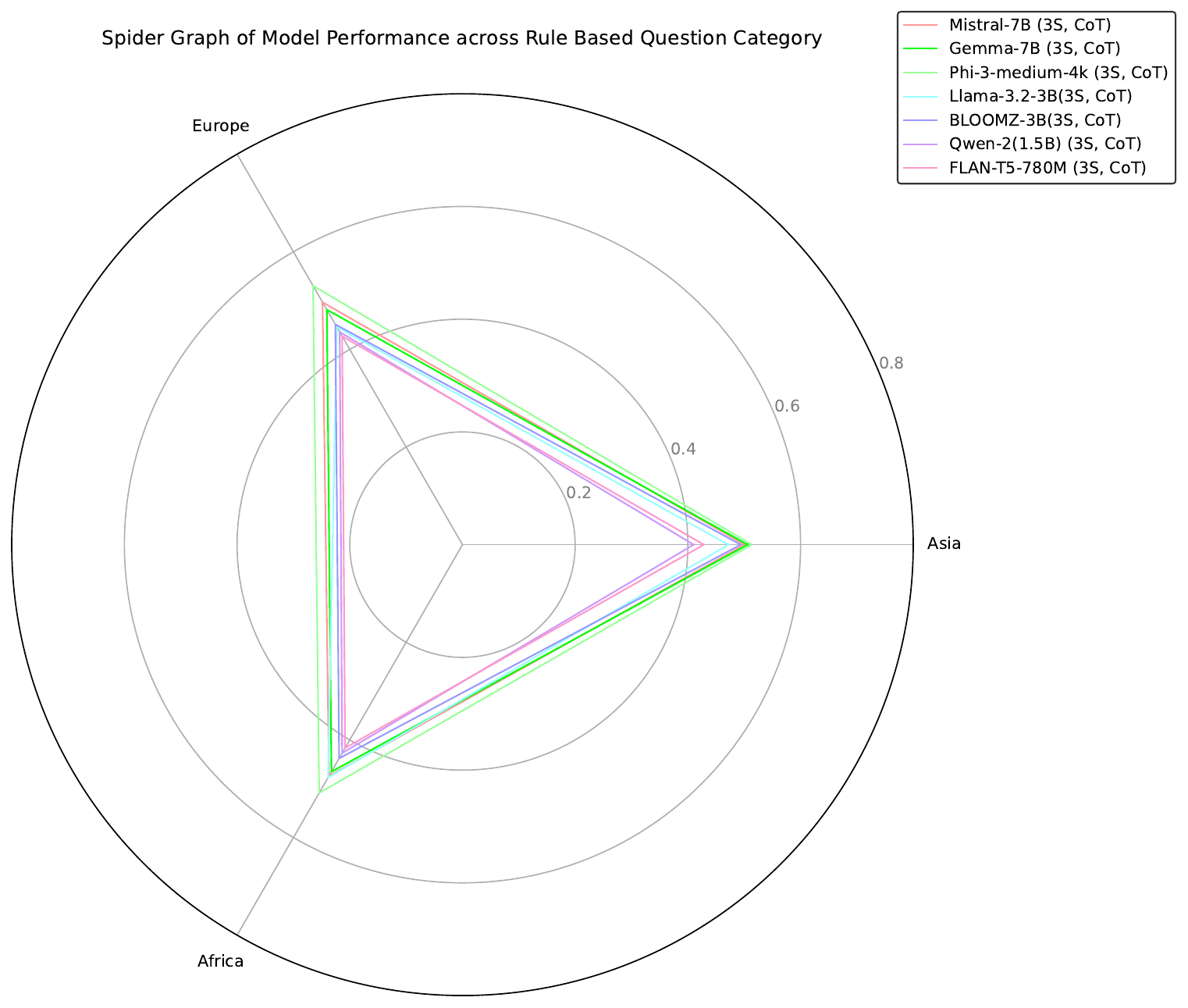}
        \caption{CoT-based results of SLMs across continents}
        \label{fig:Continent_rule_3S_COT_SLM}
    \end{subfigure}
    \caption{Results of models across continents}
    
    \label{fig:Continent_rule_3S_results}
\end{figure*}

\begin{figure*}[hbt!]
    \centering
    \begin{subfigure}[b]{0.32\textwidth}
        \centering
        \includegraphics[width=\textwidth]{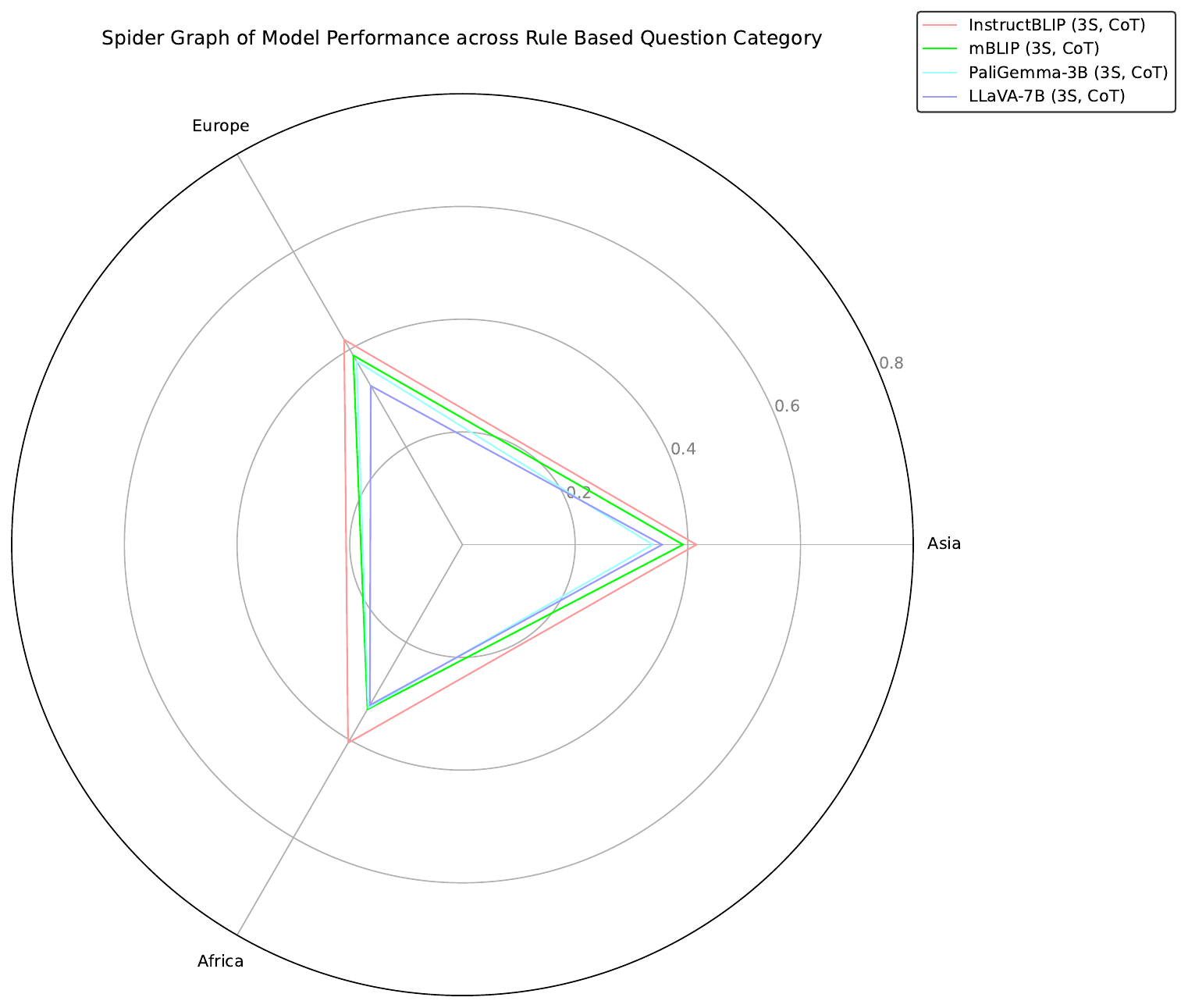}
        \caption{CoT-based results of SLMs in across continents}
        \label{fig:Continent_rule_3S_COT_VLM}
    \end{subfigure}
    \hfill
    \begin{subfigure}[b]{0.32\textwidth}
        \centering
        \includegraphics[width=\textwidth]{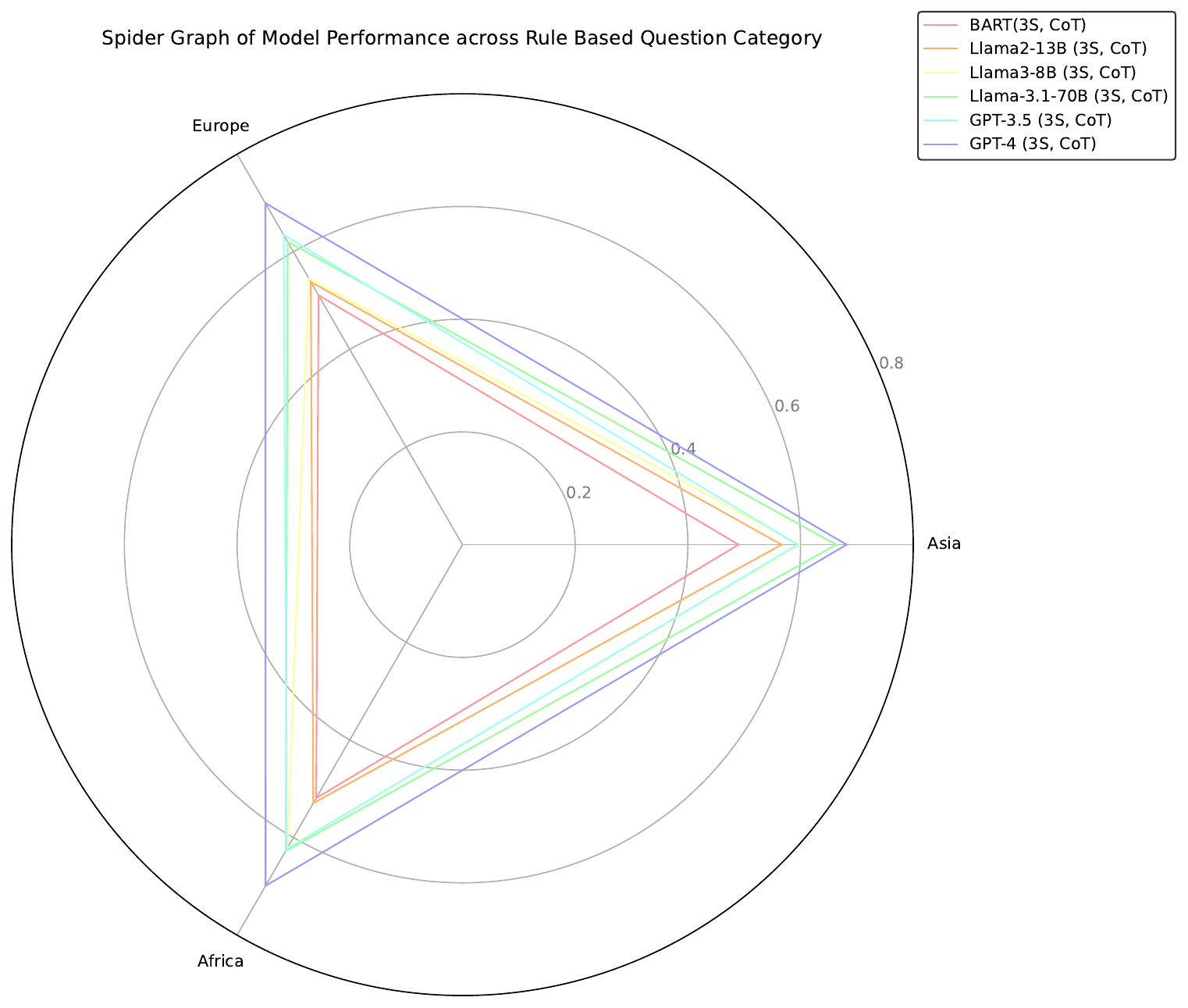}
        \caption{CoT-based results of LLMs across continents}
        \label{fig:Continent_rulee_3s-COT_LLM}
    \end{subfigure}
    \hfill
    \begin{subfigure}[b]{0.32\textwidth}
        \centering
        \includegraphics[width=\textwidth]{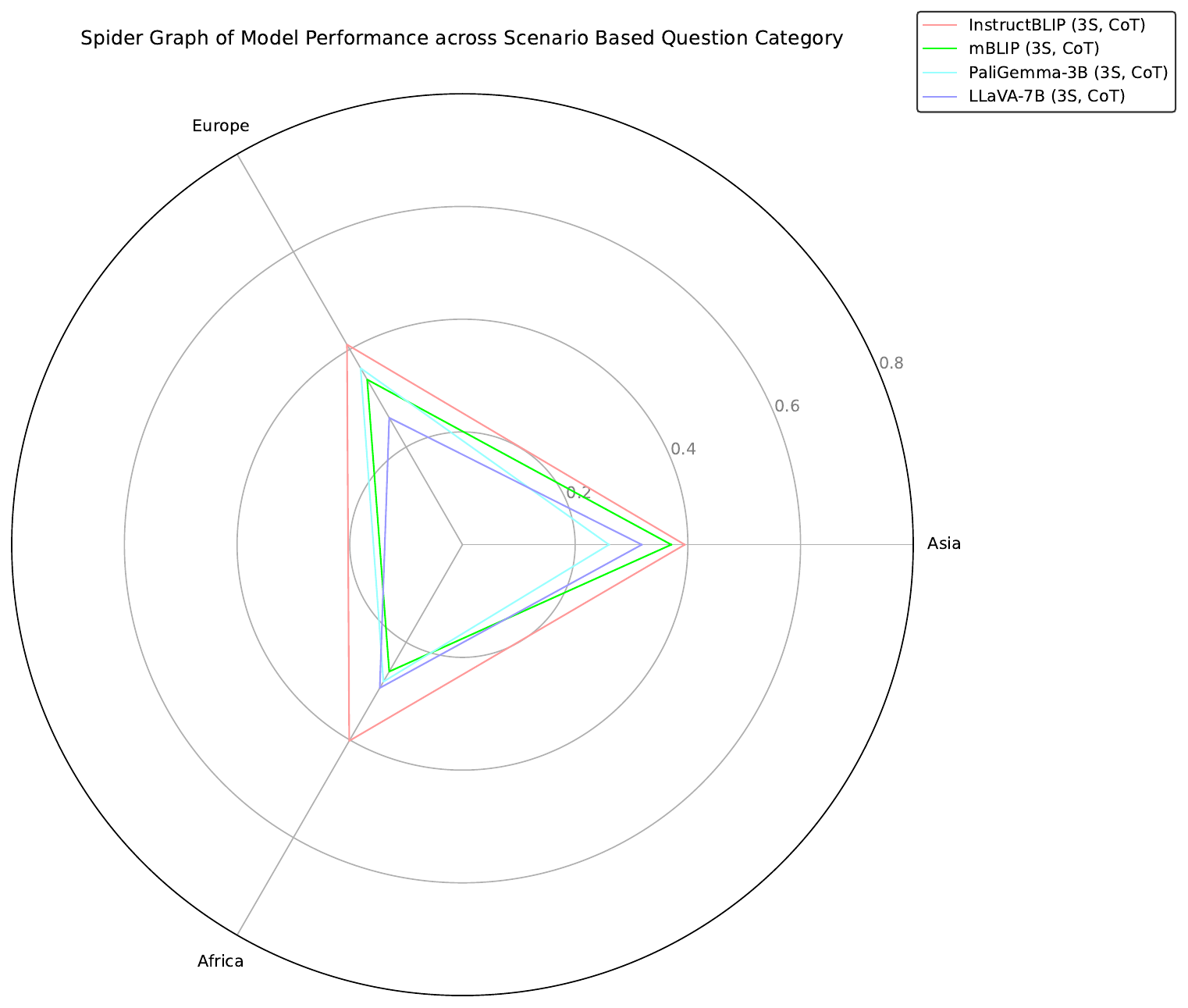}
        \caption{CoT-based results of MLLMs across continents}
        \label{fig:Contient_Scenario3S_COT_VLM}
    \end{subfigure}
    \caption{CoT-based results of models across continents}
    
    \label{fig:Contient_Scenario3S_results}
\end{figure*}

\begin{figure*}[hbt!]
    \centering
    \begin{subfigure}[b]{0.32\textwidth}
        \centering
        \includegraphics[width=\textwidth]{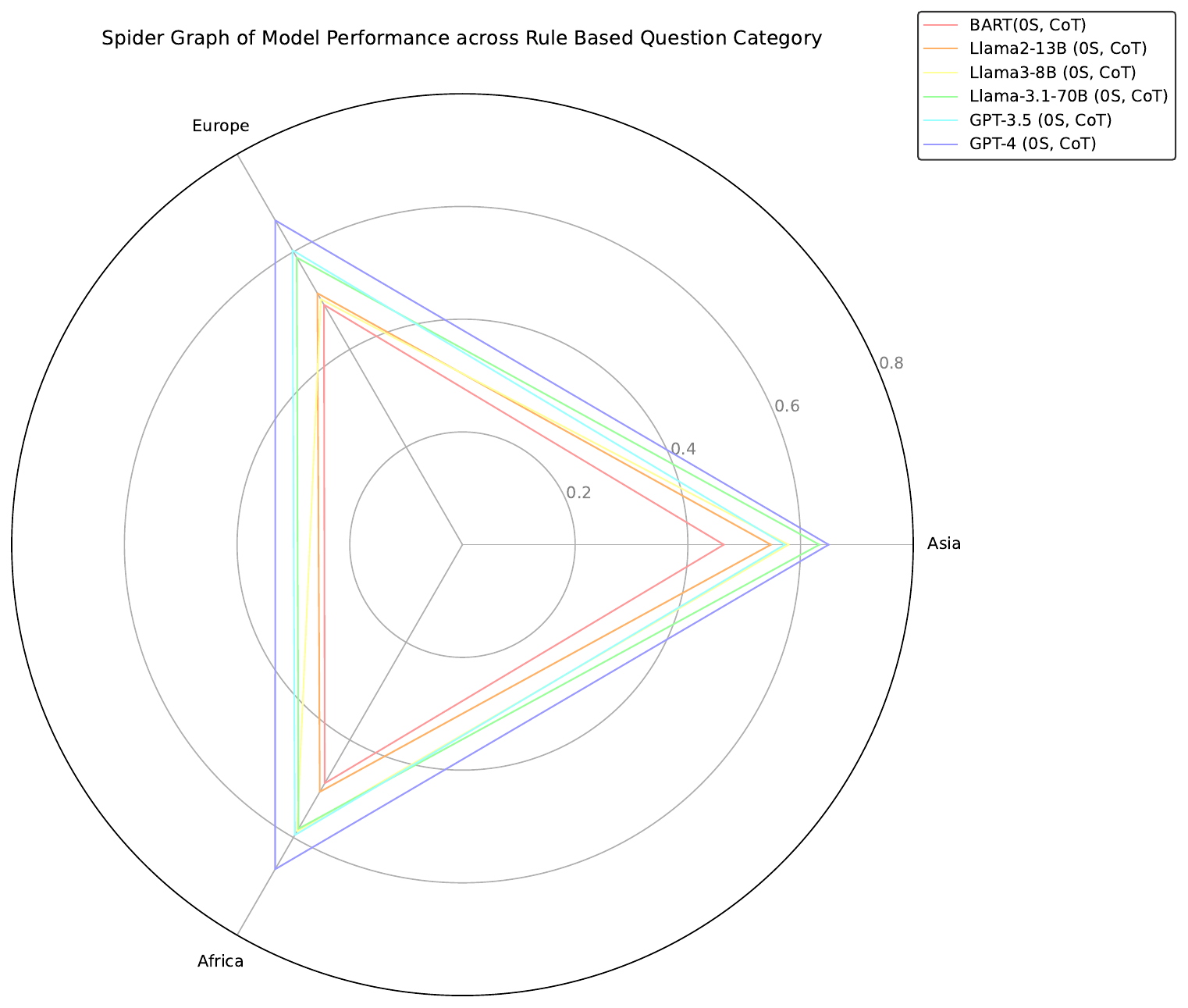}
        \caption{Zero-shot-based results of  LLMs across continents}
        \label{fig:Continent_rule_Zero_COT_LLM}
    \end{subfigure}
    \hfill
    \begin{subfigure}[b]{0.32\textwidth}
        \centering
        \includegraphics[width=\textwidth]{Continent/Continent_rulee_3s-COT_LLM.pdf}
        \caption{CoT-based results of  LLMs in across continents}
        \label{fig:Continent_rule_3s-COT_LLM}
    \end{subfigure}
    \hfill
    \begin{subfigure}[b]{0.32\textwidth}
        \centering
        \includegraphics[width=\textwidth]{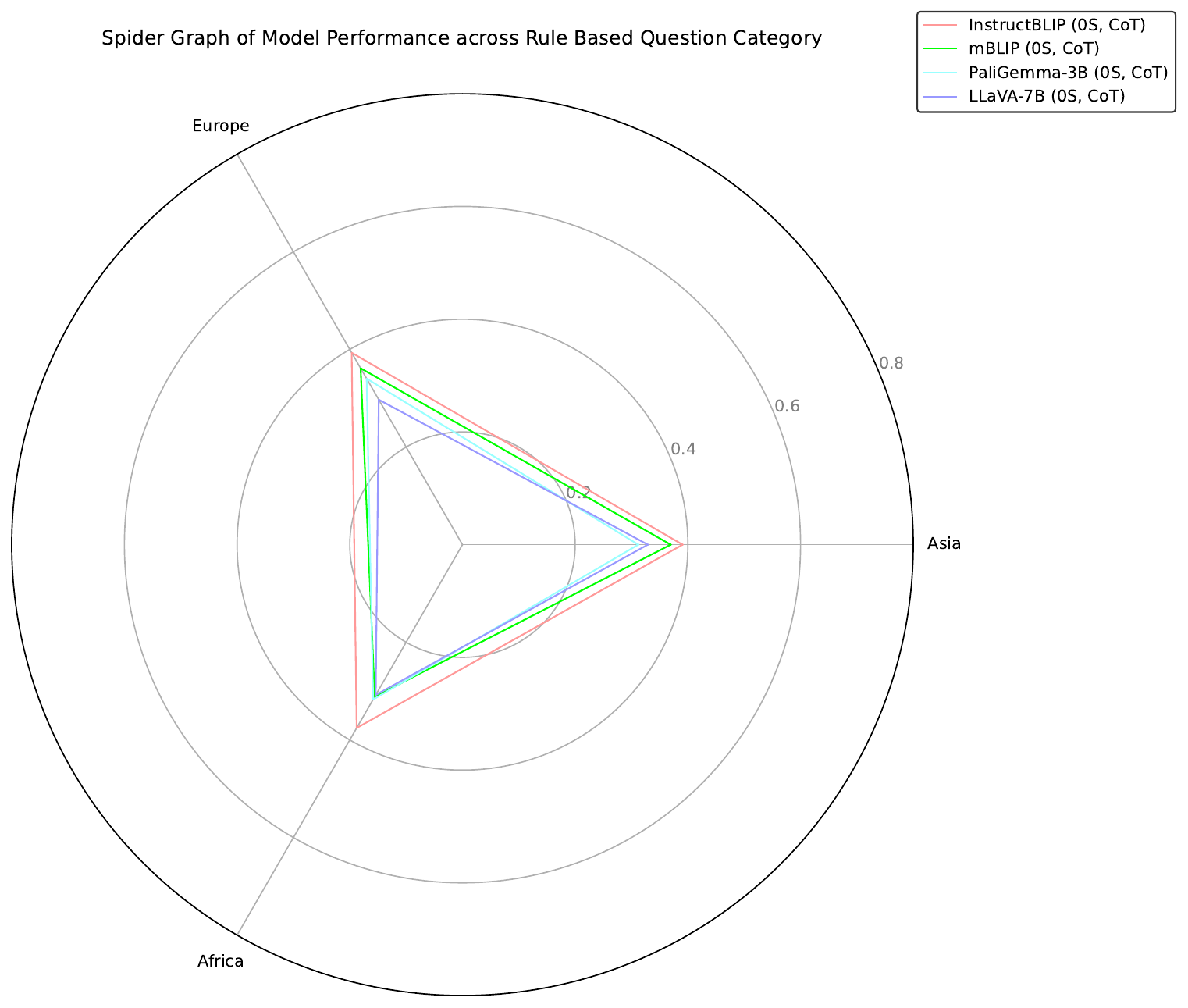}
        \caption{Zero-shot-based results of MLLMs across continents}
        \label{fig:Continent_rule_zero_COT_VLM}
    \end{subfigure}
    \caption{}
    
    \label{fig:Continent_rule_results}
\end{figure*}

\begin{figure*}[hbt!]
    \centering
    \begin{subfigure}[b]{0.32\textwidth}
        \centering
        \includegraphics[width=\textwidth]{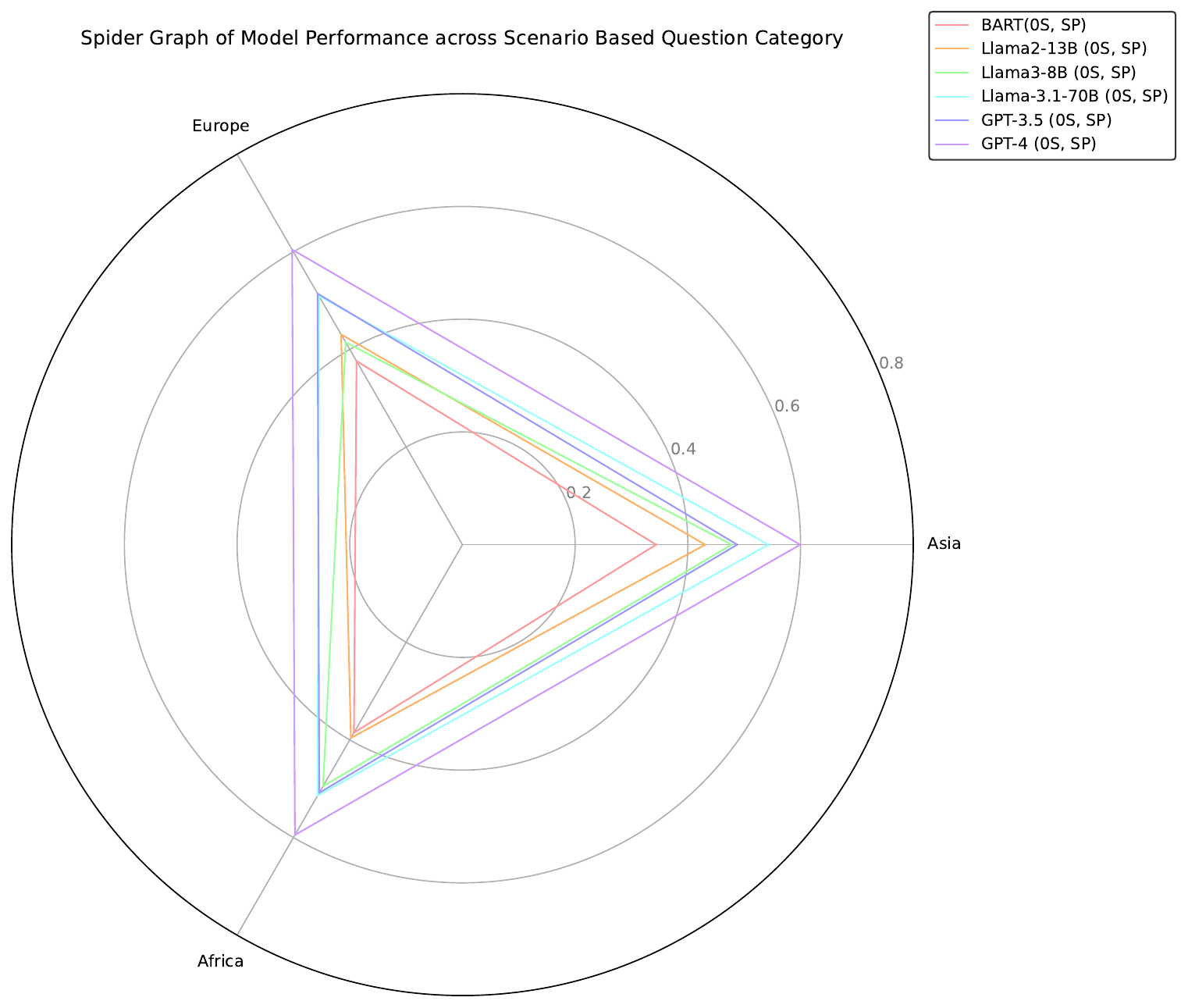}
        \caption{Zero-shot-based results of LLMs across continents}
        \label{fig:Continent_Scenario_Zero_LLM}
    \end{subfigure}
    \hfill
    \begin{subfigure}[b]{0.32\textwidth}
        \centering
        \includegraphics[width=\textwidth]{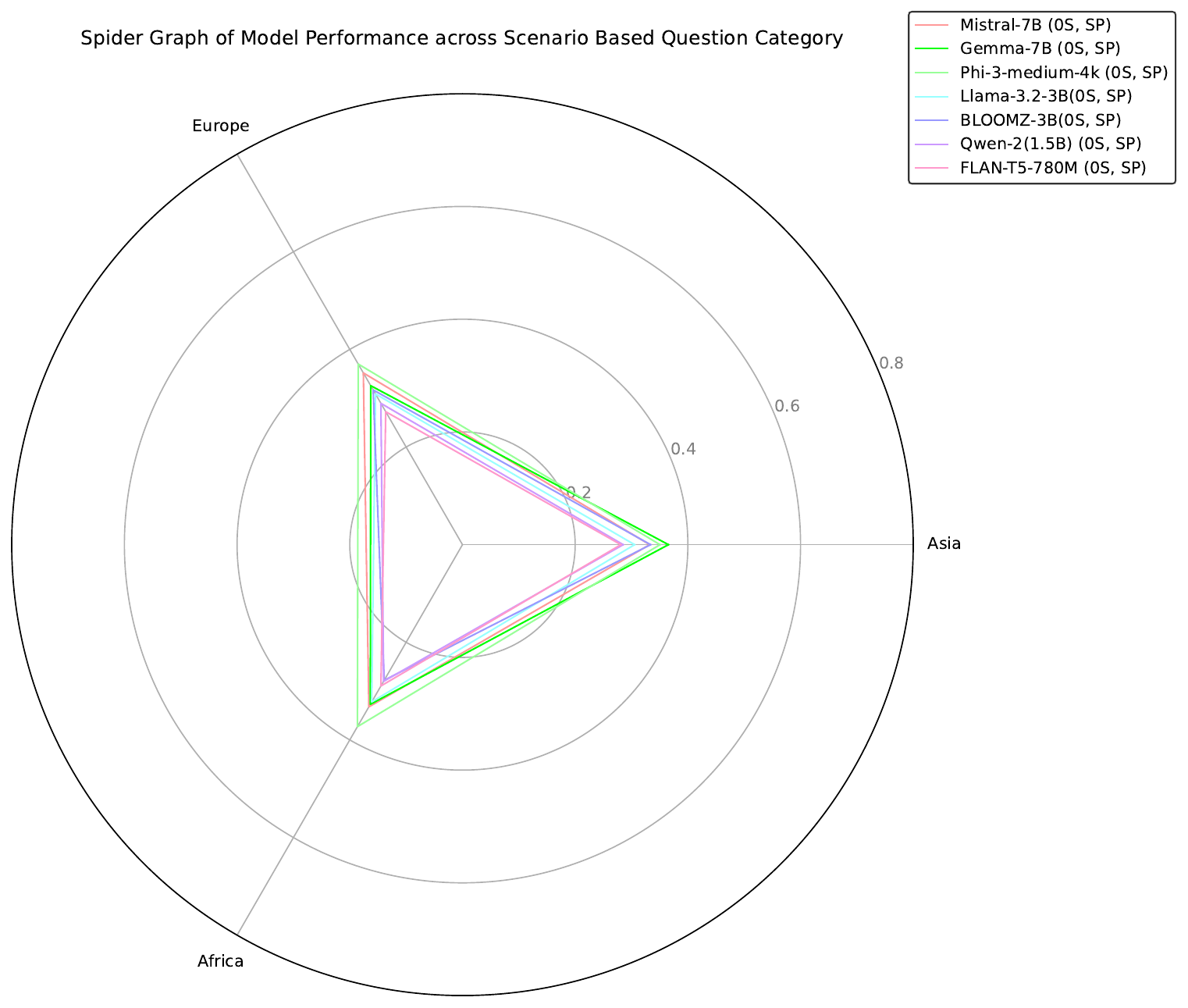}
        \caption{Zero-shot-based results of SLMs across continents}
        \label{fig:Continent_Scenario_Zero_SLM}
    \end{subfigure}
    \hfill
    \begin{subfigure}[b]{0.32\textwidth}
        \centering
        \includegraphics[width=\textwidth]{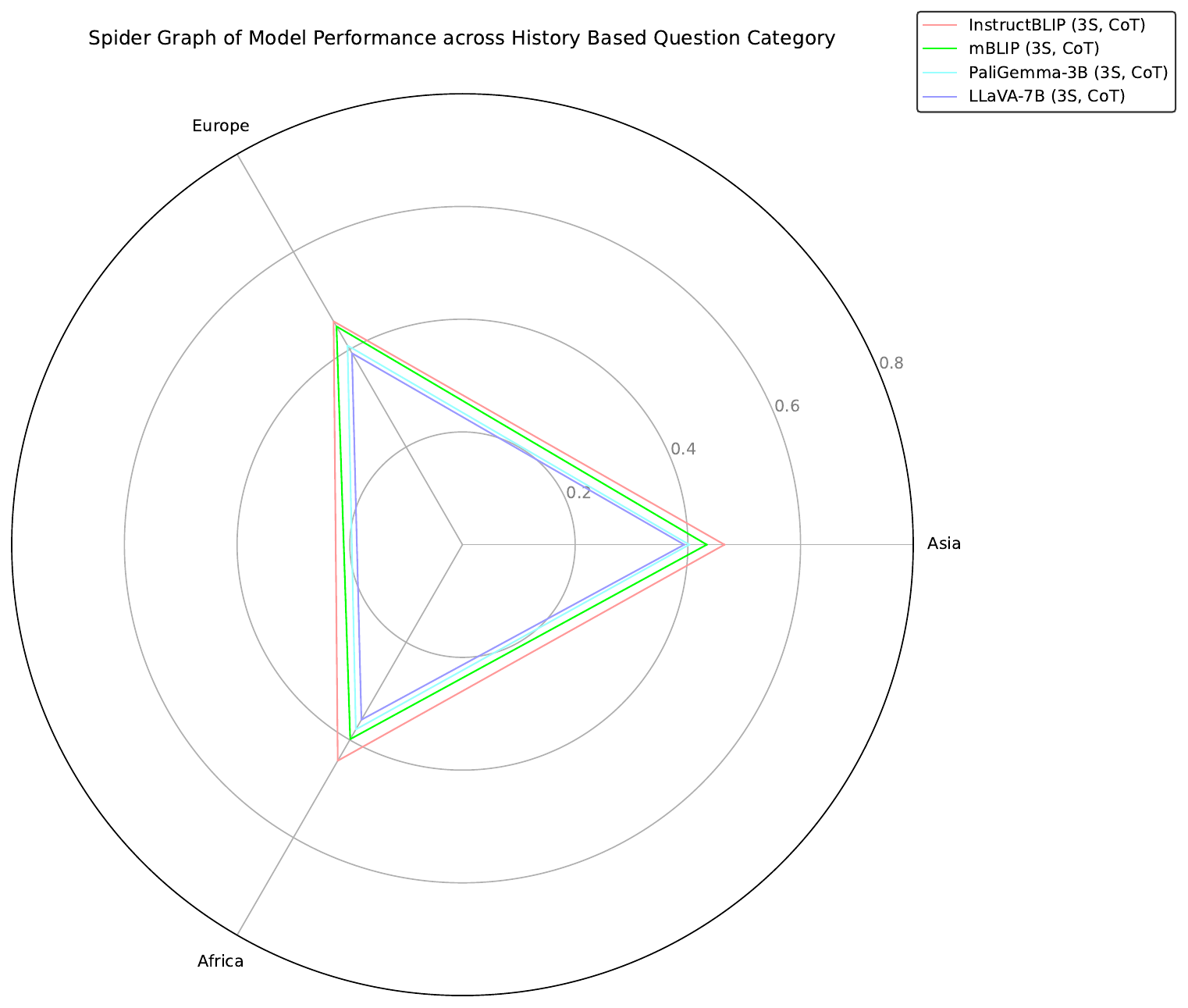}
        \caption{CoT-based results of MLLMs  across continents}
        \label{fig:Continent_History_3S_COT_VLM}
    \end{subfigure}
    \caption{}
    
    \label{fig:Continent_History_results}
\end{figure*}

\begin{figure*}[hbt!]
    \centering
    \begin{subfigure}[b]{0.32\textwidth}
        \centering
        \includegraphics[width=\textwidth]{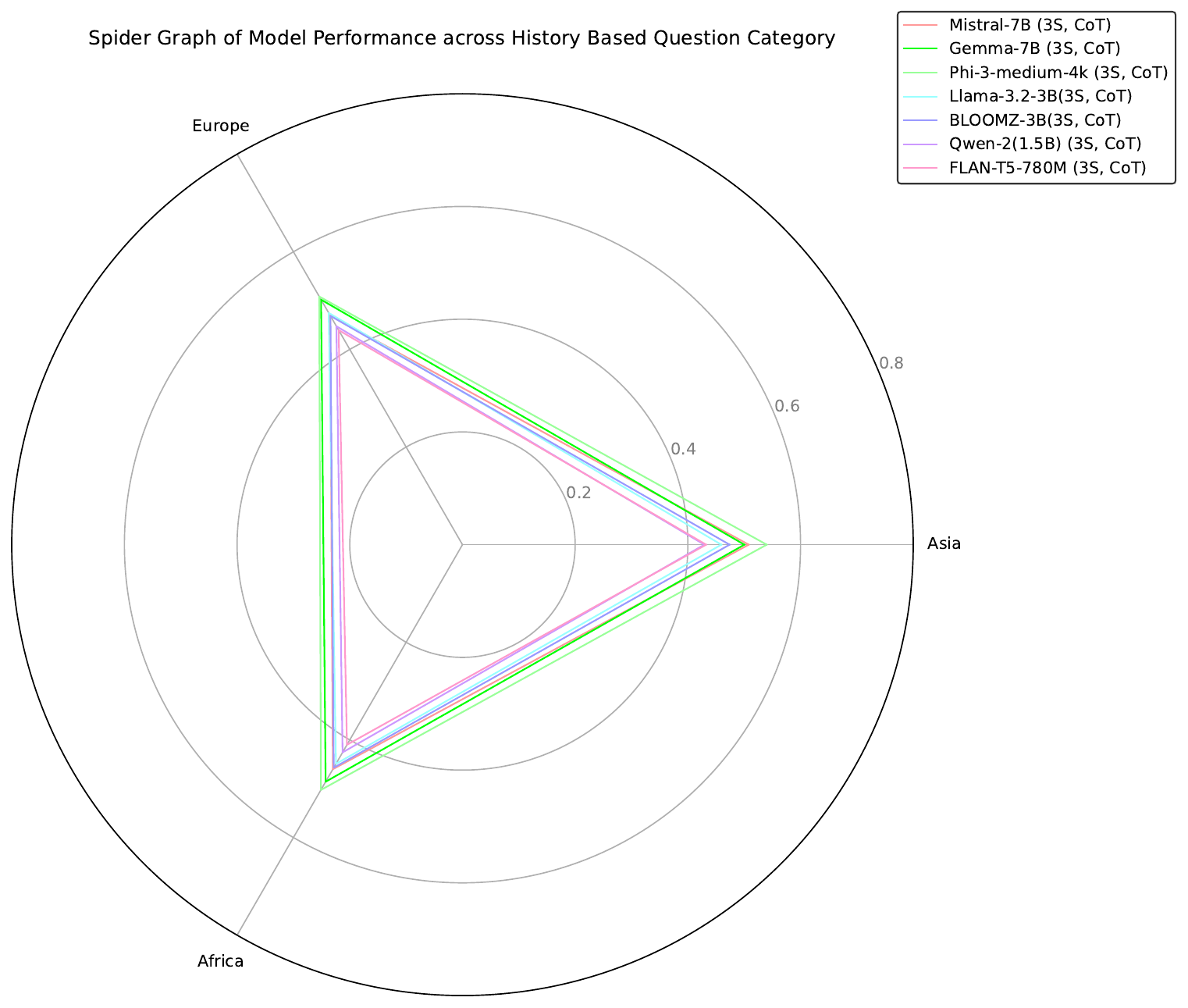}
        \caption{CoT-based results of SLMs across continents}
        \label{fig:Continent_Histroy_3S_COT_SLM}
    \end{subfigure}
    \hfill
    \begin{subfigure}[b]{0.32\textwidth}
        \centering
        \includegraphics[width=\textwidth]{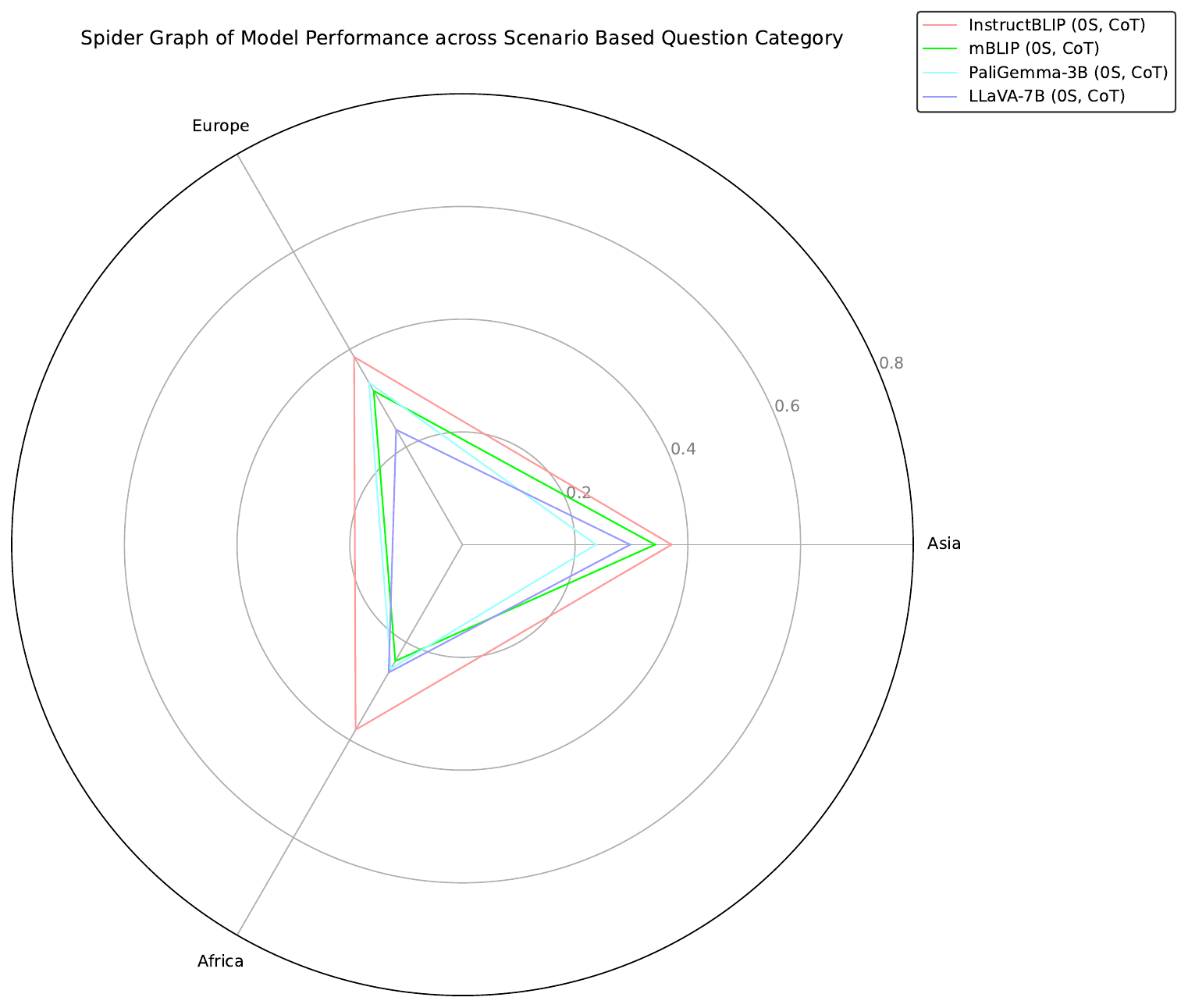}
        \caption{Few-shot-based results of MLLMs across continents}
        \label{fig:Contient_ScenarioZERO_COT_VLM}
    \end{subfigure}
    \hfill
    \begin{subfigure}[b]{0.32\textwidth}
        \centering
        \includegraphics[width=\textwidth]{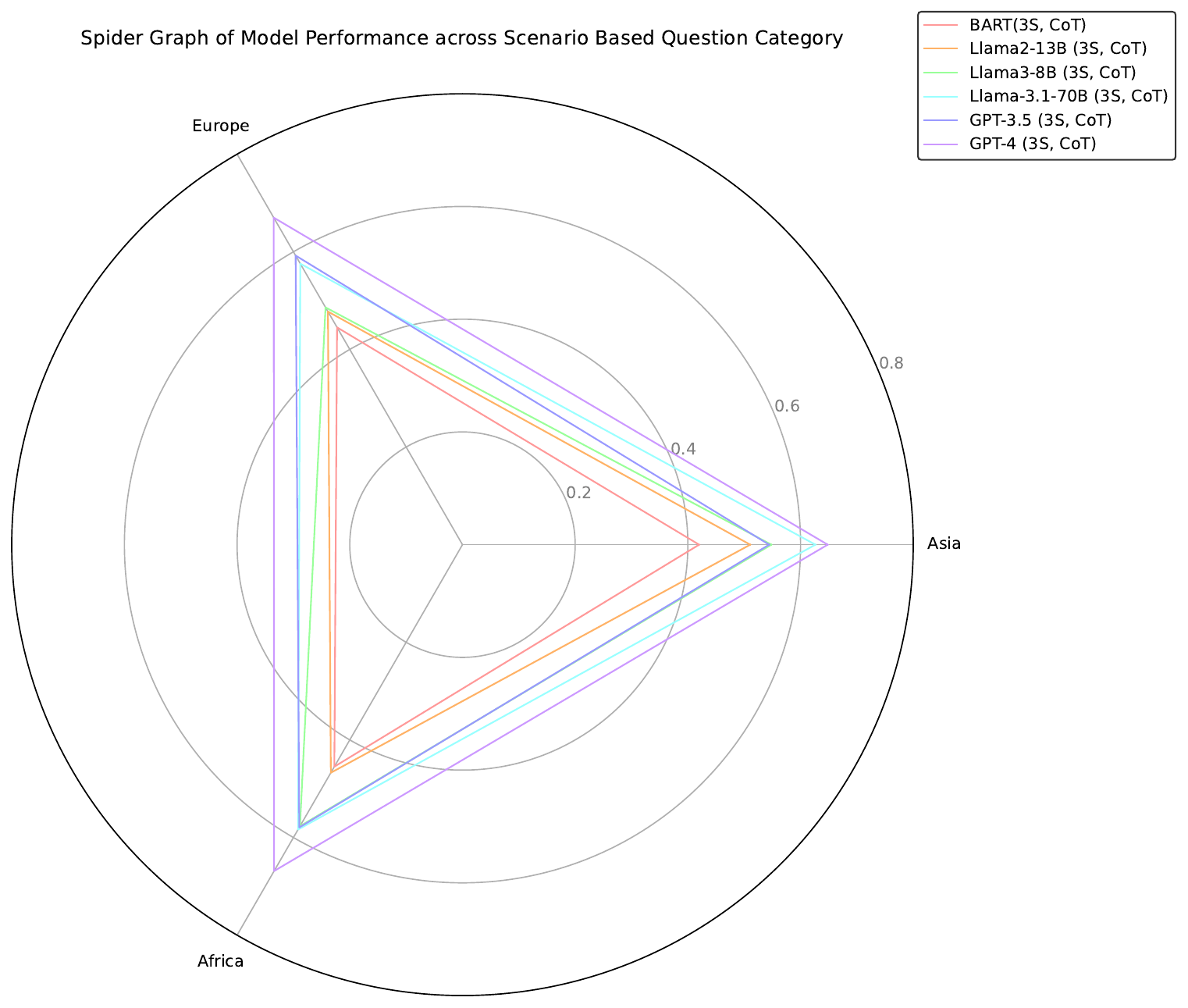}
        \caption{Few-shot-based results of SLMs across continents}
        \label{fig:Continent_Scenario_3S_COT_LLM}
    \end{subfigure}
     \caption{}
    
    \label{fig:Continent_Scenario_results}
\end{figure*}


\begin{figure*}[hbt!]
    \centering
    \begin{subfigure}[b]{0.32\textwidth}
        \centering
        \includegraphics[width=\textwidth]{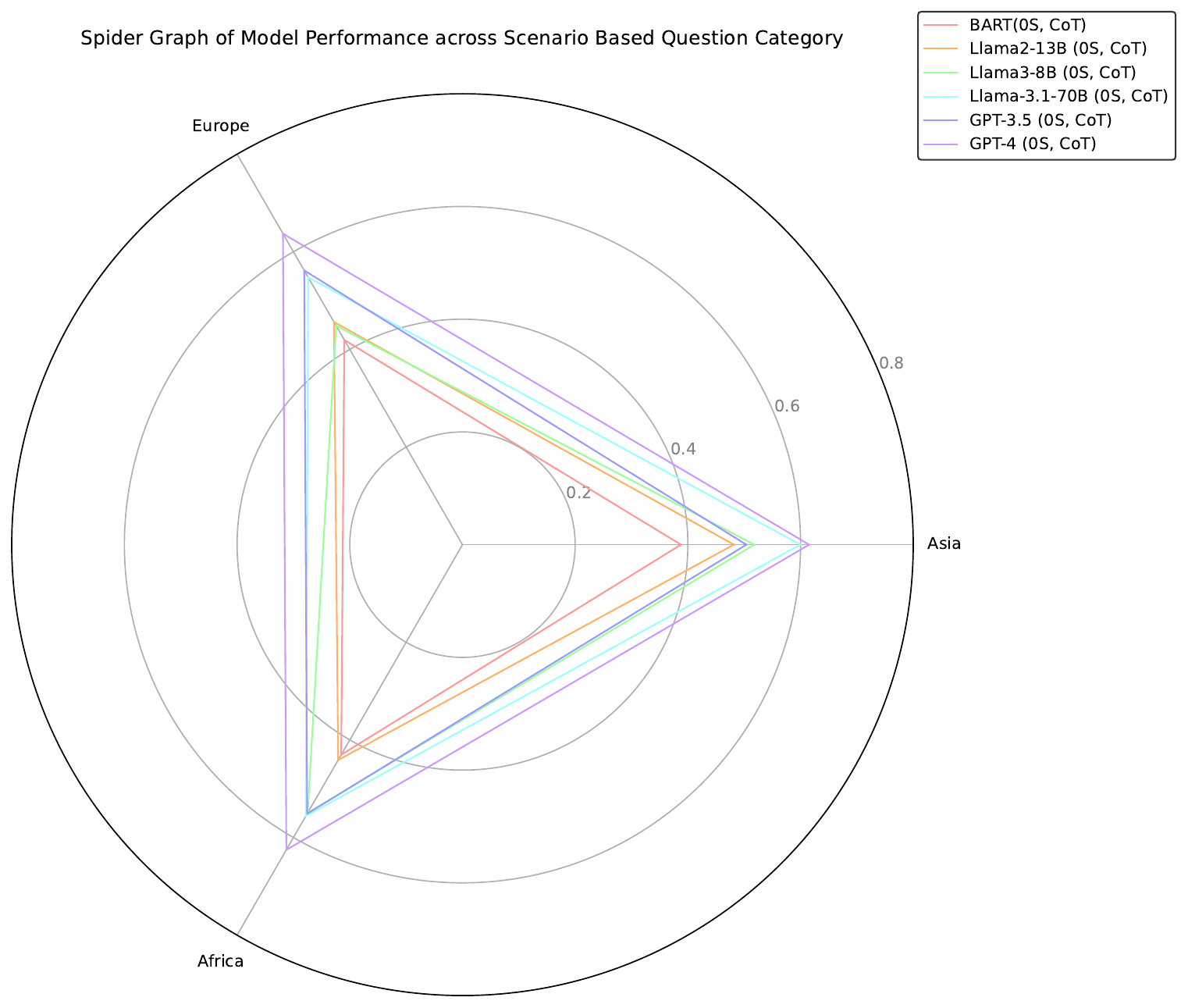}
        \caption{Few-shot-based results of LLMs across continents}
        \label{fig:Continent_Scenario_Zero_COT_LLM}
    \end{subfigure}
    \hfill
    \begin{subfigure}[b]{0.32\textwidth}
        \centering
        \includegraphics[width=\textwidth]{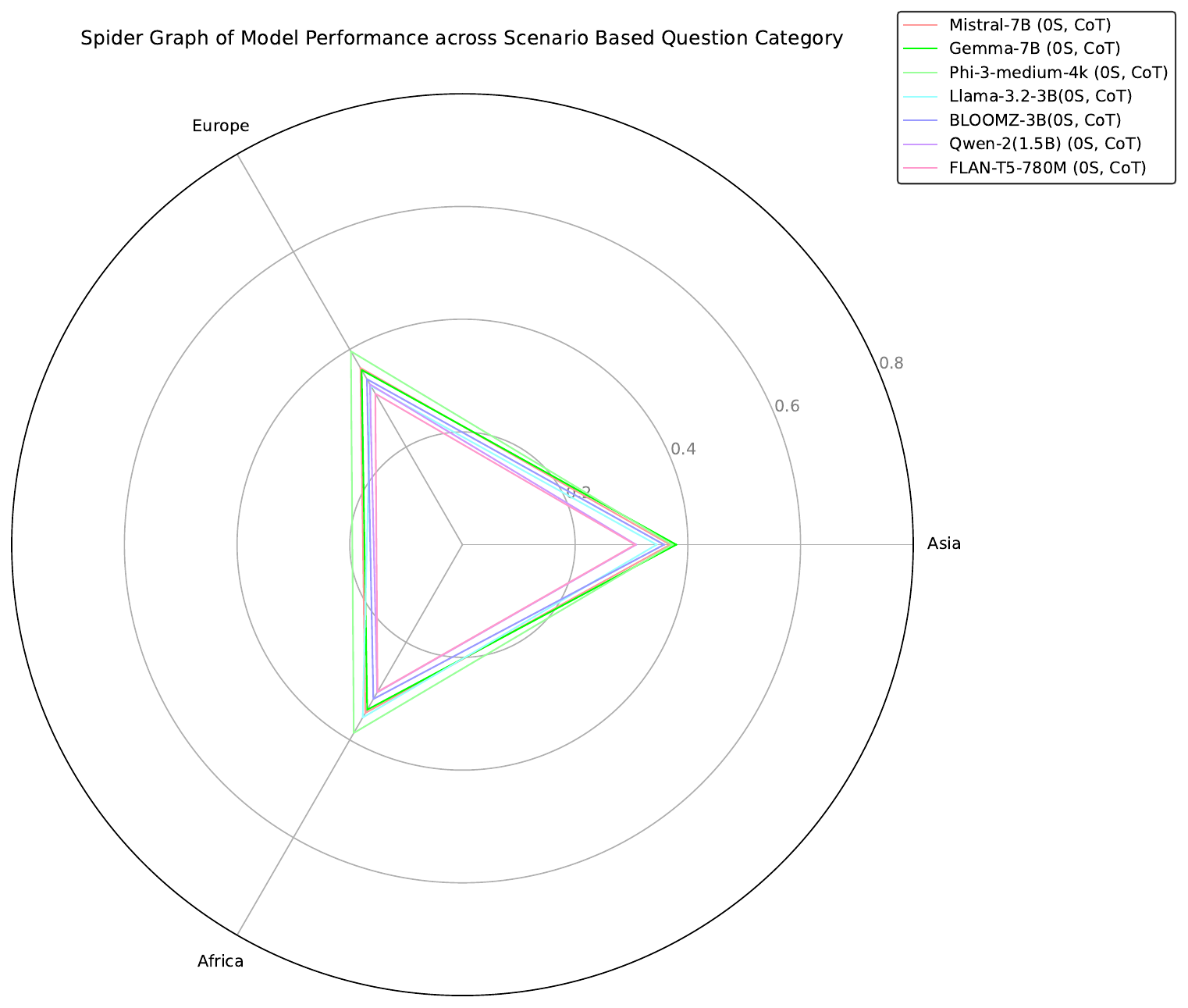}
        \caption{Few-shot-based results of SLMs across continents}
        \label{fig:Continent_Scenario_Zero_COT_SLM}
    \end{subfigure}
    
    \caption{}
    \label{fig:Continent_Scenario_Zero_results}
\end{figure*}


\begin{figure*}
\includegraphics[width=\textwidth]{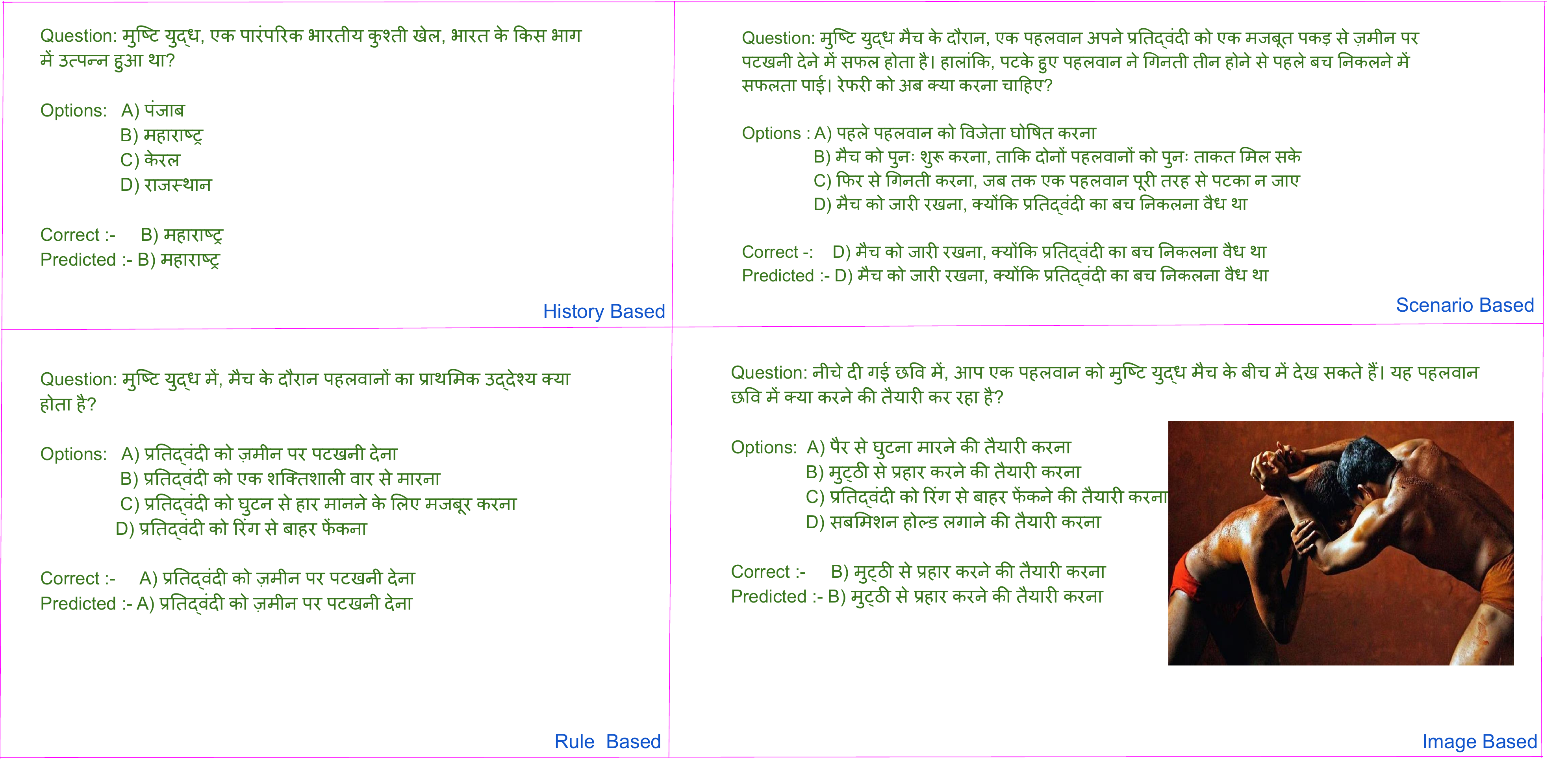}
\caption{Example Illustration of India Traditional Sports Correct Prediction.} 
\label{IndTSCP}
\end{figure*}

\begin{figure*}
\includegraphics[width=\textwidth]{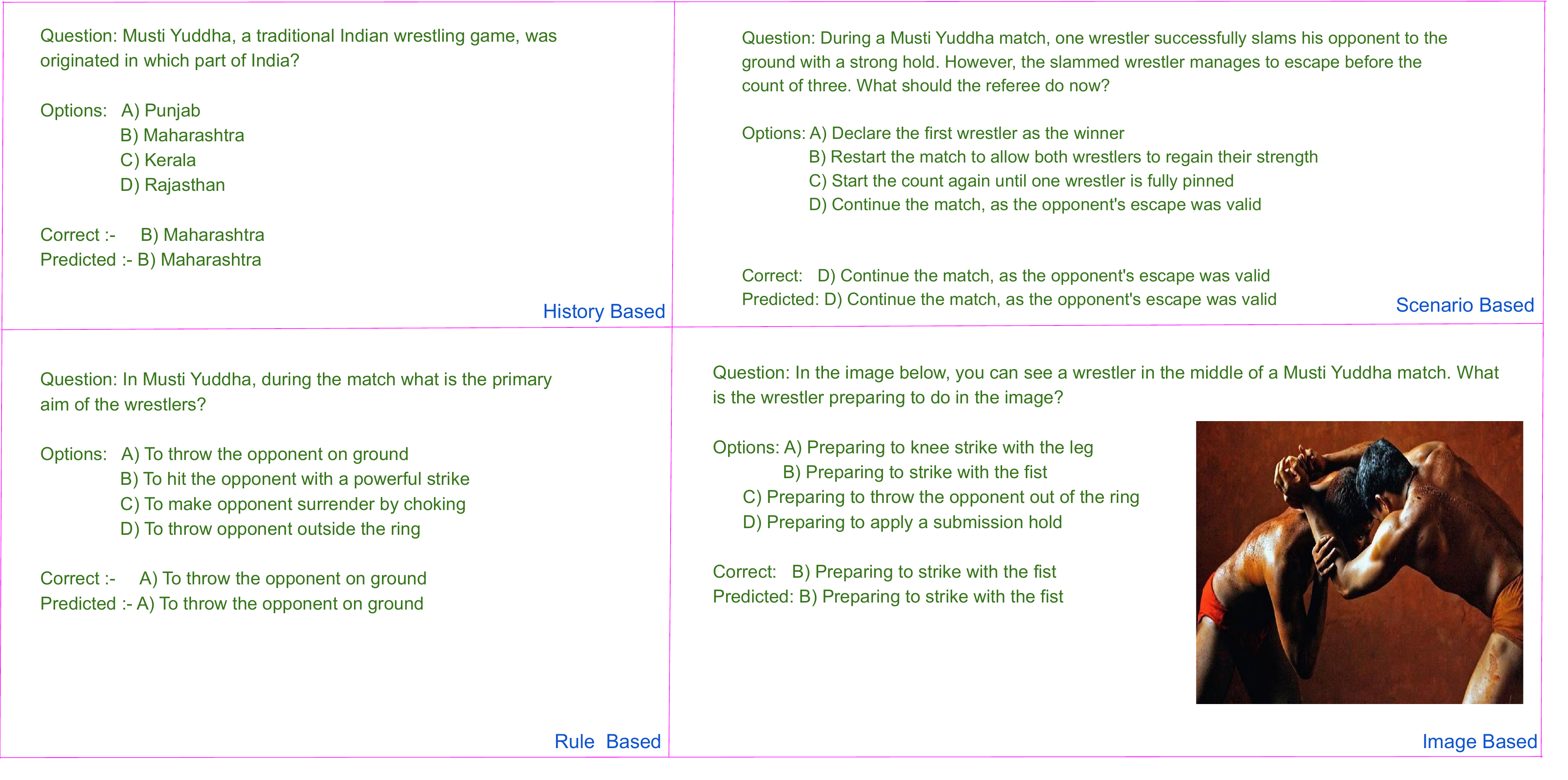}
\caption{Example Illustration of India Traditional Sports Correct Prediction (In English)} 
\label{IndTSCPE}
\end{figure*}

\begin{figure*}
\includegraphics[width=\textwidth]{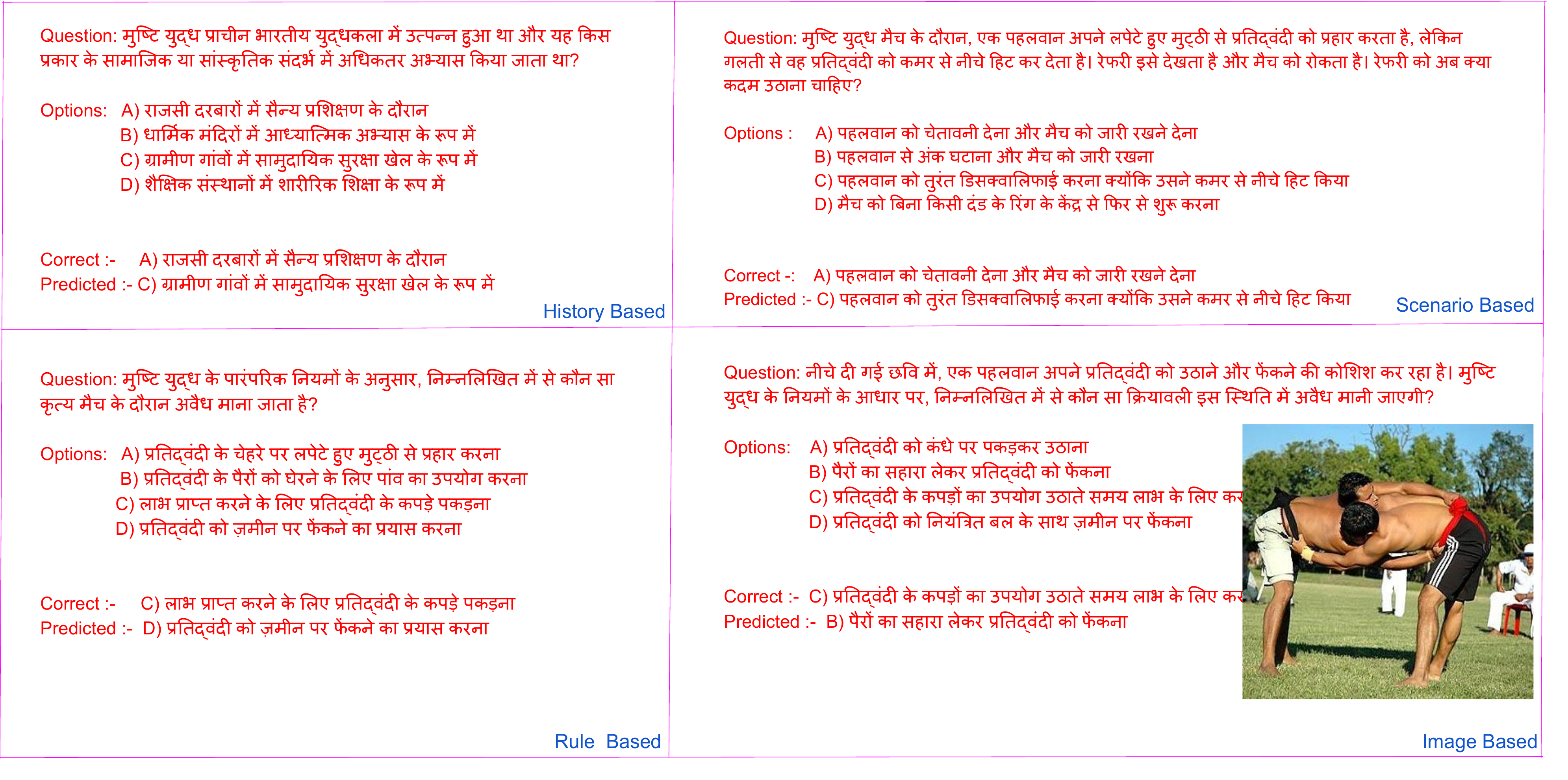}
\caption{Example Illustration of India Traditional Sports Wrong Prediction.} 
\label{IndTSWP}
\end{figure*}
\begin{figure*}
\includegraphics[width=\textwidth]{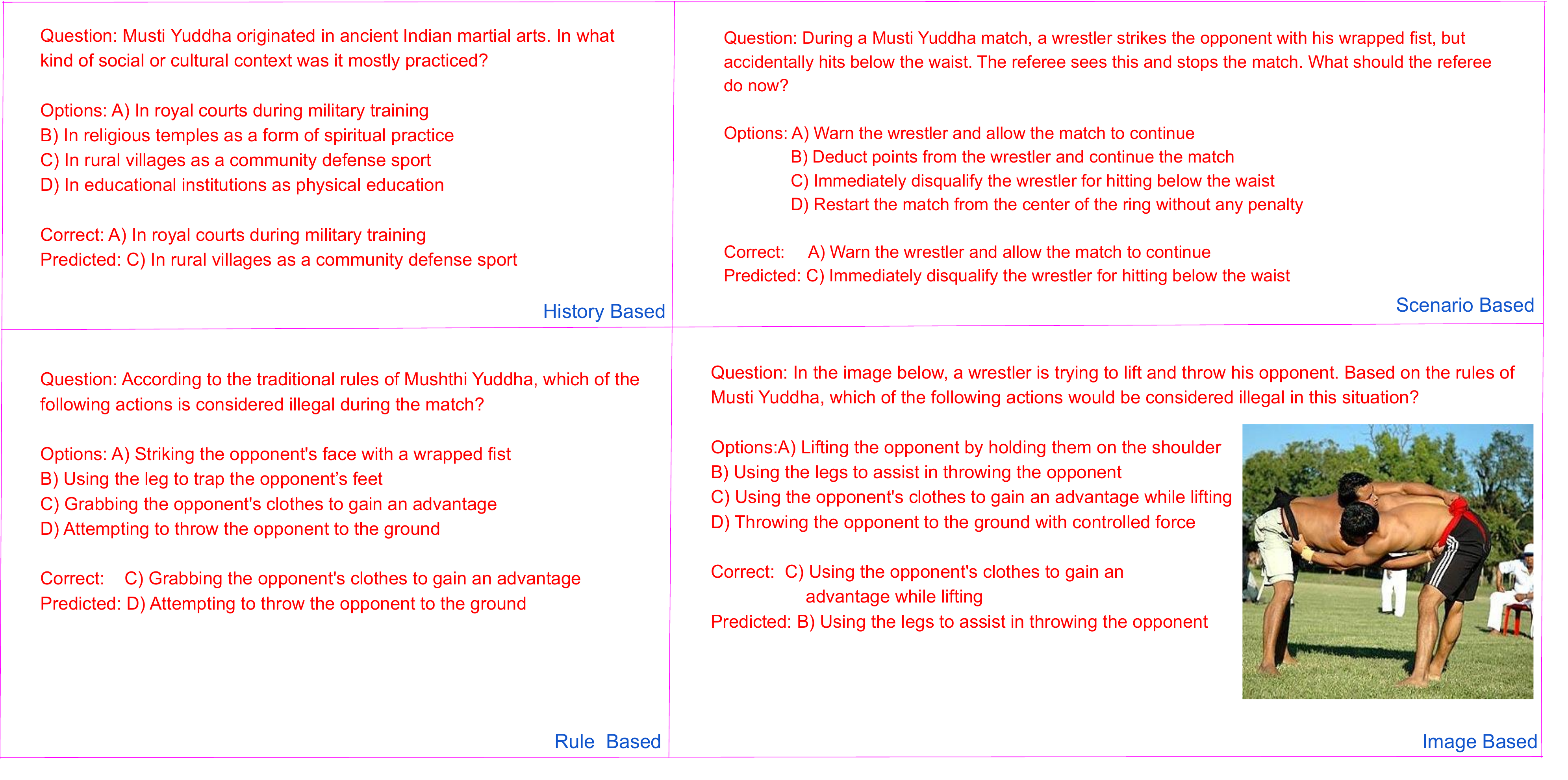}
\caption{Example Illustration of India Traditional Sports Wrong Prediction (In English)} 
\label{IndTSWPE}
\end{figure*}

\begin{figure*}
\includegraphics[width=\textwidth]{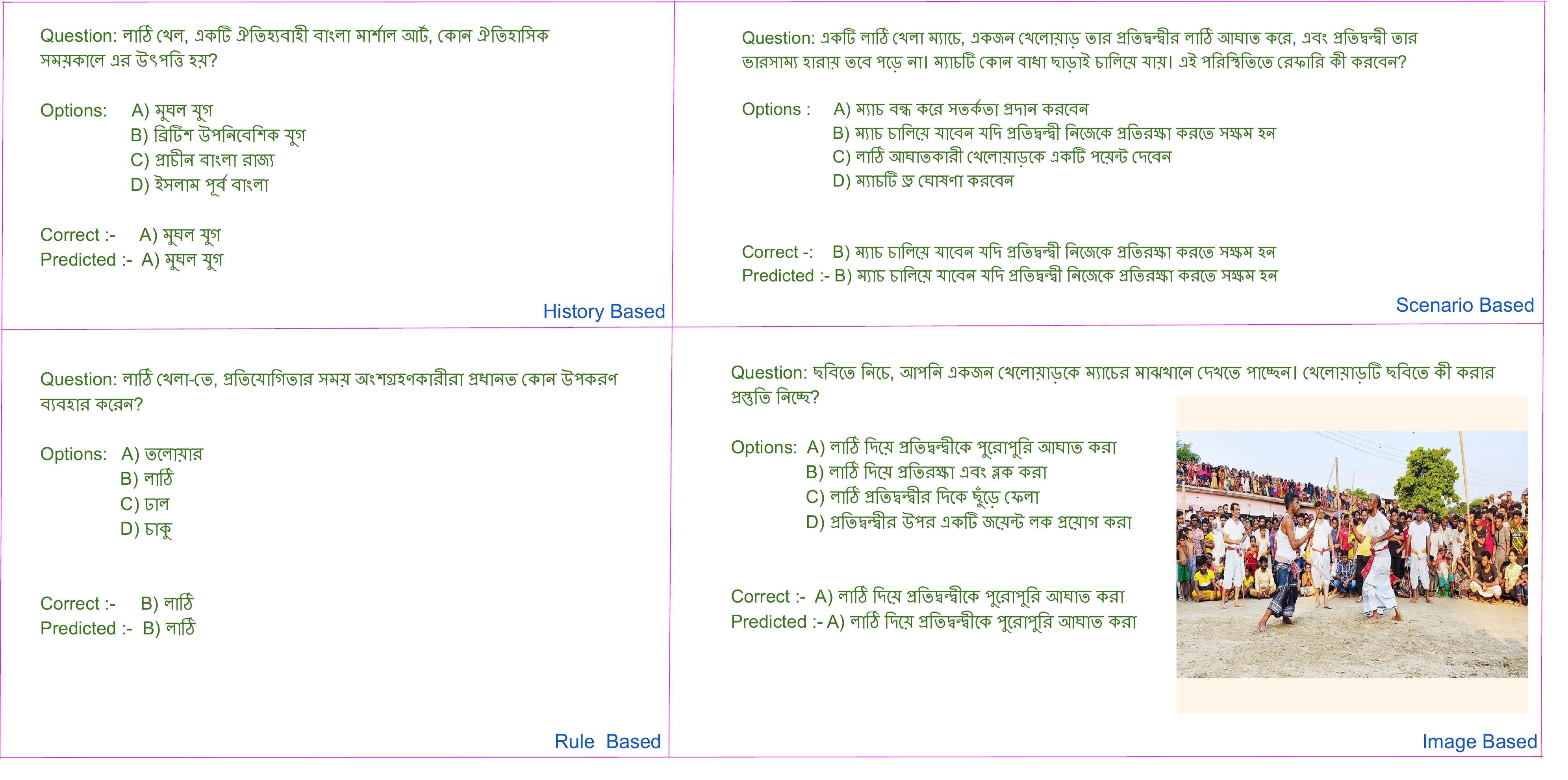}
\caption{Example Illustration of Bangladesh Traditional Sports Correct Prediction} 
\label{BTSCP}
\end{figure*}

\begin{figure*}
\includegraphics[width=\textwidth]{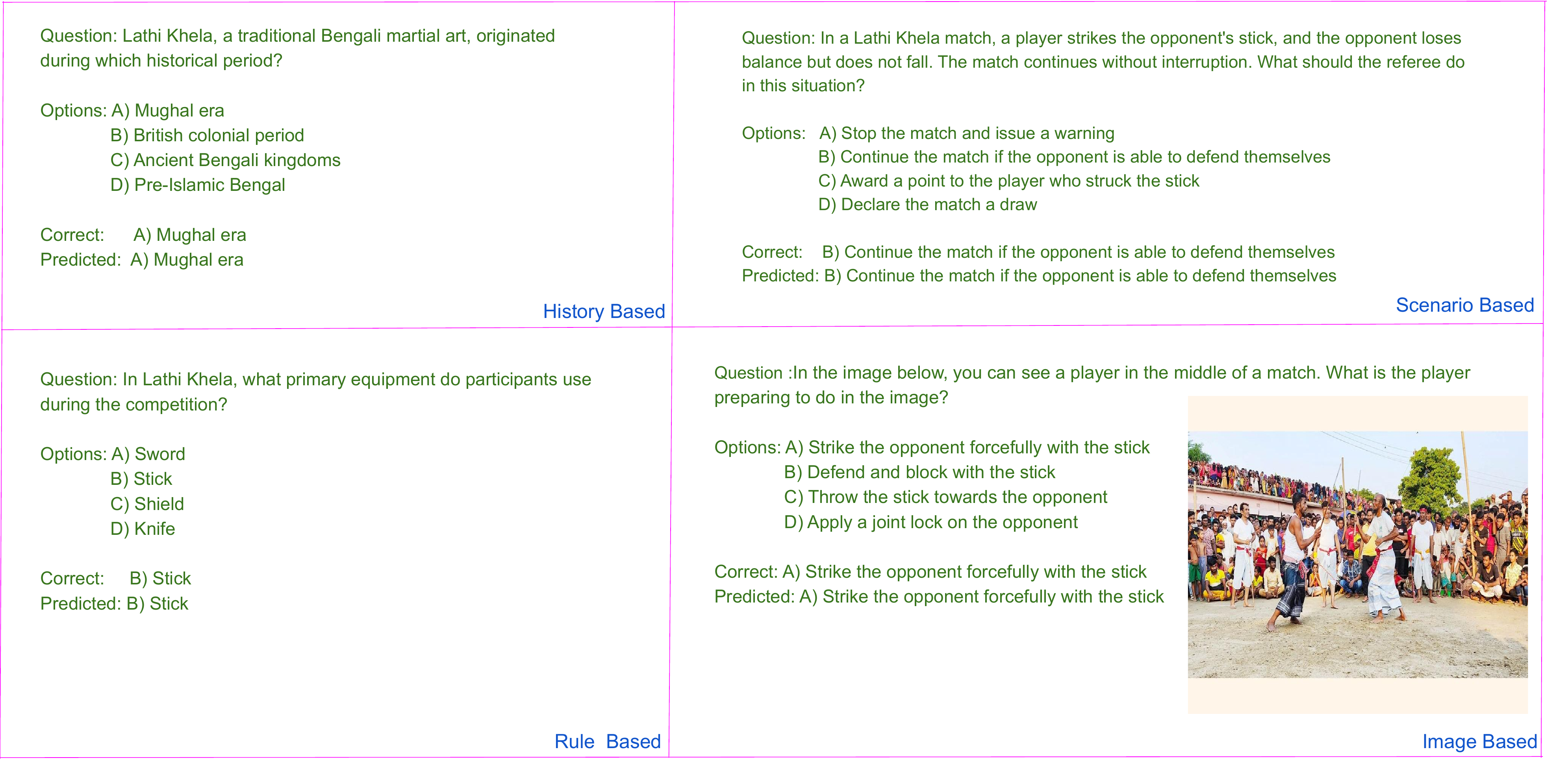}
\caption{Example Illustration of Bangladesh Traditional Sports Correct Prediction (In English)} 
\label{BTSCPE}
\end{figure*}

\begin{figure*}
\includegraphics[width=\textwidth]{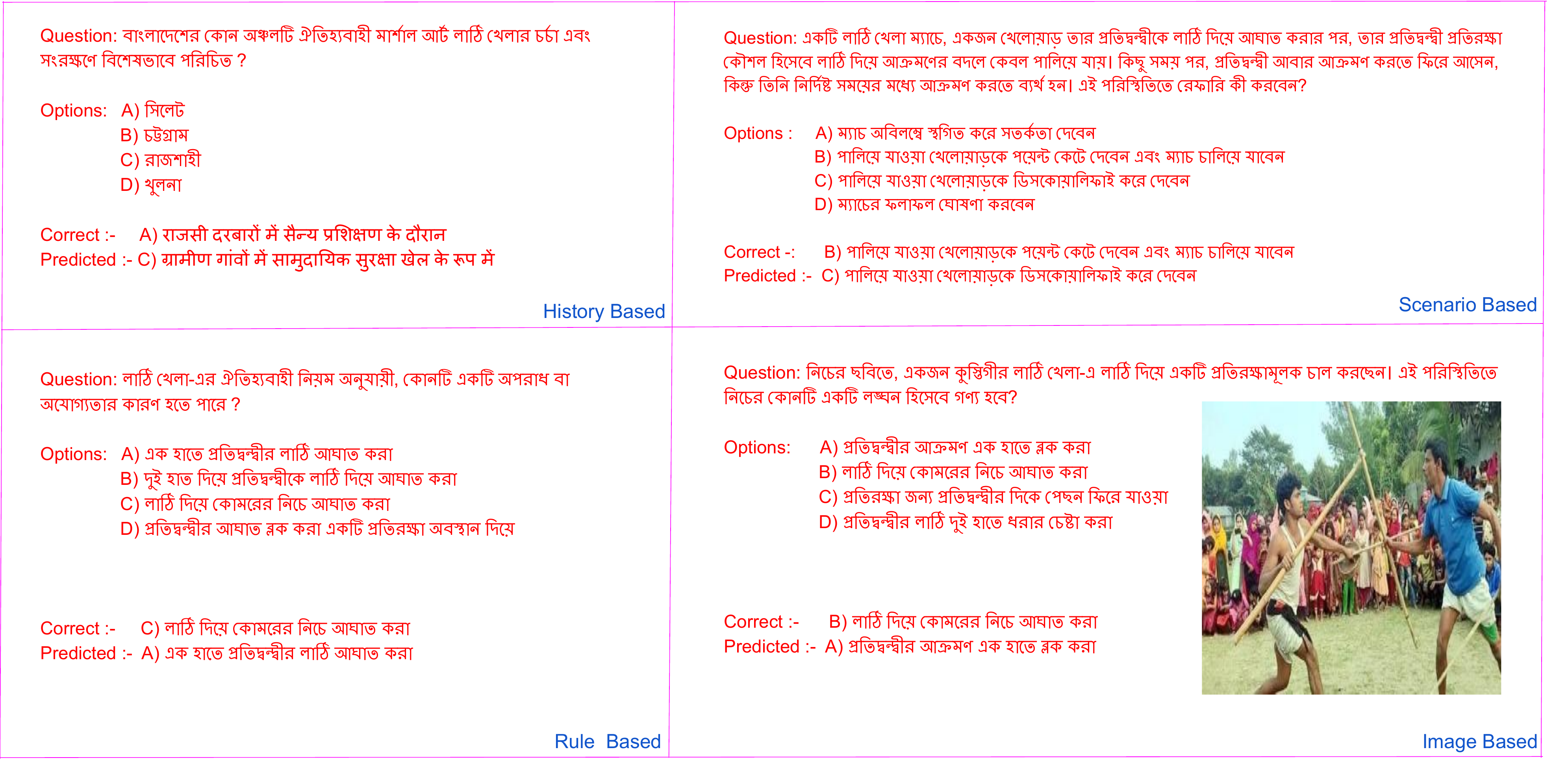}
\caption{Example Illustration of Bangladesh Traditional Sports Wrong Prediction.} 
\label{BTSWP}
\end{figure*}

\begin{figure*}
\includegraphics[width=\textwidth]{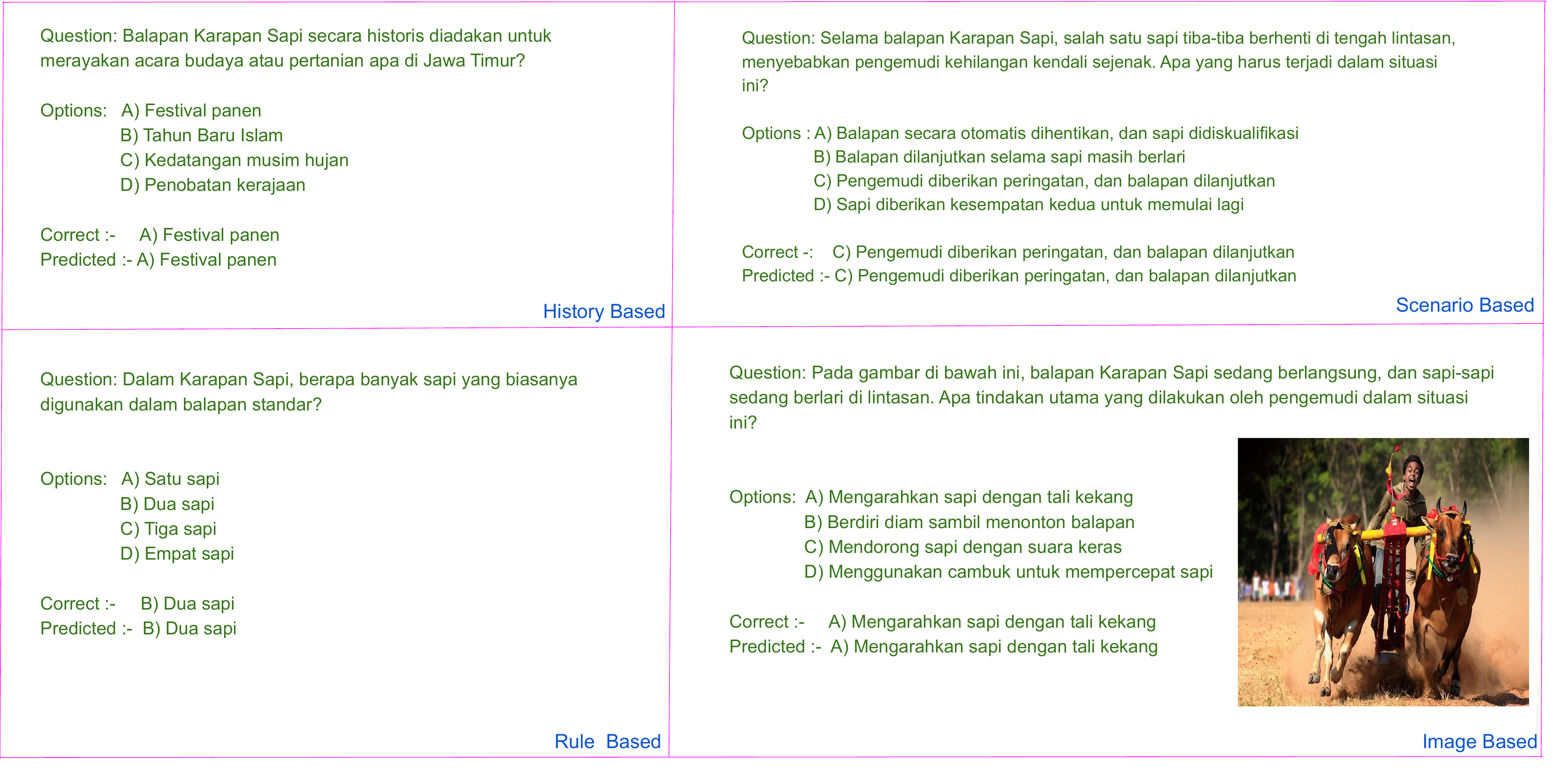}
\caption{Example Illustration of Indonesia Traditional Sports Correct Prediction.} 
\label{IndoTSCP}
\end{figure*}

\begin{figure*}
\includegraphics[width=\textwidth]{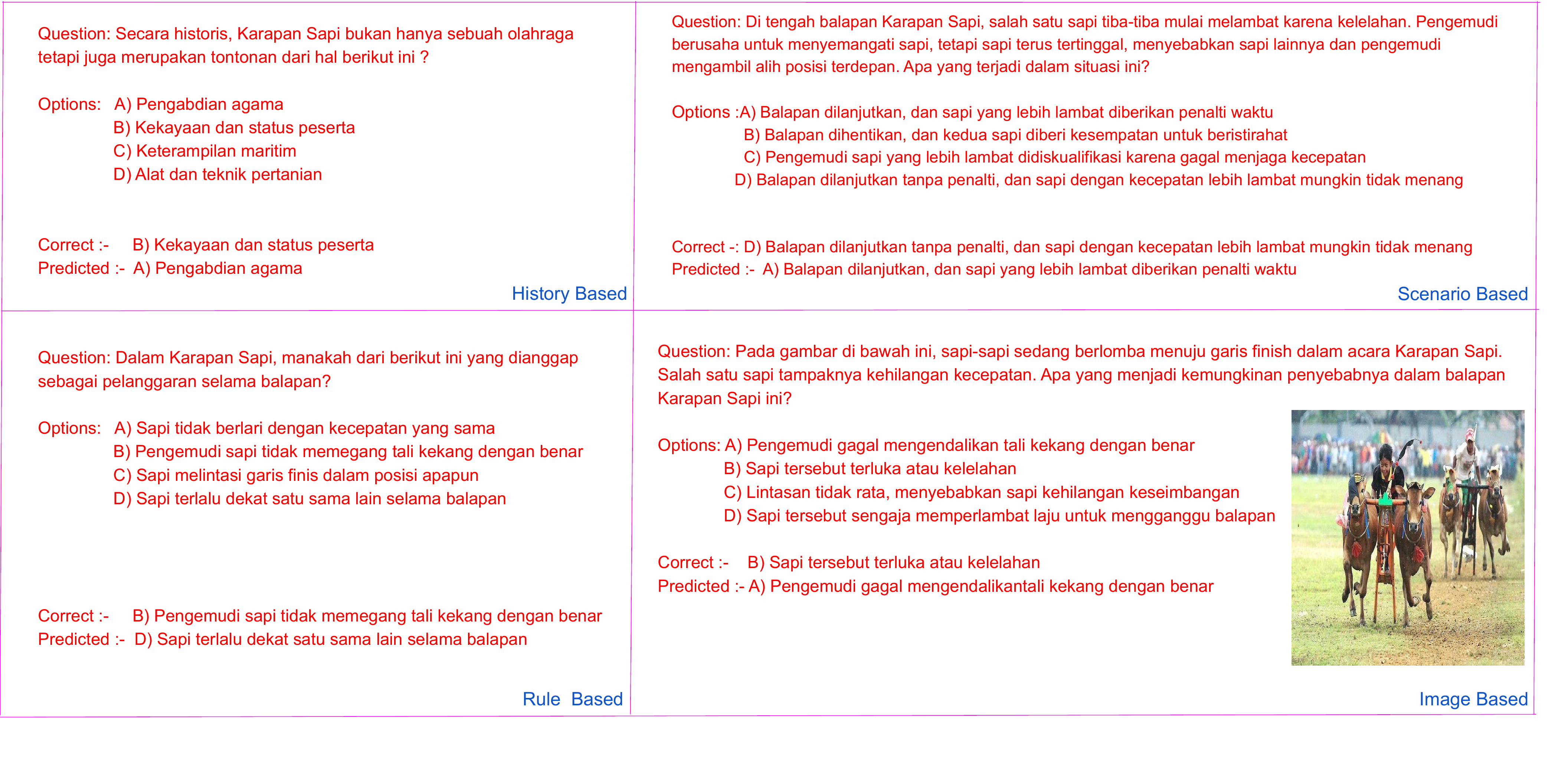}
\caption{Example Illustration of Indonesia Traditional Sports Wrong Prediction.} 
\label{IndoTSWP}
\end{figure*}

\begin{figure*}
\includegraphics[width=\textwidth]{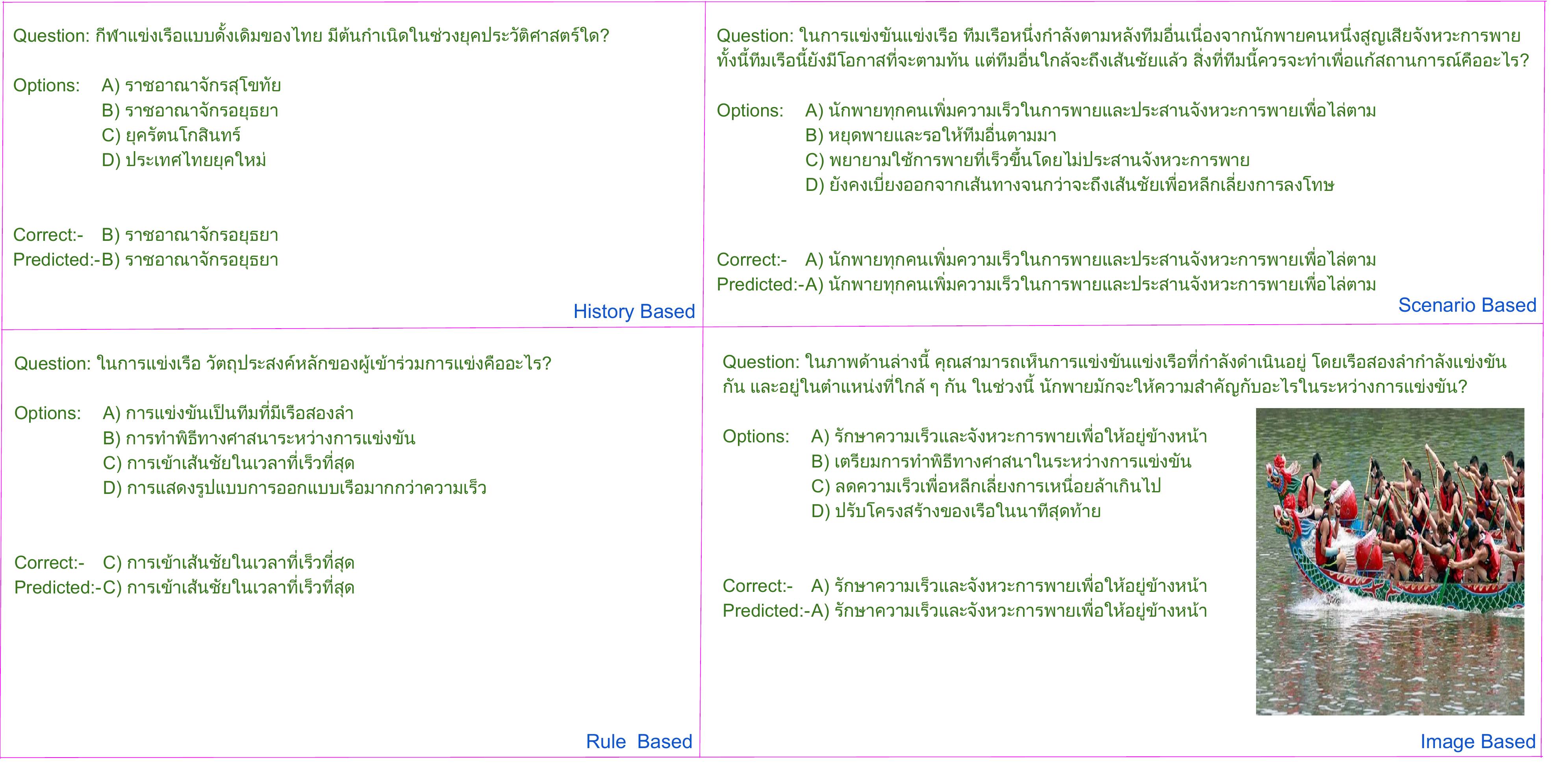}
\caption{Example Illustration of Thailand Traditional Sports Correct Prediction.} 
\label{TTSCP}
\end{figure*}

\begin{figure*}
\includegraphics[width=\textwidth]{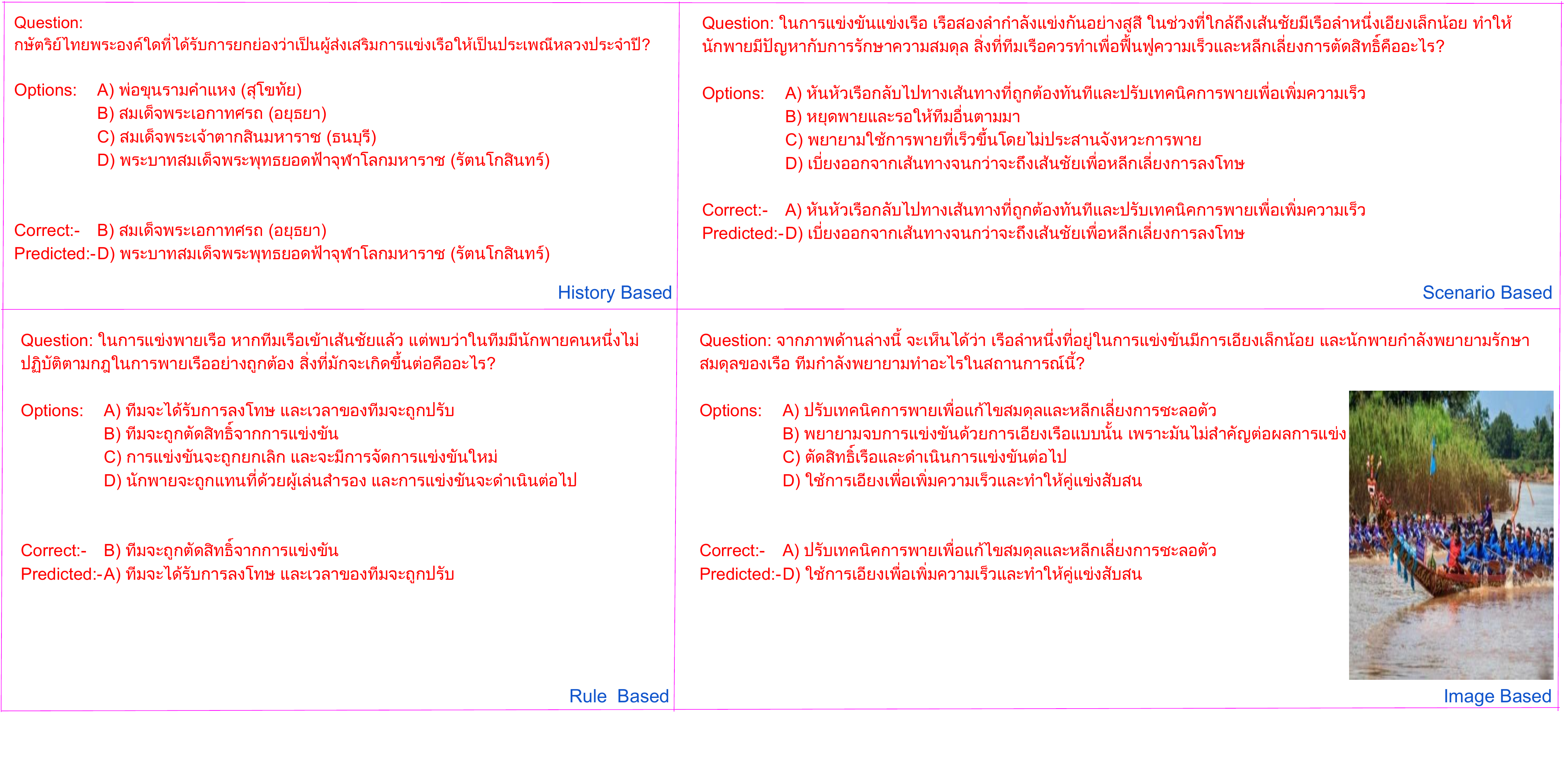}
\caption{Example Illustration of Thailand Traditional Sports Wrong Prediction.} 
\label{TTSWP}
\end{figure*}

\begin{figure*}
\includegraphics[width=\textwidth]{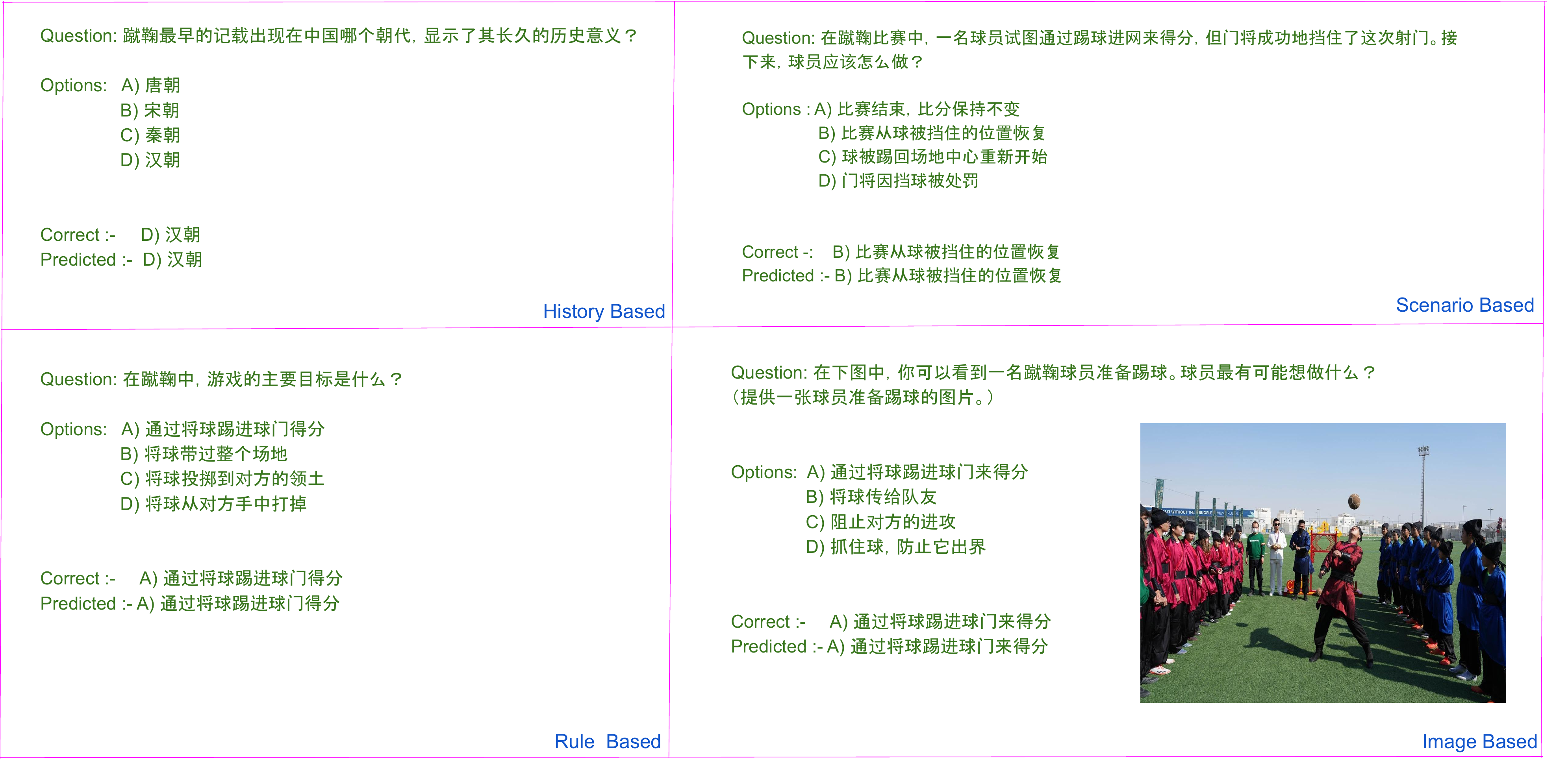}
\caption{Example Illustration of China Traditional Sports Correct Prediction.} 
\label{CTSCP}
\end{figure*}

\begin{figure*}
\includegraphics[width=\textwidth]{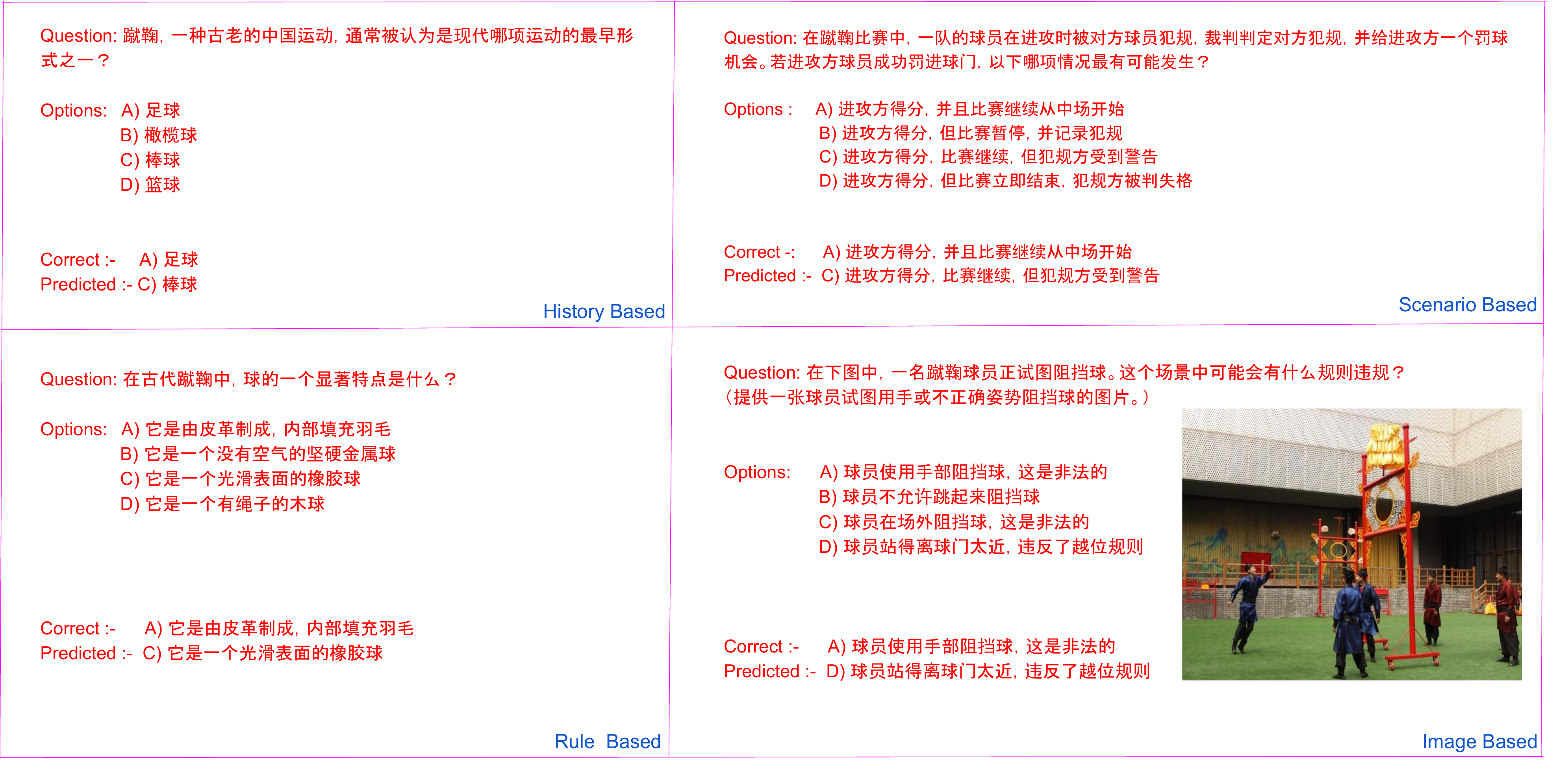}
\caption{Example Illustration of China Traditional Sports Wrong Prediction.} 
\label{CTSWP}
\end{figure*}

\begin{figure*}
\includegraphics[width=\textwidth]{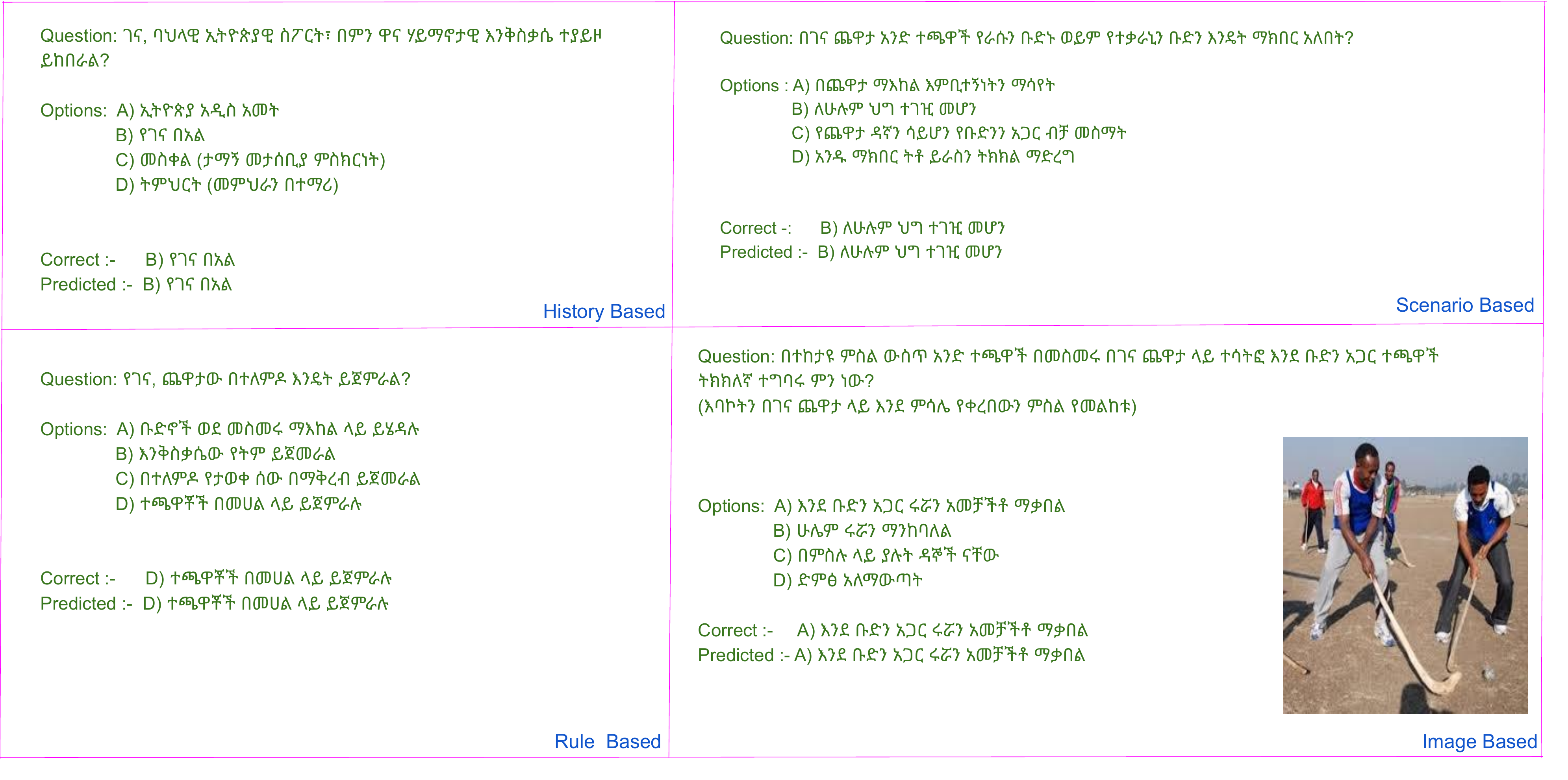}
\caption{Example Illustration of Ethiopia Traditional Sports Correct Prediction.} 
\label{ETSCP}
\end{figure*}

\begin{figure*}
\includegraphics[width=\textwidth]{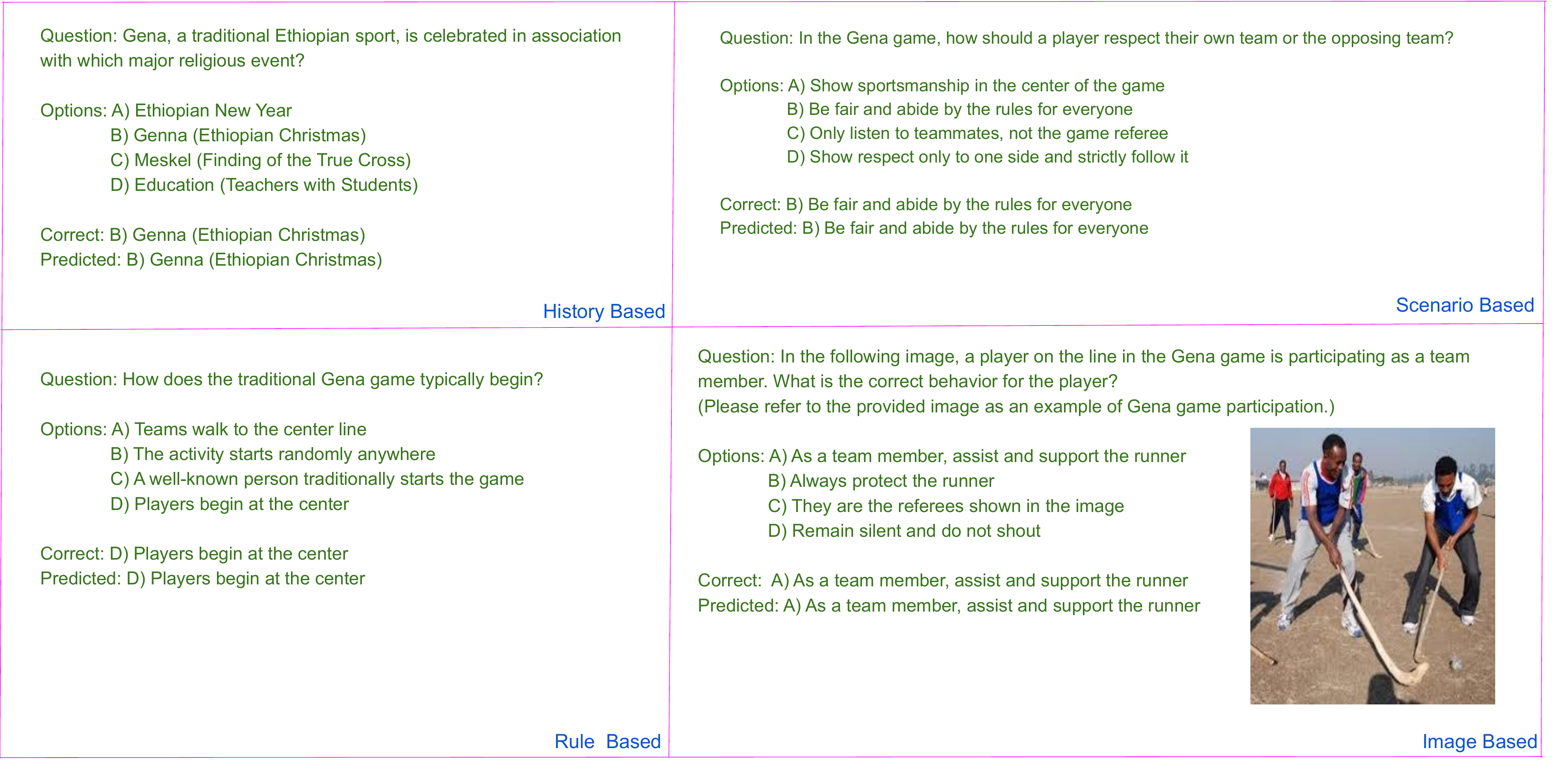}
\caption{Example Illustration of Ethiopia Traditional Sports Correct Prediction (In English)} 
\label{ETSCPE}
\end{figure*}

\begin{figure*}
\includegraphics[width=\textwidth]{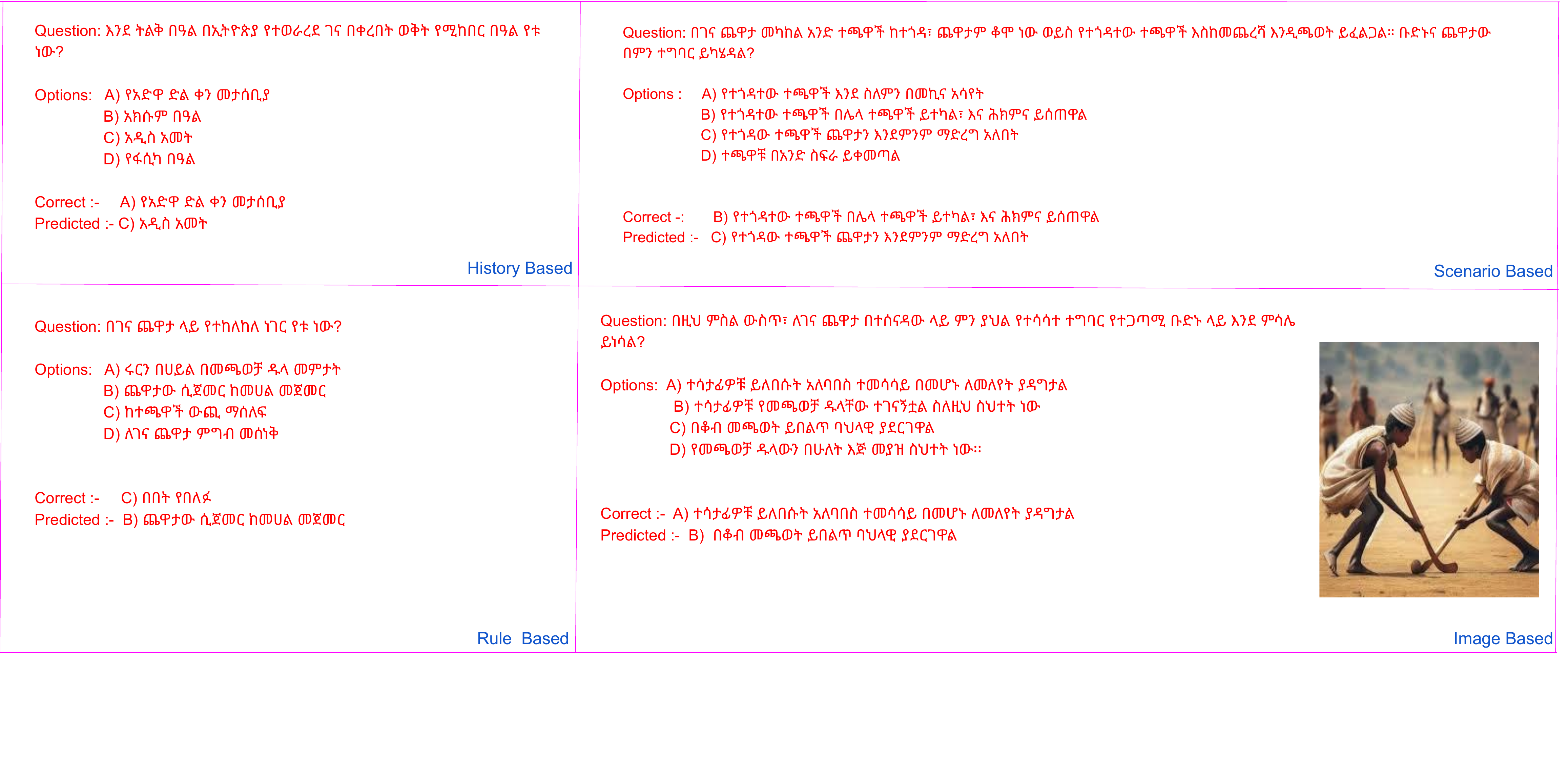}
\caption{Example Illustration of Ethiopia Traditional Sports Wrong Prediction.} 
\label{ETSWP}
\end{figure*}

\begin{figure*}
\includegraphics[width=\textwidth]{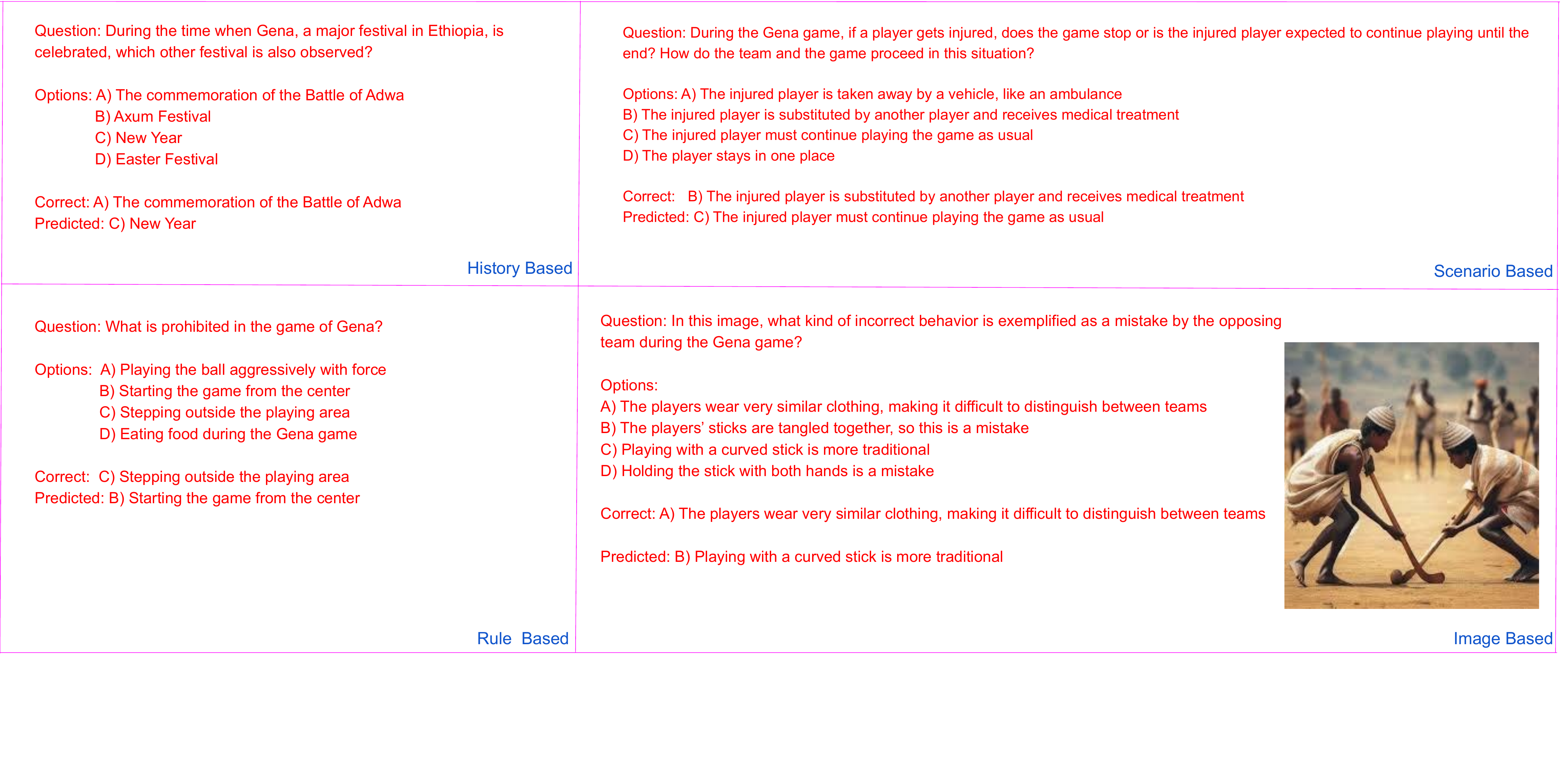}
\caption{Example Illustration of Ethiopia Traditional Sports Wrong Prediction (In English)} 
\label{ETSWPE}
\end{figure*}

\begin{figure*}
\includegraphics[width=\textwidth]{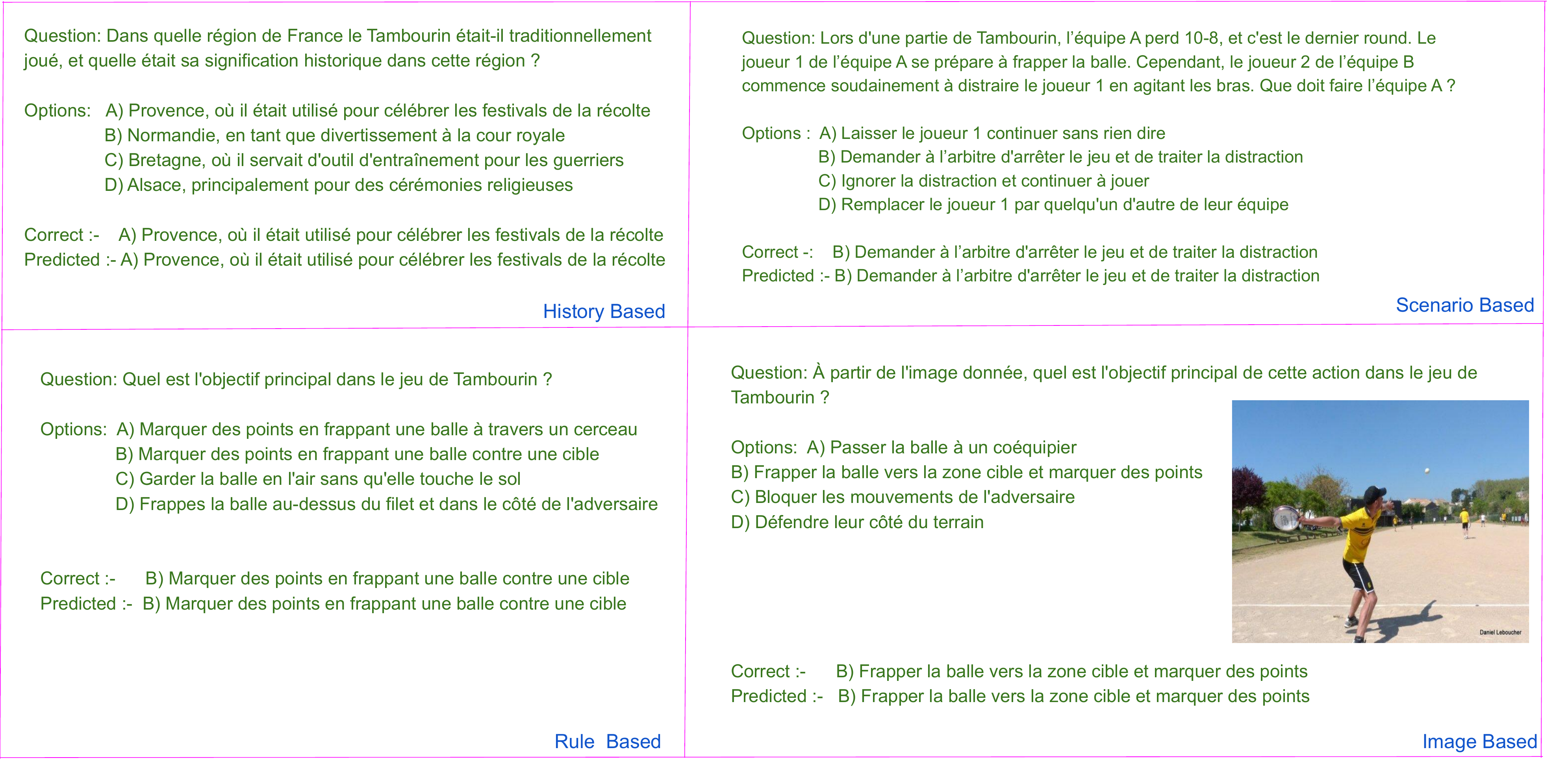}
\caption{Example Illustration of France Traditional Sports Correct Prediction} 
\label{FTSCP}
\end{figure*}

\begin{figure*}
\includegraphics[width=\textwidth]{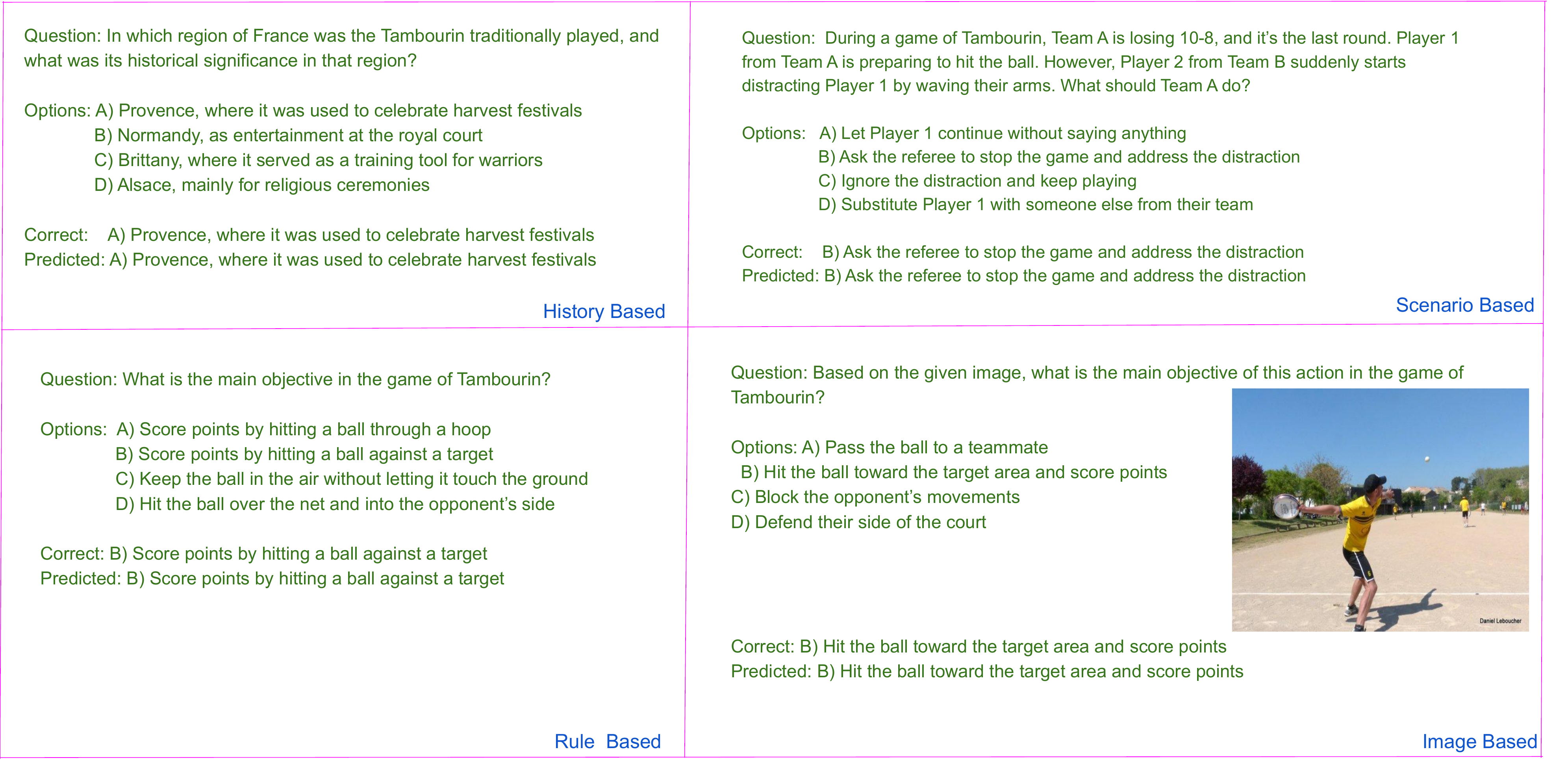}
\caption{Example Illustration of France Traditional Sports Correct Prediction (In English)} 
\label{FTSCPE}
\end{figure*}

\begin{figure*}
\includegraphics[width=\textwidth]{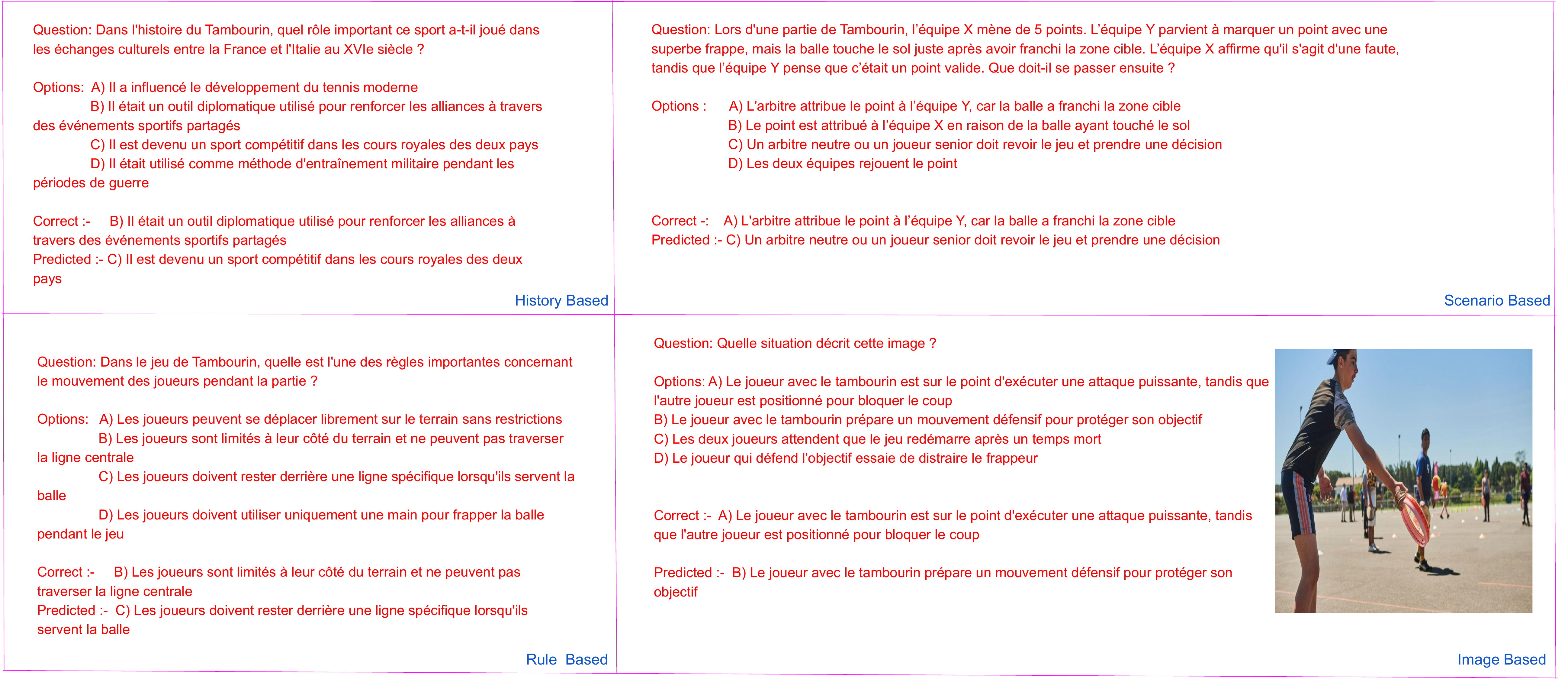}
\caption{Example Illustration of France Traditional Sports Wrong Prediction} 
\label{FTSWP}
\end{figure*}

\begin{figure*}
\includegraphics[width=\textwidth]{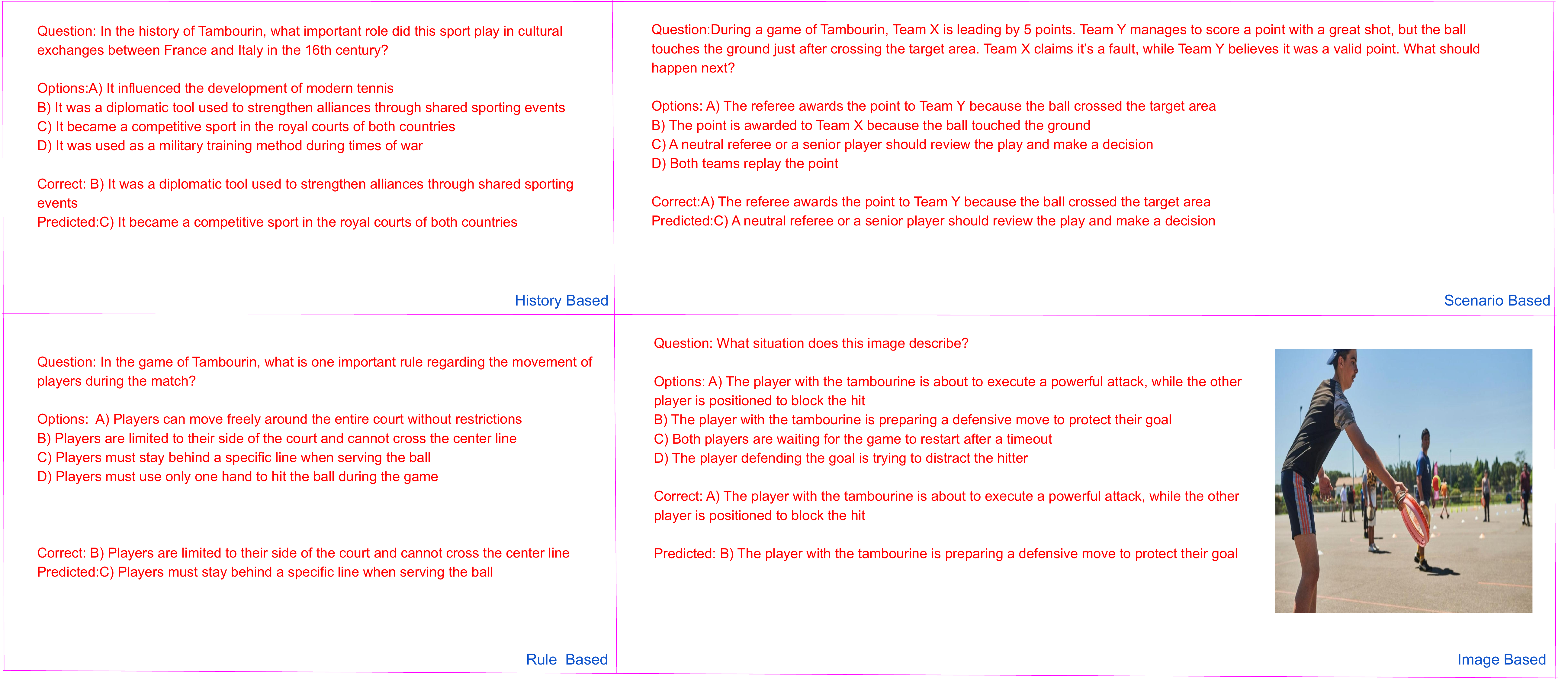}
\caption{Example Illustration of France Traditional Sports Wrong Prediction (In English)} 
\label{FTSWPE}
\end{figure*}

\begin{figure*}
\includegraphics[width=\textwidth]{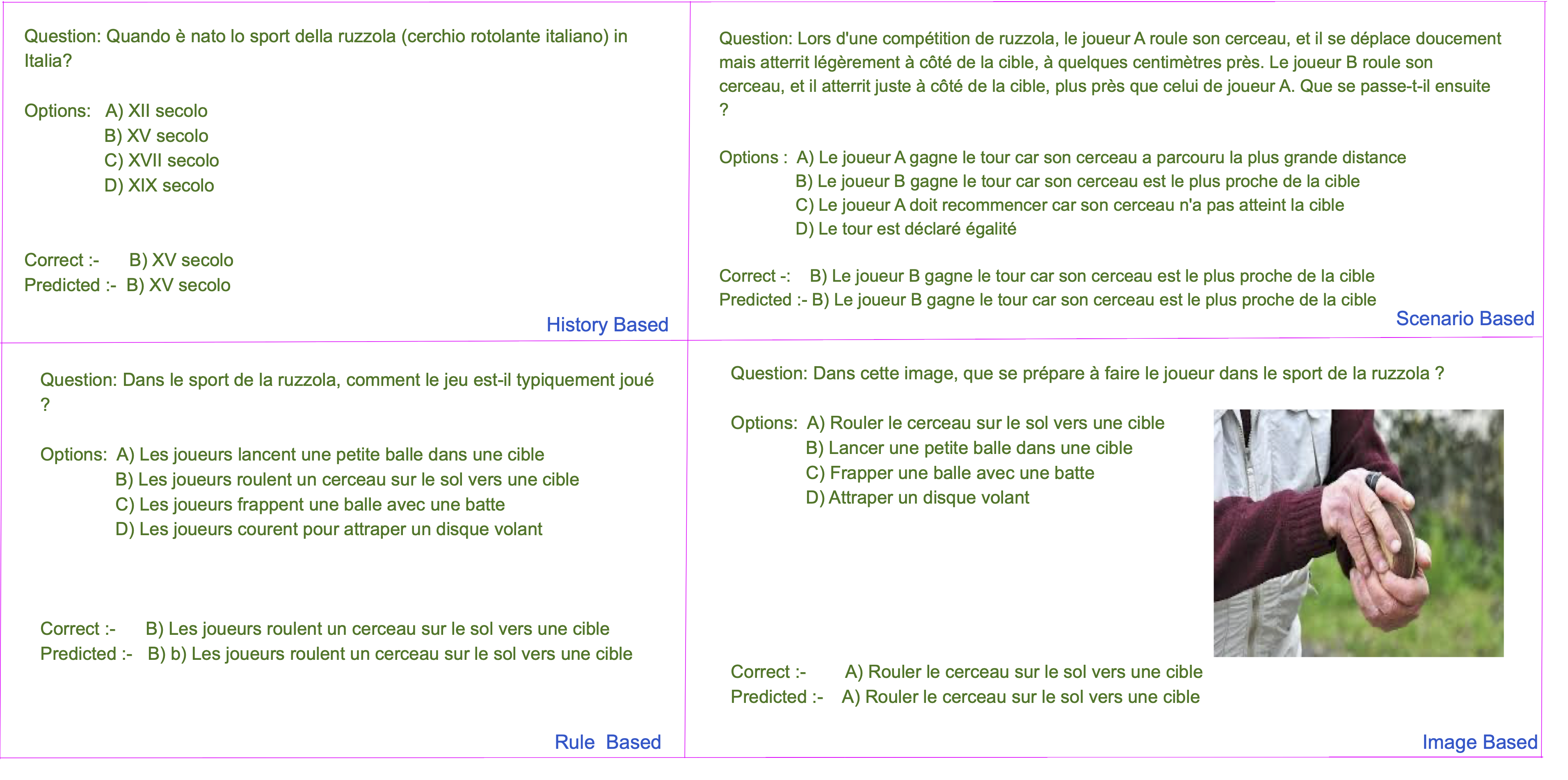}
\caption{Example Illustration of Italy Traditional Sports Correct Prediction} 
\label{ItaTSCP}
\end{figure*}

\begin{figure*}
\includegraphics[width=\textwidth]{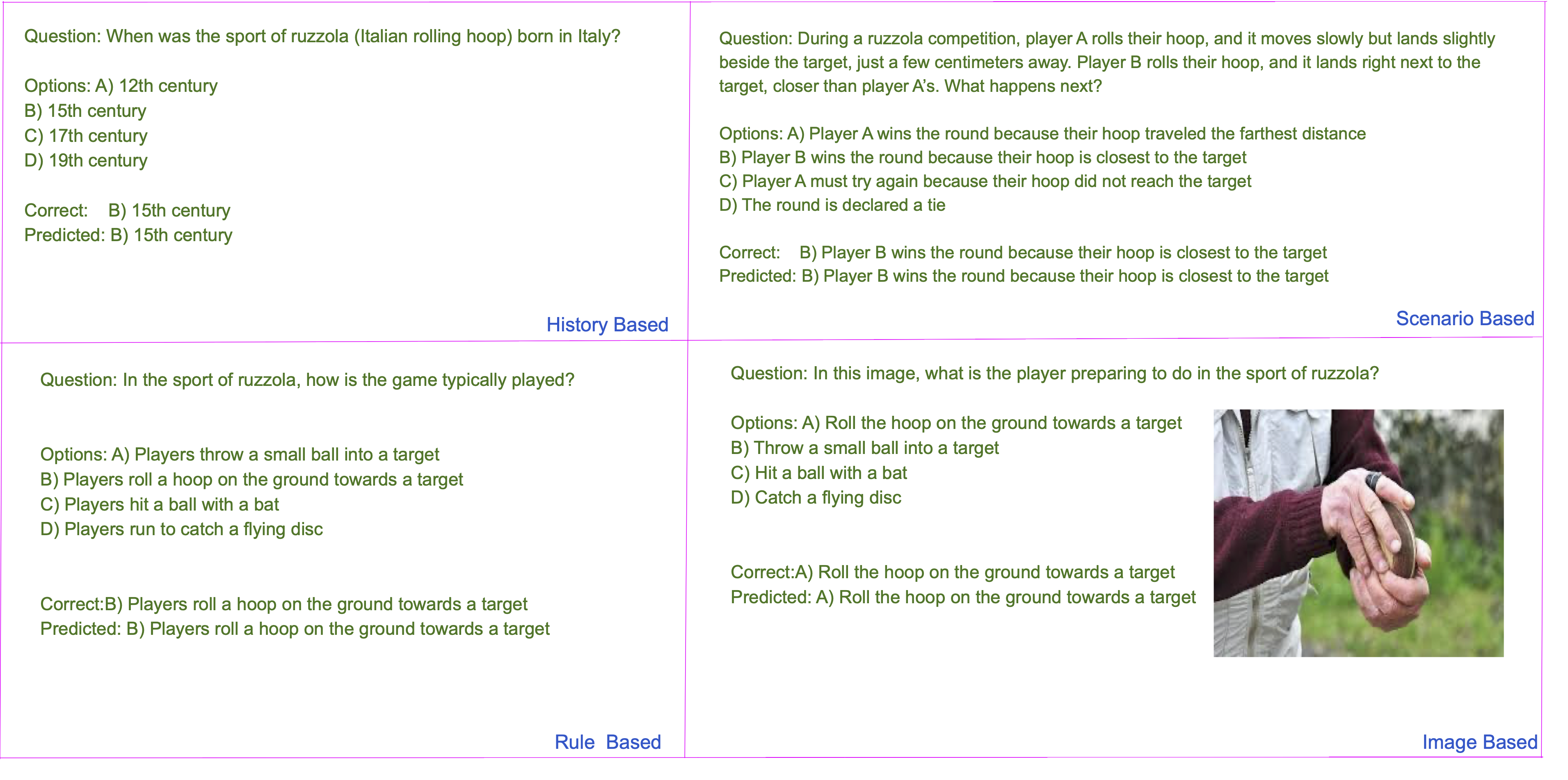}
\caption{Example Illustration of Italy Traditional Sports Correct Prediction (In English)} 
\label{ItaTSCPE}
\end{figure*}

\begin{figure*}
\includegraphics[width=\textwidth]{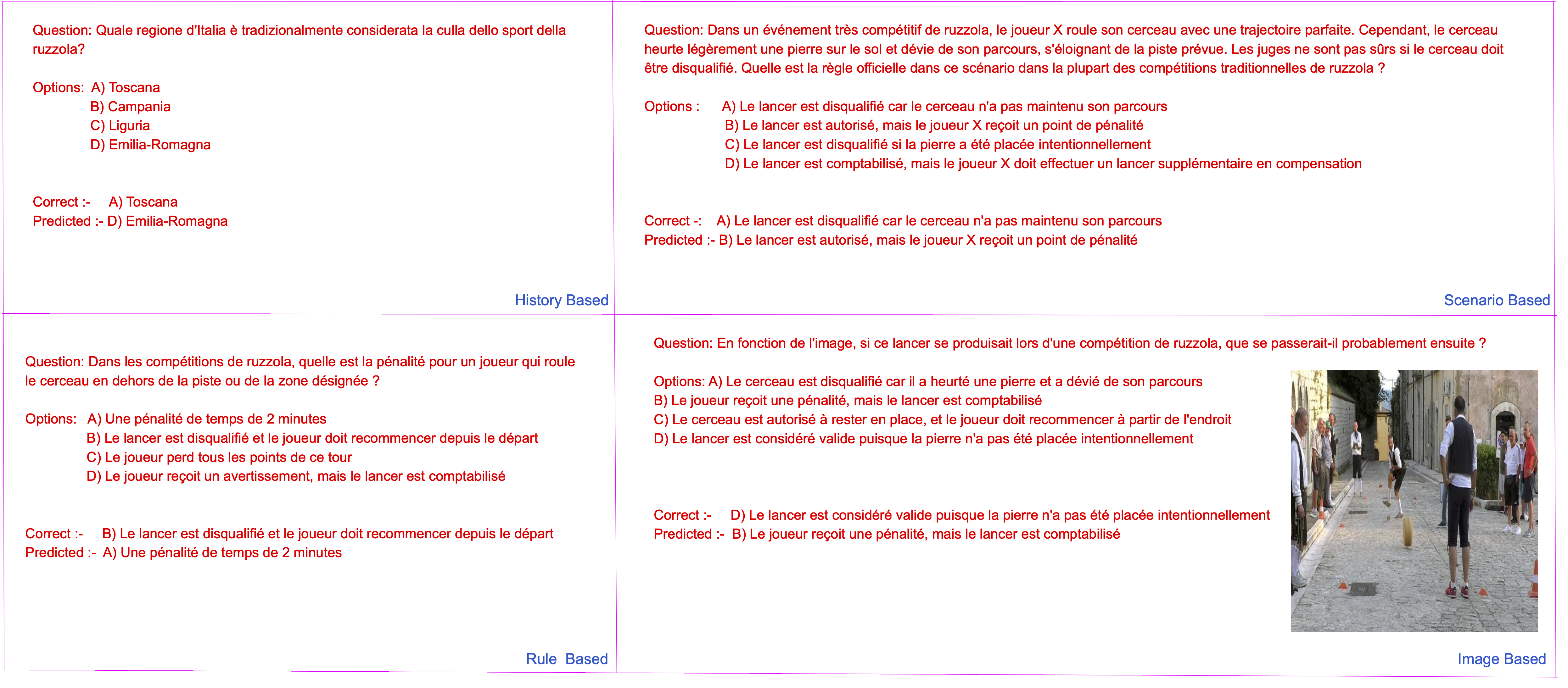}
\caption{Example Illustration of Italy Traditional Sports Wrong Prediction} 
\label{ItaTSWP}
\end{figure*}

\begin{figure*}
\includegraphics[width=\textwidth]{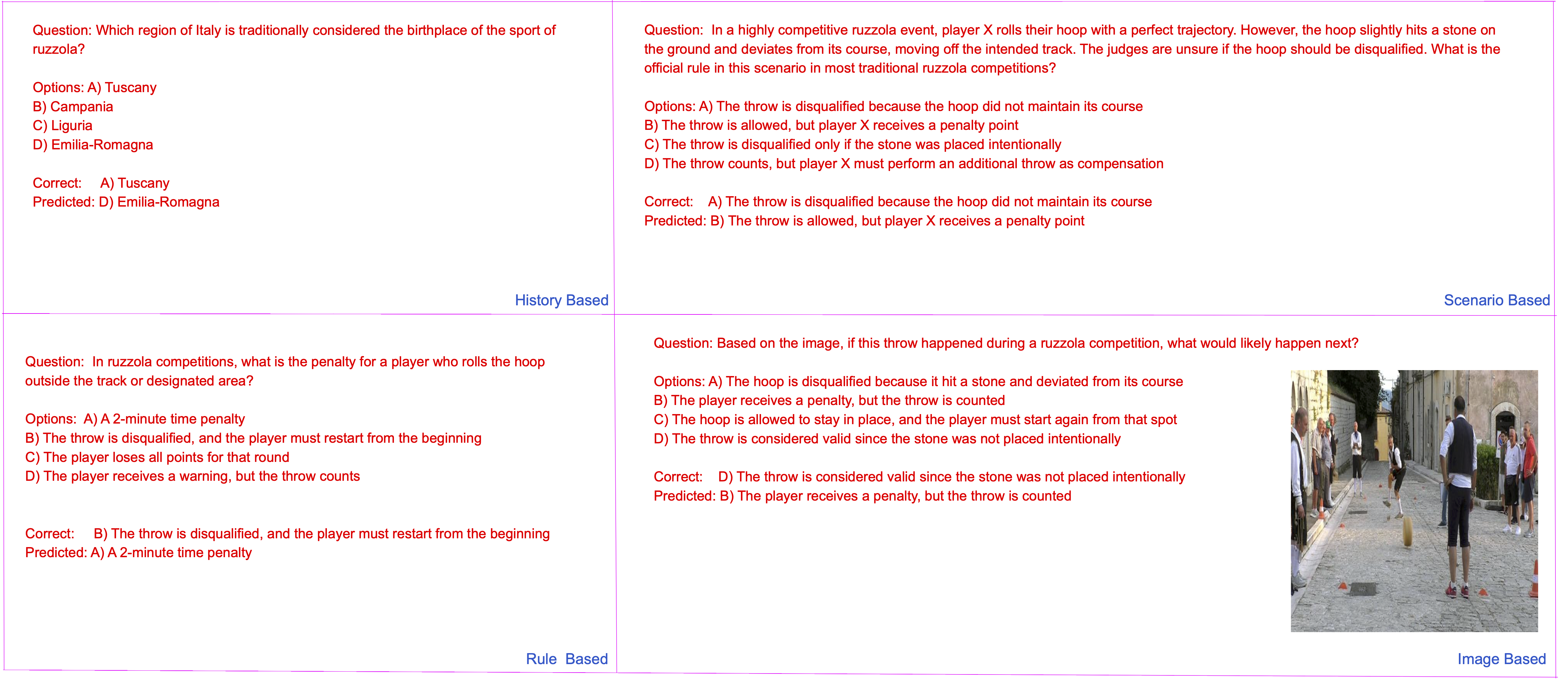}
\caption{Example Illustration of Italy Traditional Sports Wrong Prediction (In English)} 
\label{ItaTSWPE}
\end{figure*}

\begin{figure*}
\includegraphics[width=\textwidth]{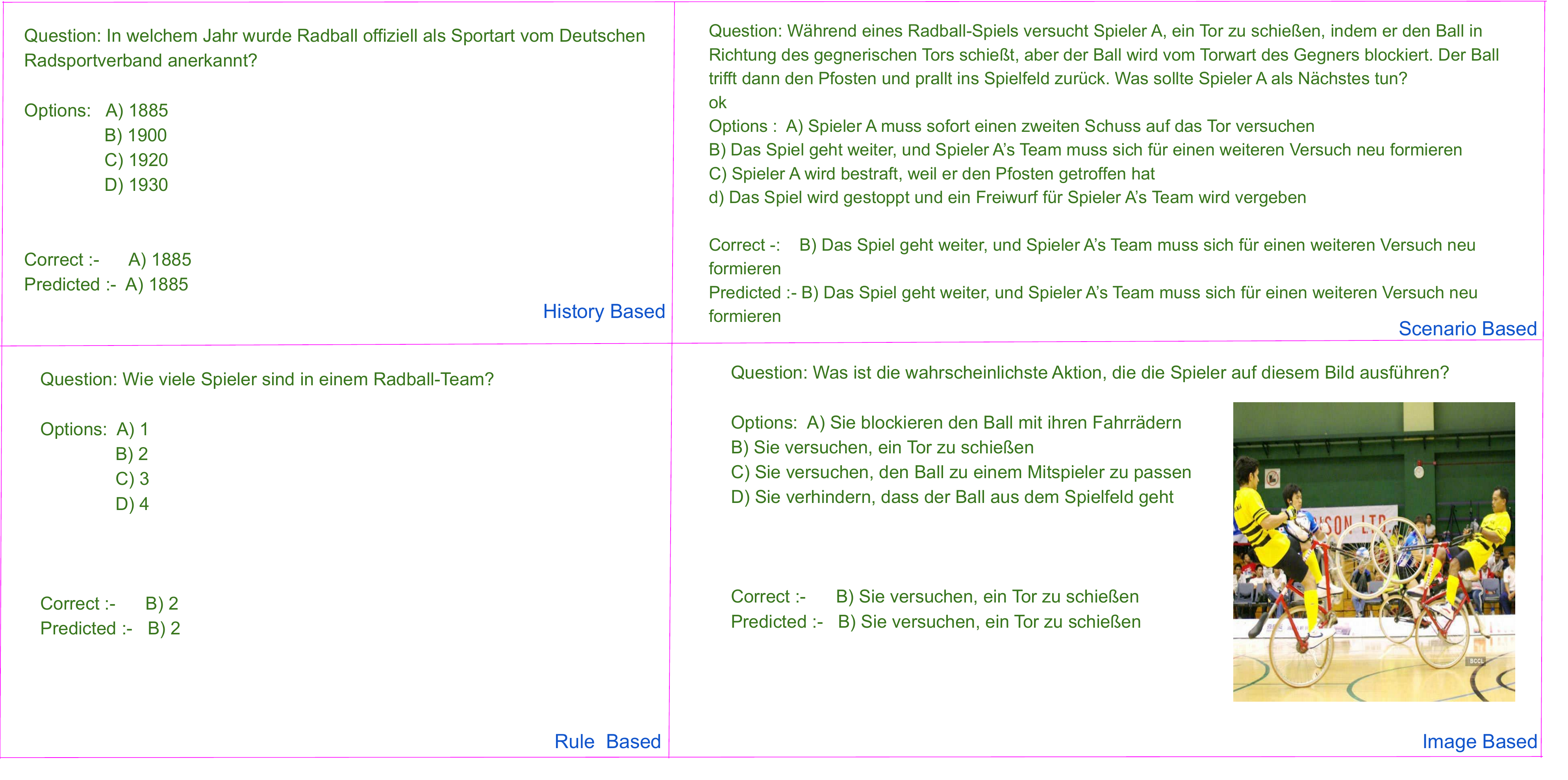}
\caption{Example Illustration of Germany Traditional Sports Correct Prediction} 
\label{GTSCP}
\end{figure*}

\begin{figure*}
\includegraphics[width=\textwidth]{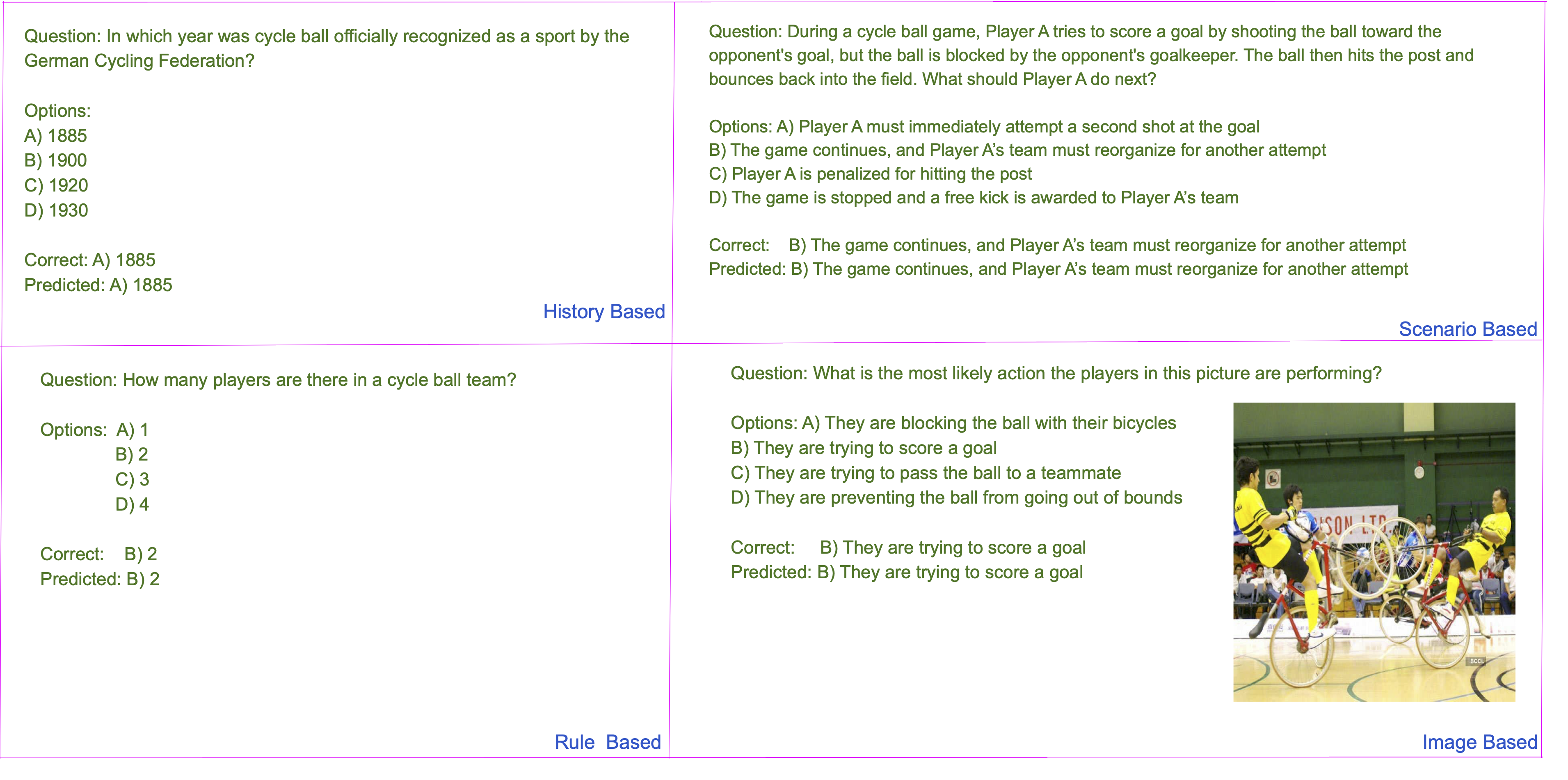}
\caption{Example Illustration of Germany Traditional Sports Correct Prediction (In English)} 
\label{GTSCPE}
\end{figure*}

\begin{figure*}
\includegraphics[width=\textwidth]{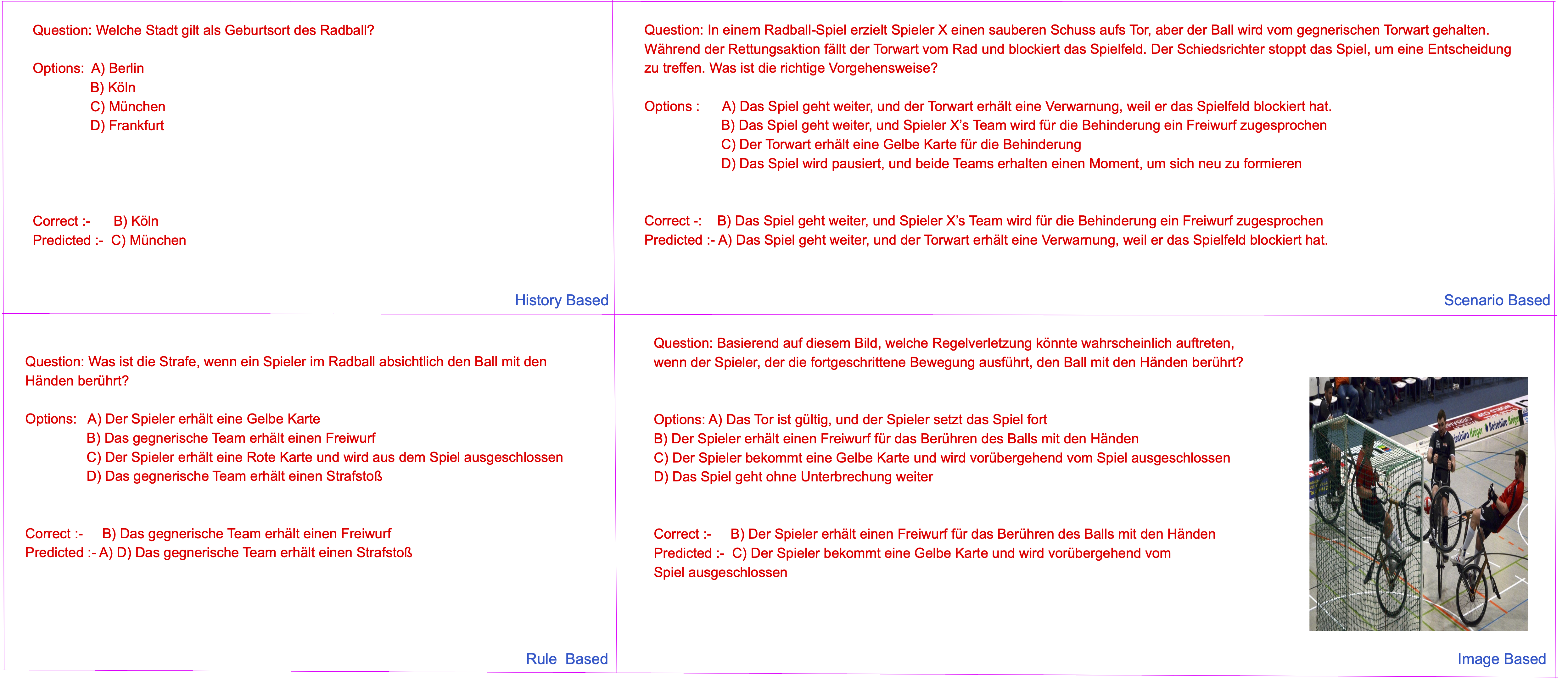}
\caption{Example Illustration of Germany Traditional Sports Wrong Prediction} 
\label{GTSWP}
\end{figure*}

\begin{figure*}
\includegraphics[width=\textwidth]{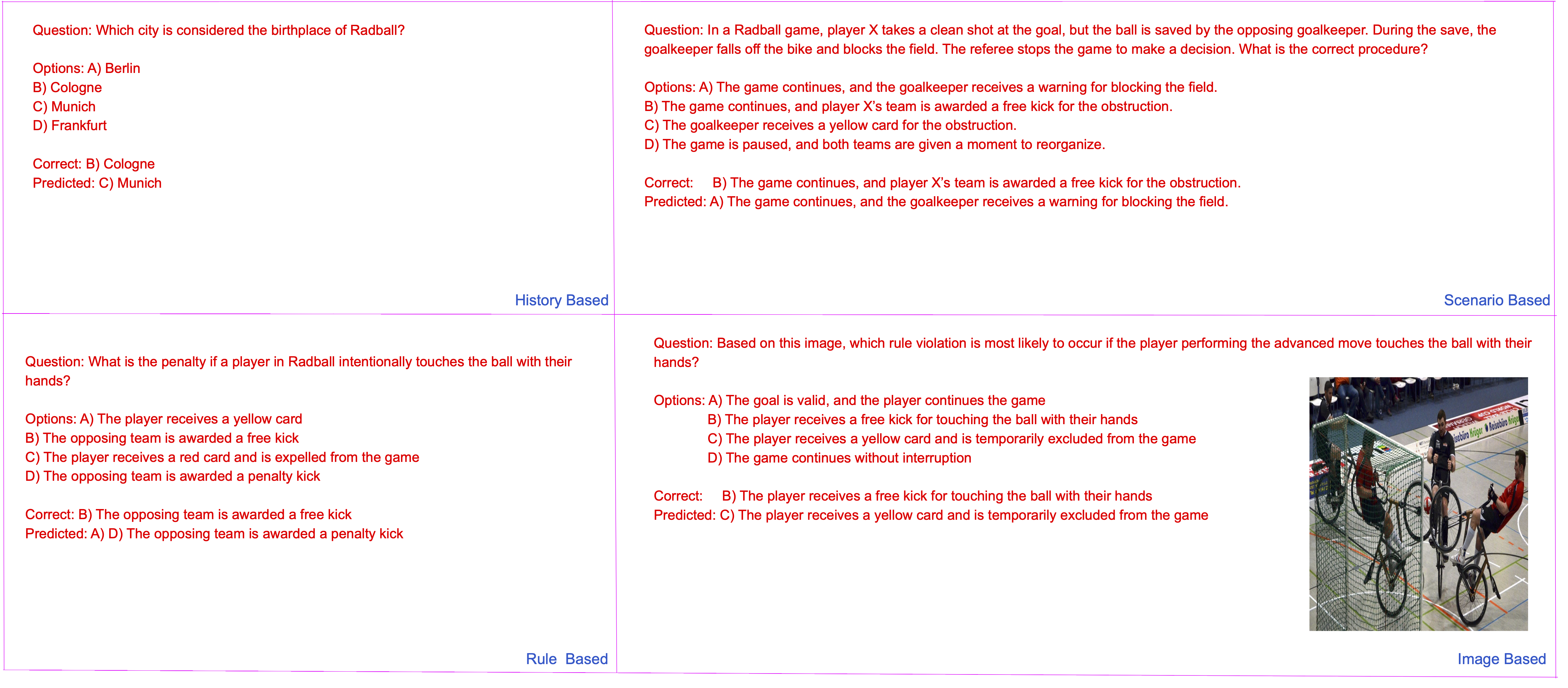}
\caption{Example Illustration of Germany Traditional Sports Wrong Prediction (In English)} 
\label{GTSWPE}
\end{figure*}

\end{document}